\documentclass[10pt, twocolumn, twoside]{IEEEtran}
\usepackage{amsmath,amsfonts}
\usepackage{algorithmic}
\usepackage{array}
\usepackage{textcomp}
\usepackage{stfloats}
\usepackage{url}
\usepackage{verbatim}
\usepackage{graphicx}
\usepackage{multirow}
\usepackage[hidelinks]{hyperref}

\usepackage[table]{xcolor}
\usepackage{subcaption}
\usepackage{lipsum}
\newcommand{\tabincell}[2]{\begin{tabular}{@{}#1@{}}#2\end{tabular}}
\usepackage{makecell}

\renewcommand{\baselinestretch}{0.96}
\newtheorem{definition}{Definition}
\begin{document}
\title{Thinking inside the Convolution for Image Inpainting: Reconstructing Texture via Structure under Global and Local Side}
\author{Haipeng Liu, Yang Wang, ~\IEEEmembership{Senior Member, IEEE}, Biao Qian, Yong Rui, ~\IEEEmembership{Fellow, ACM, IEEE} and Meng Wang, ~\IEEEmembership{Fellow, IEEE}
\thanks{Haipeng Liu, Yang Wang, Meng Wang are with School of Computer Science and Information Engineering, Hefei University of Technology, China. E-mail: hpliu\_hfut@hotmail.com, yangwang@hfut.edu.cn, eric.mengwang@gmail.com. Correspondence to Yang Wang.}
\thanks{Biao Qian is with Department of Automation, Tsinghua University, China. E-mail: hfutqian@gmail.com.}
\thanks{Yong Rui  is with Lenovo Research, China. E-mail: yongrui@lenovo.com.}}

\markboth{IEEE TRANSACTIONS ON IMAGE PROCESSING}%
{Shell \MakeLowercase{\textit{et al.}}: A Sample Article Using IEEEtran.cls for IEEE Journals}


\maketitle

\begin{abstract}
Image inpainting has earned substantial progress, owing to the encoder-and-decoder pipeline, which is benefited from the Convolutional Neural Networks (CNNs) with convolutional downsampling to inpaint the masked regions semantically from the known regions within the encoder, coupled with an upsampling process from the decoder for final inpainting output. Recent studies intuitively identify the high-frequency structure and low-frequency texture to be extracted by CNNs from the encoder, and subsequently for a desirable upsampling recovery. However, the existing arts inevitably overlook the information loss for both structure and texture feature maps during the convolutional downsampling process, hence suffer from a non-ideal upsampling output. In this paper, we systematically answer whether and how the structure and texture feature map can mutually help to alleviate the information loss during the convolutional downsampling. Given the structure and texture feature maps, we adopt the statistical normalization and denormalization strategy for the reconstruction guidance during the convolutional downsampling process, and report the followings: 1) the structure feature map can well reconstruct the texture feature map rather than the inverse; 2) we further motivate the global and local residual structure feature maps, along with local and global texture normalization feature maps, while systematically claim that reconstructing texture feature map under global (local) normalization via global (local residual) structure feature map to be fused is the best for convolutions, as it can essentially alleviate the texture feature map loss via the reconstruction (denormalization) from structure feature map under both global and local side; 3) observing that reconstructing texture feature maps from global and local residual structure feature maps dominate at different stages,  we balance the reconstruction over global and local texture feature maps via a cross-layer balance module for upsampling. The extensive experimental results validate its advantages to the state-of-the-arts over the images from low-to-high resolutions including 256*256 and 512*512, especially holds by substituting all the encoders by ours. Our code is available at \href{https://github.com/htyjers/ConvInpaint-TSGL}{https://github.com/htyjers/ConvInpaint-TSGL}.
\end{abstract}


\makeatletter
\newcommand{\xRightarrow}[2][]{\ext@arrow 0359\Rightarrowfill@{#1}{#2}}
\makeatother

\makeatletter
\newcommand{\rmnum}[1]{\romannumeral #1}
\newcommand{\Rmnum}[1]{\expandafter\@slowromancap\romannumeral #1@}
\makeatother

\newcommand\blfootnote[1]{%
\begingroup
\renewcommand\thefootnote{}\footnote{#1}%
\addtocounter{footnote}{-1}%
\endgroup
}

\section{Introduction}
\label{intro}
\IEEEPARstart{I}{mage}  inpainting attempts to inpaint the masked regions with the semantic contents of the unmasked regions for the image, which benefits many applications, e.g., deepfake detection via inpainting artifacts and inpainting-resistant watermarking. The traditional diffusion-based and patch-based methods can only inpaint the small mask or repeated patterns via simple color information over pixel level,  while fail to exploit the high-level semantics of the unmasked regions. To remedy that, substantial attention \cite{haoran2023fine, zeng2021generative, liu2018image, pathak2016context, xie2019image, yu2019free, Nazeri_2019_ICCV, yu2018generative} has shifted to the convolutional neural networks (CNNs) to encode the semantics to fill the masked regions, and subsequently conduct an upsampling process by a decoder as the final image inpainting output, yielding an encoder-decoder architecture.

Following this pipeline, recent studies \cite{liuone, liu2022delving, yu2022unbiased, Liao_2021_CVPR, li2020recurrent, yan2019PENnet, yi2020contextual, Liu2019MEDFE, wan2021high, li2022misf, wang2021parallel,wang2022progressive}  further specialize the high-frequency structure \cite{liu2022delving, yu2022unbiased, Liao_2021_CVPR, Liu2019MEDFE, wan2021high, wang2021parallel} and low-frequency texture \cite{liu2025few, li2020recurrent, yan2019PENnet, yi2020contextual, li2022misf} as the crucial information to be encoded via CNNs for image inpainting,
since they both encode complementary information, \textit{i.e.,} high-frequency and low-frequency,  for the image inpainting, especially the unmasked regions. Upon the CNNs as the encoder, a varied of strategies, such as self-attention mechanism \cite{deng2022hourglass, li2022mat} are proposed to be processed by decoder for image inpainting.  \cite{cao2022learning} distill \cite{10476709} informative texture feature map from pre-trained masked auto-encoder (MAE) \cite{he2022masked} or ground truth during convolutional downsampling; \cite{Liu2019MEDFE} equalizes the structure feature maps from deep layers and texture feature maps from shallow layers via channel and spatial equalization, then fused as input to the decoder for upsampling. \cite{yu2022unbiased, wang2021parallel} guide the decoder upsampling via the generated multi-scale \cite{wang2022progressive} structure and texture feature maps from CNNs.

\begin{figure}
\setlength{\abovecaptionskip}{0pt}
\setlength{\belowcaptionskip}{0pt}
  \centering
  \includegraphics[width=\linewidth]{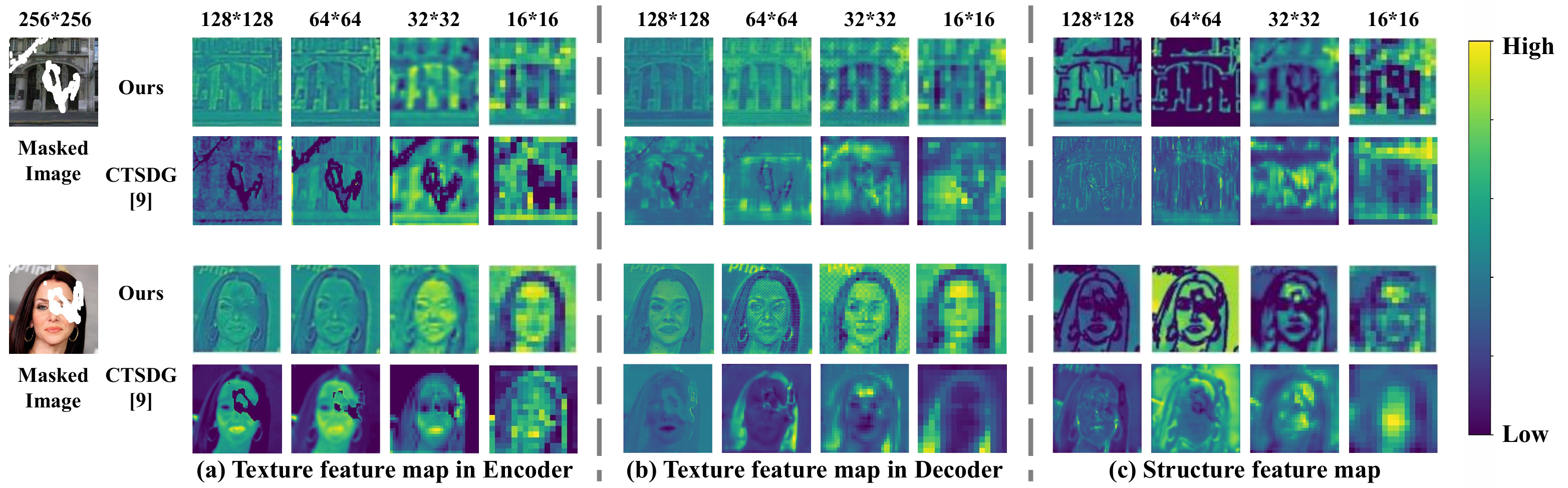}
  \caption{CTSDG \cite{Guo_2021_ICCV}  suffers from the non-ideal inpainting results due to mutual guidance between the global structure and texture {feature map} in decoder,  where the sparse structure {feature map} is broken down via the fusion from the texture (b) {feature map}, while the texture {feature map} receives no effective guidance from the structure (c) {feature map}. The higher value within the yellow area indicates more structure or texture information from feature map. Our methods preserves more information over structure and texture feature maps. Note that the feature maps in the decoder are the result of convolutional downsampling (a), resulting into the feature map loss.}\label{ctsdg}
\vspace{-15pt}
\end{figure}

Recently, Guo \textit{et al}.~\cite{Guo_2021_ICCV} proposes the mutual guidance between the structure and texture via a two-stream encoder-decoder architecture, where each encoder extracts the structure (texture) semantics via CNNs, while serves as the guidance for the upsampling from the decoder to fuse the texture (structure) feature map.  Despite the intuitions, as revealed by existing arts \cite{yu2022unbiased}, the structure semantics is always sparse,  directly fusing them can easily break down the sparse structure feature map; worse still, it remains unclear on \textit{whether} and \textit{how} the structure and texture can truly help each other for image inpainting, especially alleviating the texture feature map information loss during the convolutional process. All the above methods totally overlook the structure-and-texture feature map information loss during the convolutional downsampling process from the encoder, \emph{especially the texture feature map loss}, together with less attention on whether and how they can help each other during the convolutional process to alleviate the texture and structure feature map information loss, in particular texture feature map loss, which in essence is critical to upsampling process to decoder.

Dong \textit{et al}.~\cite{dong2022incremental} attempts to preserve the texture feature map information during the down-and-up sampling process, by combining two types of high-frequency structure feature map, to further fuse the texture feature map during the convolutional process, to alleviate the texture feature map information loss during such convolutional process.  However, such strategy fails to 1) consider the loss of the structure feature map within the convolutional process, which can benefit the texture feature map information loss during the convolutional process; 2) as aforementioned, such simple fusing manner is problematic, as it can easily suppress the high-frequency sparsity information with structure feature map, hence may not help preserve the texture feature map well during the convolutional process.

\begin{figure*}
\setlength{\abovecaptionskip}{0pt}
\setlength{\belowcaptionskip}{0pt}
  \centering
  \includegraphics[width=0.7\textwidth]{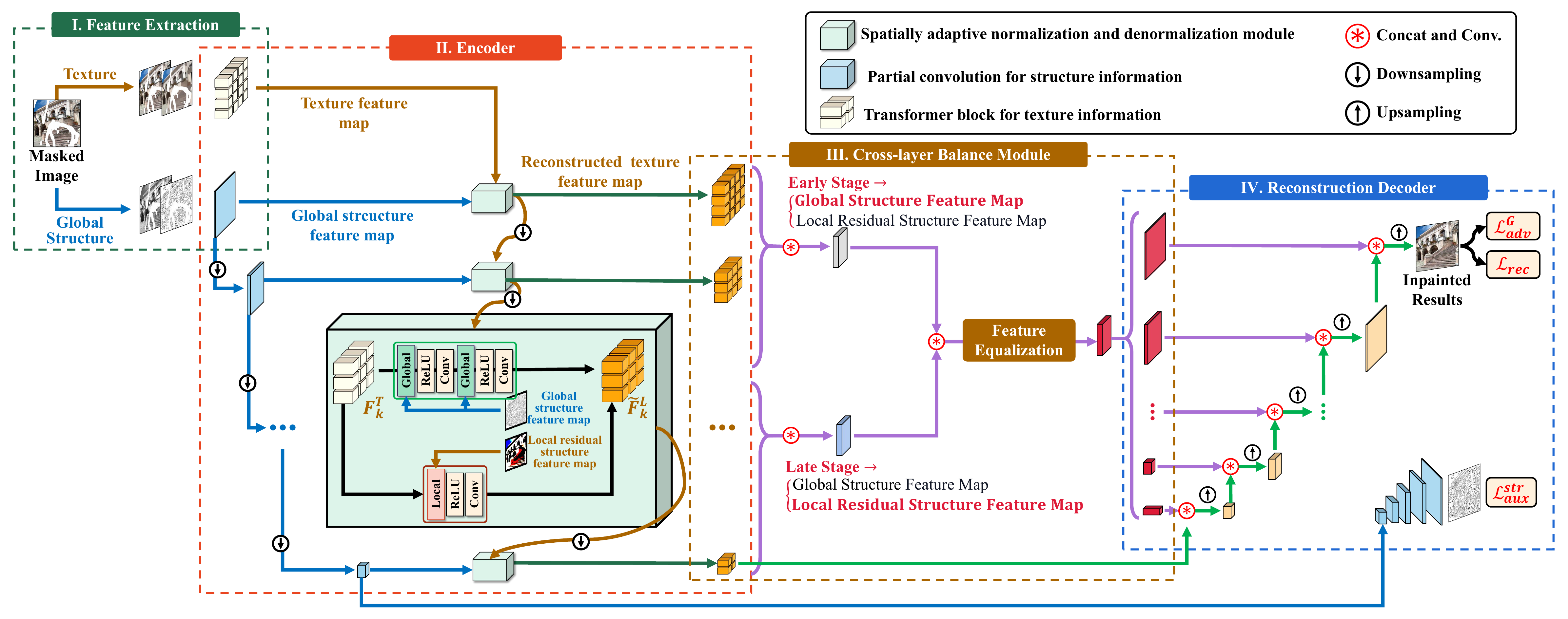}
  \caption{Illustration of our proposed overall pipeline. Our basic idea is to reconstruct  the texture feature map via the structure feature map under both global and local side during the convolutional downsampling process from \emph{encoder} (\Rmnum{2}), which is achieved via the statistical normalization and denormalization strategy between them. We augment the guidance from global structure feature map \emph{twice} to keep balance with local residual structure feature map, upon the \emph{cross-layer balance module} (\Rmnum{3}) via the feature equalization strategy to benefit the upsampling from \emph{decoder} (\Rmnum{4}); see Sec.\ref{our_arch} for more details.}\label{architecture}
\vspace{-15pt}

\end{figure*}
The above standing-out problem naturally promotes us to delve into the following questions: \textit{whether and how can the structure and texture can help each other to alleviate the information loss, especially the loss for texture feature map, during the convolutional downsampling process within encoder?} That will further benefit the subsequent decoder process for final image inpainting output. To answer the question,  given the structure and texture feature map within one layer upon CNNs as encoder,  we propose the following:
Instead of simply fusing texture and structure feature map during the convolutional downsampling process from the encoder to suppress sparse structure feature map,  we adopt the spatially-adaptive normalization and denormalization strategy (formulated as Eq.\ref{normalization}) to reconstruct the texture feature map from structure feature map after the convolutional downsampling for each layer, to incorporate the statistical information of structure feature map over the texture feature map across varied layers to alleviate the information loss for texture feature map, while find that the structure feature map can better reconstruct the texture feature map rather than the inverse, since the structure is always \textit{sparse} that can be easily broken down via the reconstruction from \textit{dense} texture feature map. Instead, such sparsity can promote the dense texture's reconstruction. Hence, the existing art \cite{Guo_2021_ICCV} on mutually guidance between structure and texture feature maps \textit{failed to} exhibit the desirable inpainting output, as validated in Fig.\ref{ctsdg};

Motivated by the above, we propose to reconstruct the texture feature map via structure feature map, by global and local normalization over texture feature map for denormalization via structure feature map; given the texture feature map, \textit{global} and \textit{local} normalization are defined as:
    \begin{definition}
    The \textit{global} normalization computes the statistical mean and variance of the pixels over the whole texture feature map from any channel (see Eq.\ref{global_normalization} for more details); while the \textit{local} normalization computes the statistical mean and variance of the pixels across different channels given the specific position (see Eq.\ref{local_normalization} for more details).
    \end{definition}

Based on the above, the global normalization can highlight the \emph{global} statistical property for the whole feature map, named \textit{global texture} feature map, as illustrated in Fig.\ref{global_local_texture}(a); while local normalization can highlight the \emph{local} statistical property for each position, named \textit{local texture} feature map, as illustrated in Fig.\ref{global_local_texture}(b). Afterwards, the denormalization is performed to reconstruct the texture feature map by incorporating the statistical information of the structure feature map based on both global and local normalization, while observing that the structure feature map can better reconstruct the \textit{local texture normalization than the global one}, as the structure feature map prefers the global sparsity so as to better complement the local dense texture feature map information via local normalization;

The above observation further motivates us on how to reconstruct the global texture feature map normalization, we propose to extract the residual local structure from the texture feature map rather than the global structure feature map above. Taken all together, we discuss 4 variant modules to deeply understand the texture feature map normalization and structure denormalization under both global and local side, while interestingly find that the reconstruction of texture feature map under global normalization from structure feature map, paired with the reconstruction from local residual structure feature map to texture feature map under local normalization is the best. The reconstructed two texture feature maps  from global and local residual structure are fused together via element-wise addition to be then convolved for downsampling to the next layer, the output is further divided into two stream as per two texture normalization strategies, ready to be reconstructed by global and local residual structure feature map for next layer;

We further observe that, at the early stage of the whole convolutional process within the encoder, the information from global structure feature map decreases \textit{slower} than that from the local residual structure feature map, while surpasses that at the late stage, hence we propose to: augment the reconstruction from global structure feature map \textit{twice} to keep balance with local structure feature map, while group the feature maps into two groups as per global and local residual structure, coupled with a feature equalization in cross-layer balance module for upsampling from the decoder, as shown in Fig.\ref{architecture}.

The extensive experiments validate its superiority to all SOTAs, while still holds throughout substituting the encoders by ours from all SOTAs.
Notably, our strategy can be well applied to the masked images from low-to-high resolutions including 256*256 and 512*512 over the SOTAs, this is because it can alleviate the texture feature map loss when the convolutional downsampling progresses to deeper layers, so as to capture the semantic information. Instead, the conventional methods failed to capture that within only shallow layers, as they cannot preserve the information loss for texture feature maps during convolutional downsampling process.

\section{Related Work} \label{rework}
To address the limitations of traditional diffusion-based and patch-based methods, recent advancements in image inpainting have focused on encoder-decoder paradigms \cite{10476709,qian2023adaptive,qian2022switchable,qian2023rethinking}. These approaches specialize in capturing both low-frequency texture , such as PENNet \cite{yan2019PENnet}, RFR \cite{li2020recurrent}, HAN \cite{deng2022hourglass}, FAR \cite{cao2022learning}, HiFill \cite{yi2020contextual}, MI-GAN \cite{Sargsyan_2023_ICCV}, and high-frequency structure like MEDFE \cite{Liu2019MEDFE}, MMT \cite{yu2022unbiased}, and DGTS \cite{liu2022delving}, StrDiffusion \cite{Liu_2024_CVPR}, as the crucial information to be encoded via CNNs for image inpainting. \textit{In particular, MI-GAN \cite{Sargsyan_2023_ICCV} distills an informative texture feature map from the heavyweight neural network, CoModGAN \cite{zhao2021large}, using multiple convolutional kernels in each downsampling layer to closely match the network’s capacity with that of its teacher network. However, this texture-only method overlooks structural information, and the extreme lightweight design sacrifices the performance, while suffers from heavy computational complexity for training, which cost nearly one month with 8 Nvidia A6000 GPUs.} PENNet \cite{yan2019PENnet} introduces a cross-layer attention module to compute attention scores across deep encoder layers and performs patch substitution based on these scores in shallow layers before final upsampling. MEDFE \cite{Liu2019MEDFE} equalizes structure and texture feature maps from deep and shallow layers before fusing them in the decoder for upsampling.

More recently, approaches such as CTSDG \cite{Guo_2021_ICCV} proposes mutual guidance between structure and texture feature maps, where each encoder extracts structure (texture) feature maps via CNNs and guides upsampling to fuse with each other. Meanwhile, ZITS \cite{dong2022incremental} aims to preserve texture feature maps during down-and-up sampling by combining two types of high-frequency structure feature maps and fusing texture feature maps during convolutional processes to mitigate information loss. However, it remains unclear whether and how the structure and texture feature maps can truly help each other for image inpainting, especially in alleviating the texture feature map loss during the convolutional process. Based on the aforementioned reasons, extending previous methods to higher-resolution image inpainting poses a challenge. To this end, LaMa \cite{suvorov2022resolution} and MAT \cite{li2022mat} endeavor to achieve a global receptive field in the early layers of the texture feature map.

However, LaMa encounters the difficulties of capturing advanced semantic information due to fewer convolutional downsampling layers, resulting into a tendency to generate the stable repeating texture feature maps rather than clear inpainted regions. In contrast, our method reconstructs the texture feature map under global (local) normalization via global (local residual) structure feature map, which offers several critical advantages: 1) the global texture feature map under global normalization is augmented by the reconstruction from the global structure feature map; 2) similarly, the local texture feature maps under local normalization is augmented by the reconstruction from the local residual structure feature map;  3) we balance the global and local residual structure feature map to reconstruct texture feature map for upsampling to decoder.

\section{Overall Framework}
For the input image $I_{gt} \in \mathbb{R}^{3\times H \times W}$ with mask $M \in \mathbb{R}^{1\times H \times W}$(0 for masked, 1 for unmasked), image inpainting transforms the masked image $I_{m} = I_{gt} \odot M$ into a complete image $I_{out}$.  Central to our pipeline, illustrated in Fig.\ref{architecture}, is what should be convolved upon high frequency structure and low-frequency texture feature map during the convolutional downsampling process, so as to alleviate their feature map loss, especially the texture feature map.  Before shedding light on our architecture, we describe how to extract the multi-scale feature maps of structure and texture w.r.t. different layers for Convolutional Neural Networks (CNNs) in Sec.\ref{msf}, followed by how they collaborate during the convolutional downsampling process in Sec.\ref{ilb}.
Finally, we offer the overall loss functions in Sec.\ref{loss_fun}.

\begin{table*}
\setlength{\abovecaptionskip}{0pt}
\setlength{\belowcaptionskip}{0pt}
\centering
\caption{Illustration of why the structure feature map can better reconstruct the texture feature map than the inverse. Notably, the texture feature map (marked by the red box) is well reconstructed by the structure feature map (a); while the sparse structure feature map is broken down from the reconstruction over the dense texture feature map (b).  FID evaluates the perceptual quality by measuring the feature distribution difference between the inpainted image and ground truth, $\downarrow$: Lower is better. The higher value within the yellow area indicates more structure or texture information from feature map.}\label{lgh}
\resizebox{0.8\textwidth}{!}{\begin{tabular}{c||c|c|c|c||c|c|c|c||c|c|c}
\hline
\multirow{2}{*}{{Method}}
& \multicolumn{4}{c||}{{Fréchet Inception Score (FID) $\downarrow$}} & \multicolumn{4}{c||}{{Feature Visualization}} &  \multicolumn{3}{c}{{Inpainted}}\\
&{10-20\%} & {20-30\%} &{30-40\%} &{40-50\%} &{256 $\times$ 256} &{128 $\times$ 128} &{64 $\times$ 64} &{32 $\times$ 32} &  \multicolumn{3}{c}{{Results}}
\\\hline\hline
 \multicolumn{12}{c}{\cellcolor{blue!20}{\textbf{(a) Texture  {feature map} reconstruction via structure  {feature map}}}} \\\hline
\textit{\textbf{w. structure feature map}}  &\cellcolor{green!20}\textbf{23.18} &\cellcolor{green!20}\textbf{29.71} &\cellcolor{green!20}\textbf{40.52} &\cellcolor{green!20}\textbf{57.37}
    & \multirow{2}{*}{\begin{minipage}{0.12\columnwidth}
        \centering
    	{\includegraphics[width=\columnwidth]{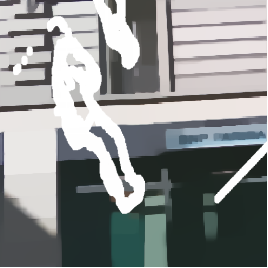}
            \\\textbf{Texture}}
	   \end{minipage}}
    & \begin{minipage}{0.12\columnwidth}
        \centering
    	\raisebox{-.5\height}{\includegraphics[width=\columnwidth]{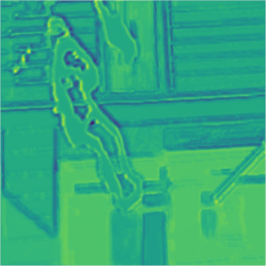}}
	   \end{minipage}
    & \begin{minipage}{0.12\columnwidth}
        \centering
    	\raisebox{-.5\height}{\includegraphics[width=\columnwidth]{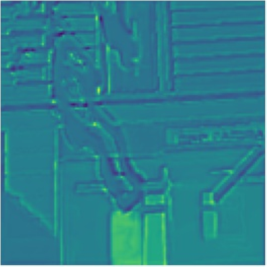}}
	   \end{minipage}
    & \begin{minipage}{0.12\columnwidth}
        \centering
    	\raisebox{-.5\height}{\includegraphics[width=\columnwidth]{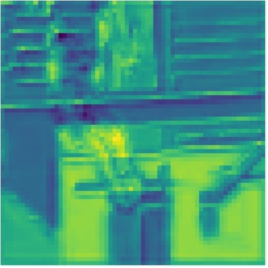}}
	   \end{minipage}
    & \begin{minipage}{0.12\columnwidth}
        \centering
    	\raisebox{-.5\height}{\includegraphics[width=\columnwidth]{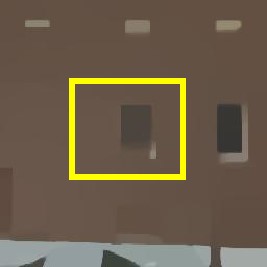}}
	   \end{minipage}
    & \begin{minipage}{0.12\columnwidth}
        \centering
    	\raisebox{-.5\height}{\includegraphics[width=\columnwidth]{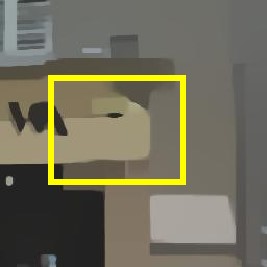}}
	   \end{minipage}
    & \begin{minipage}{0.12\columnwidth}
        \centering
    	\raisebox{-.5\height}{\includegraphics[width=\columnwidth]{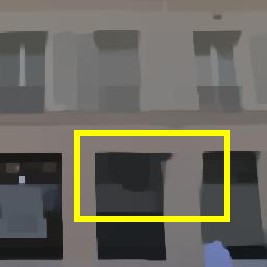}}
	   \end{minipage} \\
\textit{\textbf{w/o. structure feature map}} &26.55 &31.33 &42.64 &58.35 &
    & \begin{minipage}{0.12\columnwidth}
        \centering
    	\raisebox{-.5\height}{\includegraphics[width=\columnwidth]{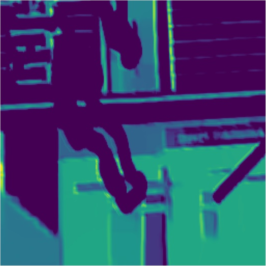}}
	   \end{minipage}
    & \begin{minipage}{0.12\columnwidth}
        \centering
    	\raisebox{-.5\height}{\includegraphics[width=\columnwidth]{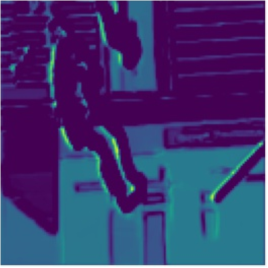}}
	   \end{minipage}
    & \begin{minipage}{0.12\columnwidth}
        \centering
    	\raisebox{-.5\height}{\includegraphics[width=\columnwidth]{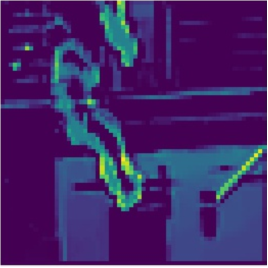}}
	   \end{minipage}
    & \begin{minipage}{0.12\columnwidth}
        \centering
    	\raisebox{-.5\height}{\includegraphics[width=\columnwidth]{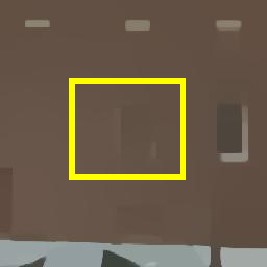}}
	   \end{minipage}
    & \begin{minipage}{0.12\columnwidth}
        \centering
    	\raisebox{-.5\height}{\includegraphics[width=\columnwidth]{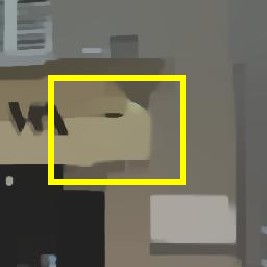}}
	   \end{minipage}
    & \begin{minipage}{0.12\columnwidth}
        \centering
    	\raisebox{-.5\height}{\includegraphics[width=\columnwidth]{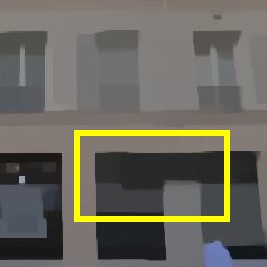}}
	   \end{minipage} \\ \hline\hline

 \multicolumn{12}{c}{\cellcolor{blue!20}{\textbf{(b) Structure  {feature map} reconstruction via texture  {feature map}}}}\\
\textit{\textbf{w. texture feature map}}   &15.41 &29.76 &43.49 &64.04
    & \multirow{2}{*}{\begin{minipage}{0.12\columnwidth}
        \centering
    	{\includegraphics[width=\columnwidth]{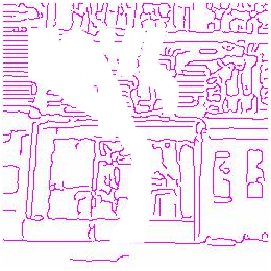}
            \\\textbf{Structure}}
	   \end{minipage}}
    & \begin{minipage}{0.12\columnwidth}
        \centering
    	\raisebox{-.5\height}{\includegraphics[width=\columnwidth]{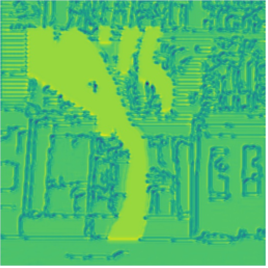}}
	   \end{minipage}
    & \begin{minipage}{0.12\columnwidth}
        \centering
    	\raisebox{-.5\height}{\includegraphics[width=\columnwidth]{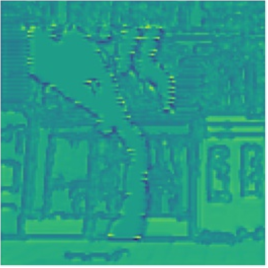}}
	   \end{minipage}
    & \begin{minipage}{0.12\columnwidth}
        \centering
    	\raisebox{-.5\height}{\includegraphics[width=\columnwidth]{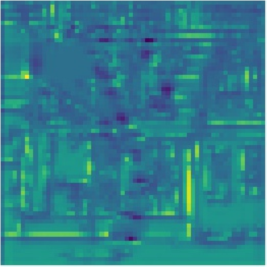}}
	   \end{minipage}
    & \begin{minipage}{0.12\columnwidth}
        \centering
    	\raisebox{-.5\height}{\includegraphics[width=\columnwidth]{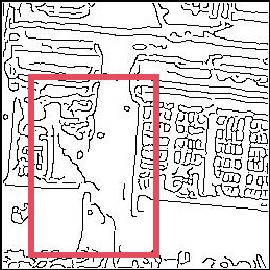}}
	   \end{minipage}
    & \begin{minipage}{0.12\columnwidth}
        \centering
    	\raisebox{-.5\height}{\includegraphics[width=\columnwidth]{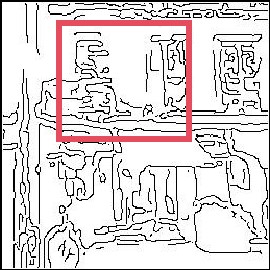}}
	   \end{minipage}
    & \begin{minipage}{0.12\columnwidth}
        \centering
    	\raisebox{-.5\height}{\includegraphics[width=\columnwidth]{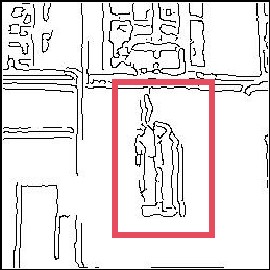}}
	   \end{minipage} \\
\textit{\textbf{w/o. texture feature map}}&\cellcolor{green!20}\textbf{15.00} &\cellcolor{green!20}\textbf{29.64} &\cellcolor{green!20}\textbf{42.97} &\cellcolor{green!20}\textbf{63.69} &
    & \begin{minipage}{0.12\columnwidth}
        \centering
    	\raisebox{-.5\height}{\includegraphics[width=\columnwidth]{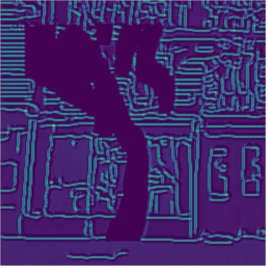}}
	   \end{minipage}
    & \begin{minipage}{0.12\columnwidth}
        \centering
    	\raisebox{-.5\height}{\includegraphics[width=\columnwidth]{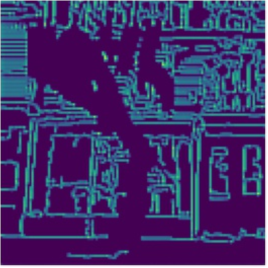}}
	   \end{minipage}
    & \begin{minipage}{0.12\columnwidth}
        \centering
    	\raisebox{-.5\height}{\includegraphics[width=\columnwidth]{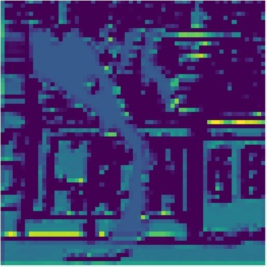}}
	   \end{minipage}
    & \begin{minipage}{0.12\columnwidth}
        \centering
    	\raisebox{-.5\height}{\includegraphics[width=\columnwidth]{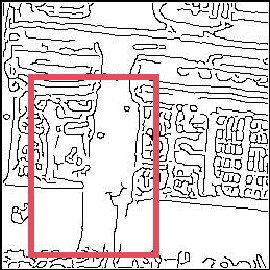}}
	   \end{minipage}
    & \begin{minipage}{0.12\columnwidth}
        \centering
    	\raisebox{-.5\height}{\includegraphics[width=\columnwidth]{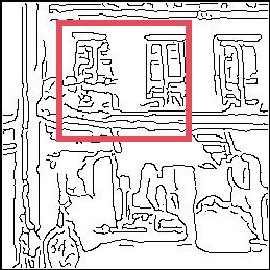}}
	   \end{minipage}
    & \begin{minipage}{0.12\columnwidth}
        \centering
    	\raisebox{-.5\height}{\includegraphics[width=\columnwidth]{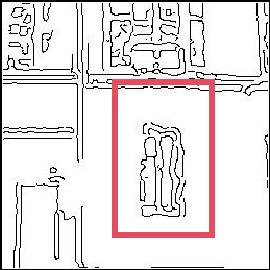}}
	   \end{minipage} \\\hline

\end{tabular}}
\vspace{-15pt}
\end{table*}

\subsection{Extracting Multi-Scale Feature Map of Structure and Texture}\label{msf}
\subsubsection{Partial Convolution for Structure Feature Map}\label{pc}
Given the input image $I_{gt}$, we first employ the canny edge detector \cite{canny1986computational} to construct the edge map of $I_{gt}$, denoted as $I_{S}$,  and the grayscale counterpart  $I_{gray} \in \mathbb{R}^{1\times H \times W}$. Similarly, the edge map of $I_{m}$ is given as $\tilde{I}_{S} = I_{S}\odot M$ and the corresponding grayscale is $\tilde{I}_{gray} = I_{gray}\odot M$ (\textit{more analysis can be seen in Appendix Sec.F}).  Based on that, we adopt the \textit{partial} convolutional layer \cite{liu2018image} to extract the structure feature map from the \textit{unmasked / known} regions for $I_{m}$. Specifically, we concatenate $\tilde{I}_{S}$, $\tilde{I}_{gray}$ and $M$ as the input of the first partial convolution layer, denoted as $X_{S}$, where the corresponding mask is $M$. Given the current sliding window $\mathbf{X_{s}}$ over the corresponding region of  $X_{S}$ and its mask $\mathbf{M_{s}}$, $\frac{\text{sum}(\mathbf{1})}{\text{sum}(\mathbf{M_{s}})}$ denotes the scaling factor to adjust the ratio of the known region, thus the partial convolution operation at every location can be expressed as
\begin{equation}
\label{eq:partconv}
\begin{split}
    x'_{s} = \begin{cases}
     \mathbf{W}^{Tr}(\mathbf{X_{s}}\odot\mathbf{M_{s}})\frac{\text{sum}(\mathbf{1})}{\text{sum}(\mathbf{M_{s}})} + b,  & \text{if }  \text{sum}(\mathbf{M_{s}})>0  \\
        0,  & \text{otherwise}
    \end{cases}
    \end{split}
\end{equation}
where $x'_{s}$ denotes one pixel of the output feature map $F^{S}_{1}$. $\mathbf{W}$ and $b$ are weight matrix and bias vector under convolutional filter; $Tr$ is the transpose operation; $\odot$ represents the element-wise multiplication.
After each partial convolution operation, the mask is then updated below:
\begin{equation}
\label{eq:partconv}
\begin{split}
    m'_{s}=\begin{cases}
        1, & \text{if } \text{sum}(\mathbf{M_{s}})>0,  \\
        0, & \text{otherwise},
    \end{cases}
    \end{split}
\end{equation}
where $m'_{s}$ is the mask value corresponding to $x'_{s}$.
Based on that, we cascade $N$ partial convolutional layers together to inpaint the masks, while updated the feature maps. After passing through the partial convolution layers, the feature maps are processed by a normalization layer and an activation function. Furthermore, the structure information decreases progressively with convolutional downsampling. To reduce such structure loss, we add more structure feature channels to endow the network with the capacity to encode more structure information. Such intuition is validated in \cite{ronneberger2015u} that more feature channels can contain more context information. Alternatively, \cite{deng2022hourglass} proposes the self-attention mechanism to alleviate feature maps loss during convolutional downsampling process. However, unlike the dense texture information, the structure is always sparse so as to be not easily reconstructed as the input of attention mechanism.
Based on the above, we can obtain the structure feature maps from the known regions, \emph{i.e.}, the feature maps from the different partial convolution layers, denoted as $F^{S}_{t}, t=1,2,...,N$, where $N$ is the number of the layers. In the next, we turn to discuss how to extract the multi-scale feature maps of texture information.

\subsubsection{Partial Convolution + Transformer for Texture Feature Map}
Similar in spirit to \cite{ren2019structureflow}, we first remove the structure information and retain the texture information in the input image $I_{gt}$ by adopting the edge-preserved smooth method \cite{xu2011image}.  Accordingly, we denote the edge-preserved smooth image of  ${I}_{gt}$ as ${I}_{T} \in \mathbb{R}^{3\times H \times W}$, while the corresponding masked image is computed by $\tilde{I}_{T} = I_{T}\odot M$.
To well extract the texture feature map with both \textit{global} and \textit{local} pattern, besides the above partial convolution over $\tilde{I}_{T}$ for \textit{local} pattern, we deploy the vision transformer (ViT) to extract the \textit{global} correlation over the patches from texture feature map, specialized by the transformer block \cite{li2022mat} strategy through a total of $N$ convolutional downsampling stages w.r.t. $N$ layers. Notably, the vision transformer is designed to capture global correlations at the patch level within the feature map, hence adaptive to texture features. However, unlike dense texture features, structural information within one patch is often sparse, which can be easily disrupted by surrounding dense information from other patches caused by self-attention from transformer. Therefore, we adopt partial convolution for our scenario.
At the beginning, to facilitate the transformer, we utilize a partial convolutional head consisting of two partial convolutional block for downsampling as the first stage,  to obtain the $1/2$ sized feature map ${F}^{T}_{1}$ used for the tokens, whose input is the concatenation of $I_{m}$, $\tilde{I}_{T}$ and $M$ together, where the first convolution layer is exploited to change the input dimension, the second serves as a downsampled layer to reduce the resolution.

Based on ${F}^{T}_{1}$,  the transformer body is further adopted to process the tokens by building long-range correspondences, which contains $N-1$ stages of the adjusted transformer blocks following the first stage.  For the $l$-th block of $t$-th stage ($t=2,...,N$), the output feature map ${F}^{T}_{t, l+1}$  is obtained by
\begin{equation}
\begin{aligned}
         {F}^{T}_{t, l+1} = {\rm MLP}\left({\rm FC}\left({\rm Concat}\left({\rm MCA}({F}^{T}_{t, l}), {F}^{T}_{t, l}\right)\right) \right)\\
\end{aligned}
    \label{eq:encoder}
\end{equation}
where ${F}^{T}_{t, l}$ is the input feature map of $l$-th block at the $t$-th stage; ${\rm FC}(\cdot)$ denotes the fully connected layer and ${\rm MLP}(\cdot)$ is the multi-layer perceptron.  ${\rm MCA}(\cdot)$ represents a multi-head contextual self-attention operation, computed by
\begin{equation}
\begin{aligned}
        &{\rm MCA}({F}^{T}_{t, l}) = {\rm Softmax}\left(\frac{Q_{t, l}(K_{t, l})^{Tr}-\tau \cdot M^{T}_{t, l}}{\sqrt{d_{t, l}}}\right)V_{t, l},\\
\end{aligned}
    \label{eq:encoder}
\end{equation}
where $Q_{t, l}$, $K_{t, l}$, $V_{t, l}$ are the query, key, value matrices and $d_{t, l}$ is the embedding dimension. $\tau$ (a large positive integer, set as 100 in our experiments) is utilized to adjust the attention region (\textit{more analysis can be seen in Appendix  Sec.G}).
$M^{T}_{t, l}$ is the mask of  the $l$-th block  at the $t$-th stage, formulated as
\begin{equation}
\label{eq:partconv}
\begin{split}
   M^{T}_{t, l}(i,j) = \begin{cases}
     0,  & \text{if }  \text{token } j \text{ is }\text{valid, }  \\
        1,  & \text{if }  \text{token } j \text{ is }\text{invalid, }  \\
    \end{cases}
    \end{split}
\end{equation}
where $i$ is the index of the pixel in each patch. Based on that, the refinement of the mask $M^{T}_{t, l}$ follows a rule that all tokens in a window are updated to be valid after attention as long as there is at least one valid token before; if all tokens in a window are invalid, they remain invalid after attention. Building on the above, we can obtain the feature maps of \emph{texture information} from the known areas, \emph{i.e.}, the feature maps from different stages, denoted as $F^{T}_{t}$, $t=1,...,N$.

With the acquired structure and texture feature map, one crucial question is how they collaborate during the convolutional process to alleviate their feature map loss, as discussed in the next.

\begin{figure}
\setlength{\abovecaptionskip}{0pt}
\setlength{\belowcaptionskip}{0pt}
  \centering
  \includegraphics[width=0.9\linewidth]{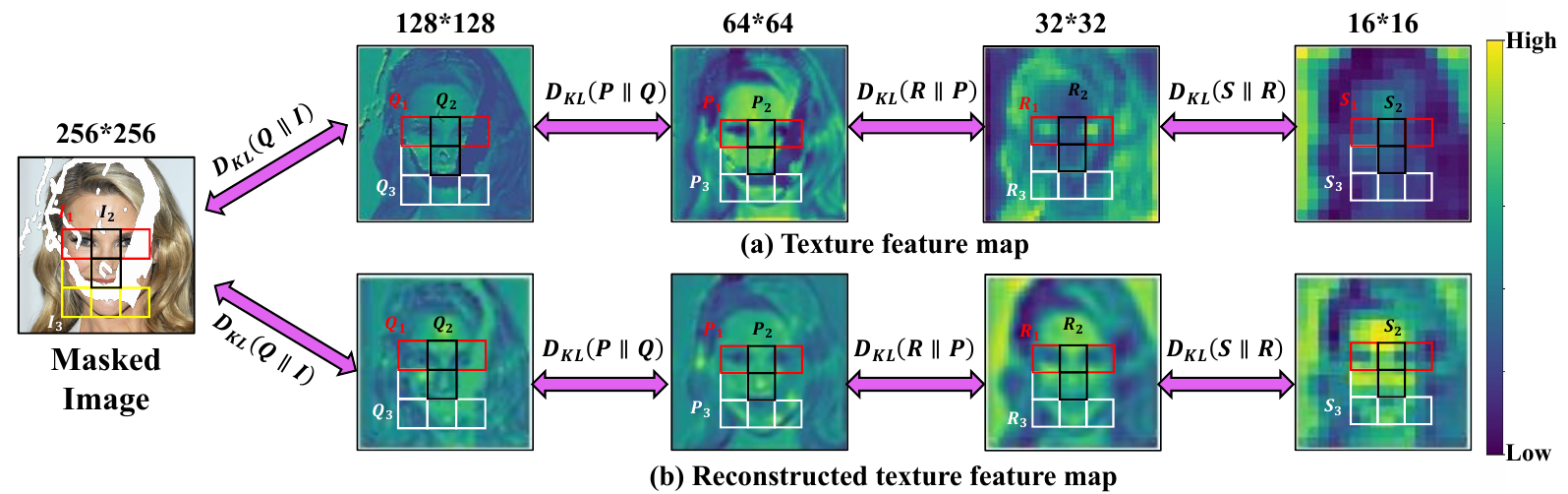}
  \caption{Comparison of the texture feature map (a) from the \textit{baseline model} and the reconstructed texture feature map (b) via the structure feature map. \textit{We highlight the semantic distribution regions in both texture feature map and reconstructed texture feature map}. The higher value within the yellow area indicates more structure or texture information from feature map. The KL divergence for (b) reconstructed texture feature map via structure feature map is smaller than that from (a) texture feature map, implying the intuition of reconstruction via structure feature map.}\label{stot_fea}

\vspace{-15pt}
\end{figure}

\subsection{What Should be Convolved for Image Inpainting?}\label{ilb}
Given structure and texture feature map as per one layer, we revisit and deploy the spatially adaptive normalization and denormalization strategy \cite{park2019semantic,liu2021pd} to incorporate the statistical information over the feature map across varied layers, formulated as
\begin{equation}\label{normalization}
\begin{aligned}
        & \tilde{F}^{T(S)}_{k} = \gamma(F_{de})\frac{{F}^{T(S)}_{k} - \mu_{k}^{T(S)}}{\sigma_{k}^{T(S)}}+\beta(F_{de})
\end{aligned}
\end{equation}

where  $F_k^{T(S)} \in \mathbb{R}^{C_{T(S)}\times H_{k} \times W_{k}}$ denotes the texture (structure) feature maps from $C_{T(S)}$ channels on the $k$-th layer of CNNs; $\tilde{F}^{T(S)}_{k}$ is the feature map of texture (structure) reconstructed via structure (texture)  {feature map} on the $k$-th layer of CNNs, and $F_{de}\in\{{F}^{T(S)}_{k}, \tilde{F}^{T(S)}_{k-1}-Upsample({F}^{T(S)}_{k})\}$ $Upsample(\cdot)$ upsamples ${F}^{T(S)}_{k}$ to match $\tilde{F}^{T(S)}_{k-1}$; in such situation, one is normalized and reconstructed by the other via their statistical mean $\mu_{k}^{T(S)}$ and variance $\sigma_{k}^{T(S)}$ over ${F}^{T(S)}_{k}$. We study two cases by alternatively reconstructing one by the other, while illustrate the reconstructed feature map over masked image across three layers during convolutional process in Table.\ref{lgh}, where we find that the texture feature map reconstruction by structure {feature map} as Table.\ref{lgh}(a) shows consistent superior performance over structure feature map reconstruction via texture feature map as Table.\ref{lgh}(b), especially for the unmasked regions, across three layers. This is because the structure feature map is always sparse so as to be easily broken down by the dense texture feature map reconstruction. Instead, such sparsity can well restore structure information loss in dense texture feature map, implying that the structure  {feature map} should guide the texture  {feature map} reconstruction rather the inverse.

\begin{figure}
\setlength{\abovecaptionskip}{0pt}
\setlength{\belowcaptionskip}{0pt}
  \centering
  \includegraphics[width=\linewidth]{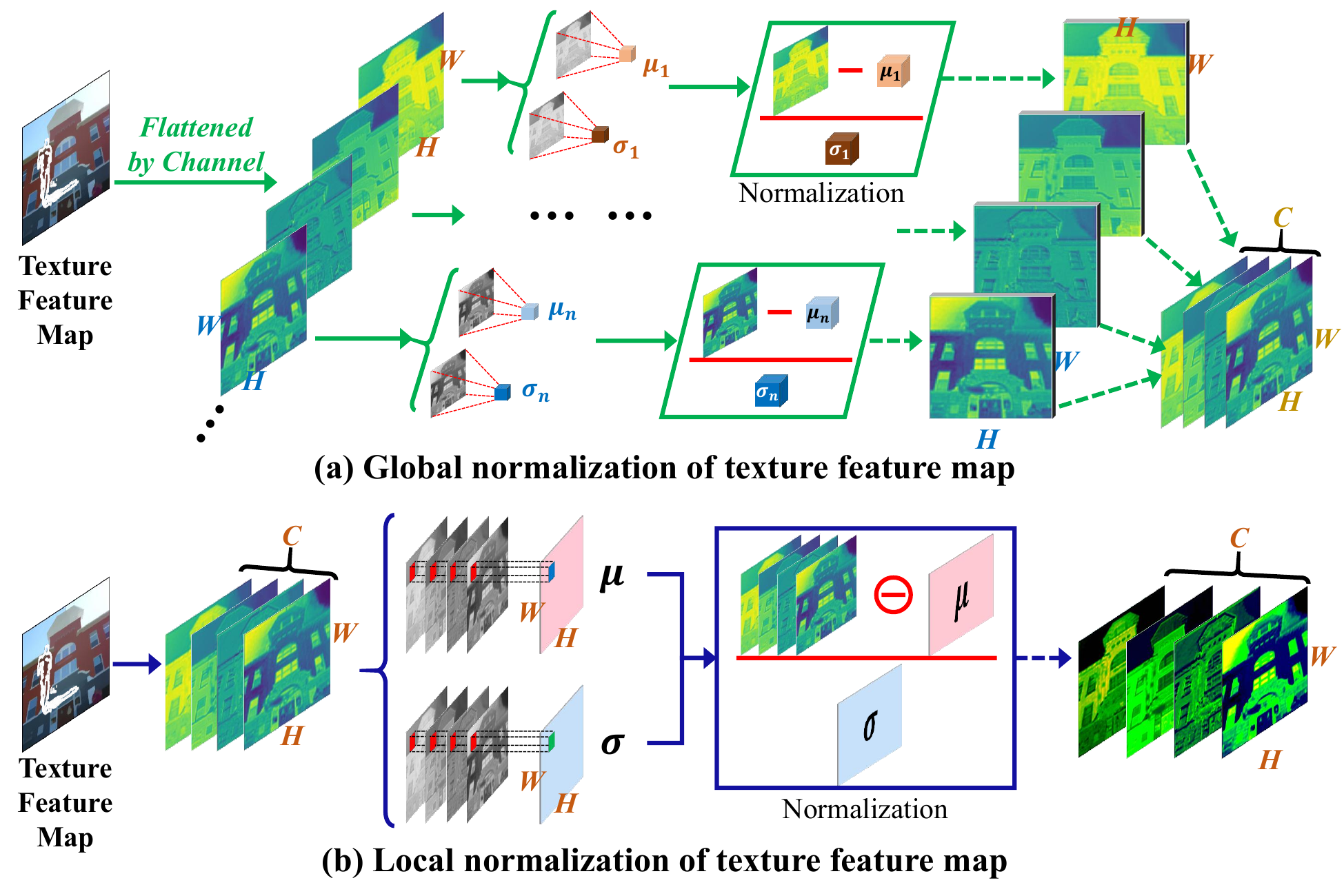}
  \caption{Illustration of how to capture the global and local texture {feature maps} via global and local normalization strategies over texture feature map, \textit{i.e.,} (a) global normalization over the texture feature map, which captures the global statistical property for the whole feature map, named the \textbf{global texture}  {feature map}. Different from that, the local normalization (b) capture the local statistical property for each position, named the \textbf{local texture} {feature map}.}
\label{global_local_texture}

\vspace{-15pt}
\end{figure}

In particular, we discuss the intuition of why the sparse structure feature map can reconstruct the dense texture feature map as shown in Fig. \ref{stot_fea}(a). We highlight different semantic regions, such as the \textit{eyes area, nose area, and the facial contour area} denoted by varied notations within different layers. Throughout the convolutional downsampling process, we can observe that, due to the lack of structure information in the texture feature map, the different semantic regions in the texture feature map become mixed with each other. The dense texture feature maps make it more likely for different semantic regions to overlap. For instance, from $128 \times 128$ to $64 \times 64$, the nose regions are covered by the face area; from $64 \times 64$ to $32 \times 32$, the facial contour area becomes unclear, making it difficult to distinguish between the neck and face areas; and from $32 \times 32$ to $16 \times 16$, the eyes area is also covered by the surrounding regions. In contrast, our reconstructed texture feature map by the structure feature map (Fig. \ref{stot_fea}(b)) preserves the structure information in the texture feature map, so that the different semantic regions remain clear even after downsampling the texture feature map to $16 \times 16$. \textit{This is because that the structure information can help bound distinct regions within the feature map}. Based on this observation, we deploy the Kullback-Leibler (KL) divergence to measure the loss of texture feature maps between adjacent layers incurred by convolutional downsampling, since the KL divergence is highly sensitive to the differences resulting from the changes in semantic region distributions, where the reconstructed texture feature map shares the similar semantics region distributions as that in the previous layer, benefiting from the reconstruction via the structure feature map to help bound the different semantic regions, with similar semantic patch regions as the last layer, while maintain a small KL divergence; instead, the texture feature map without reconstruction from structure feature map tends to be mixed together across different semantic regions during convolutional downsampling process, hence resulting into quite a different semantic distributions from the last layer, suffering from a large KL divergence.

\begin{figure}
\setlength{\abovecaptionskip}{0pt}
\setlength{\belowcaptionskip}{0pt}
  \centering
  \includegraphics[width=0.9\linewidth]{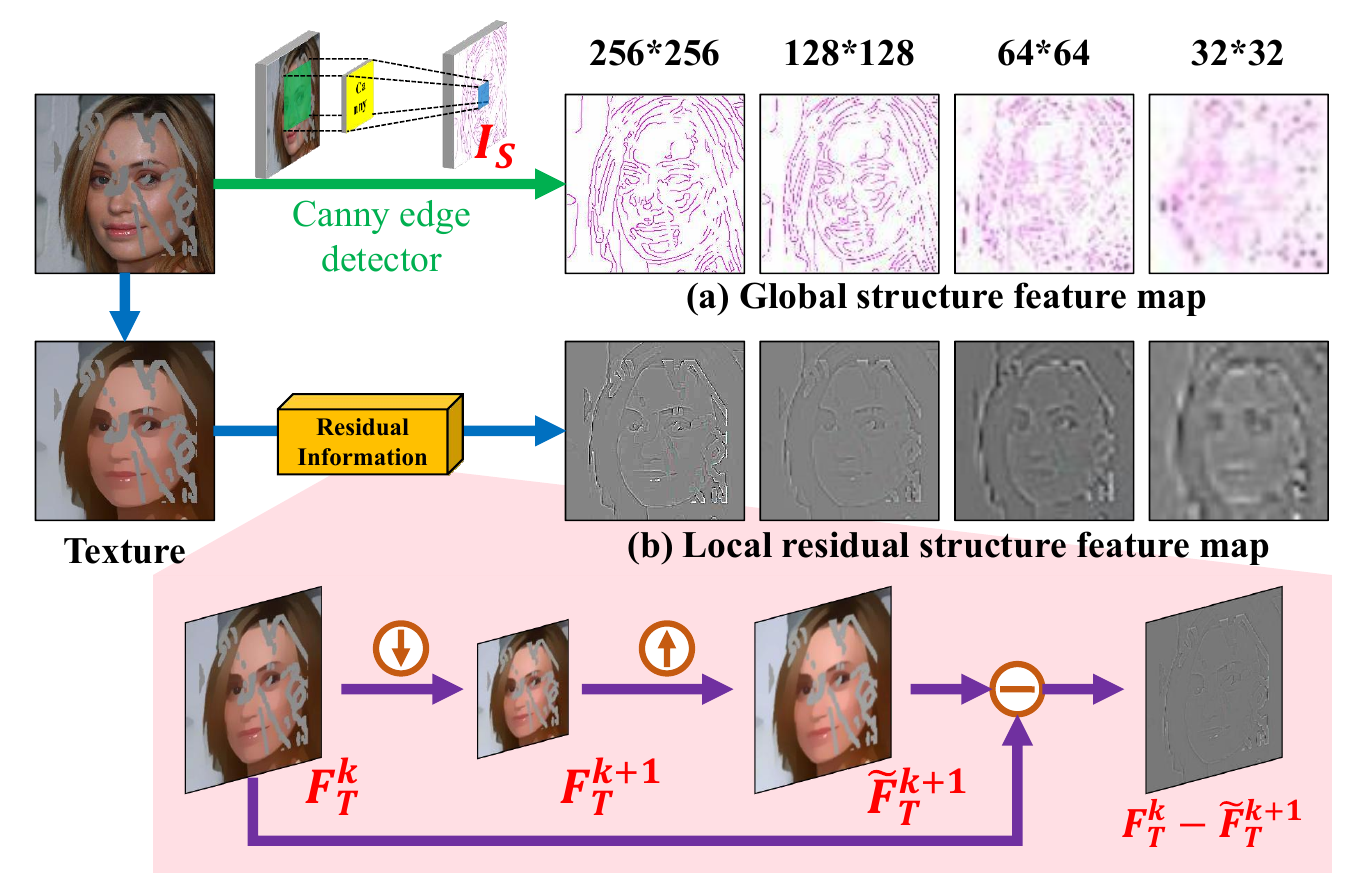}
  \caption{Illustration on how to capture the global and local residual structure feature map, respectively. (a) We employ the canny edge detector to extract the edge map as the structure information named \textit{global structure} feature map. (b) To complement  \textit{global structure} feature map, we develop the \textit{local residual structure}  feature map via residual subtraction conducts over each position, \textit{i.e.,} the texture is unsampled with the same size as previous layer, then being subtracted to yield such \textit{residual} structure feature map,  which is closely related to local texture feature map. }
\label{global_local_structure}
\vspace{-15pt}
\end{figure}
We normalize the texture feature map, then to be reconstructed by denormalization via structure {feature map}. Inspired by Liu \textit{et al.} \cite{liu2022delving} that delves globally into local texture {feature map} to capture the textural semantic correlations across the whole image for image inpainting, we perform two normalization strategies over texture feature map, \textit{i.e.,} \emph{global normalization} and \emph{local normalization}, corresponding to global and local texture {feature map}, respectively. As illustrated in Fig.\ref{global_local_texture}(a), given the texture feature map, the global normalization computes the statistical mean $\mu^{T}_{k}(c_T)$ and variance $\sigma^{T}_{k}(c_T)$ of the pixels over the whole feature map from the $c_T$-th channel as:
\begin{equation}\label{global_normalization}
\begin{aligned}
        & \mu^{T}_{k}(c_T) = \frac{1}{H_{k}W_{k}}\sum^{H_{k}}_{x=1} \sum^{W_{k}}_{y=1} F^T_k(x,y,c_{T}),\\
        & \sigma^{T}_{k}(c_T) = \sqrt{\frac{1}{H_{k}W_{k}}\sum^{H_{k}}_{x=1} \sum^{W_{k}}_{y=1} (F^T_{k}(x,y,c_{T})-\mu^{T}_{k}(c_T))^2}.
\end{aligned}
\end{equation}
Hence the global normalization can highlight the \emph{global} statistical information for the whole feature map, named \textit{global texture} feature map; as illustrated in  Fig.\ref{global_local_texture}(b), the local normalization computes the statistical mean $\mu^{T}_{k}(x,y)$ and variance $\sigma^{T}_{k}(x,y)$ of the pixels across different channels at the position $x$ and $y$ as:
\begin{equation}\label{local_normalization}
\begin{aligned}
        & \mu^{T}_{k}(x,y) = \frac{1}{C_{T}}\sum^{C_{T}}_{c_T=1}F^T_{k}(x,y, c_T),\\
        & \sigma^{T}_{k}(x,y) = \sqrt{\frac{1}{C_{T}}\sum^{C_{T}}_{c_T=1}(F^T_{k}(x,y, c_T)-\mu^{T}_{k}(x,y))^2}.
\end{aligned}
\end{equation}
Hence local normalization can highlight the \emph{local} statistical information for each position, named \textit{local texture} feature map. Following that, we further study how structure feature map reconstructs the varied texture feature map under both global and local normalization during the convolutional downsampling, as discussed in the next.

\subsubsection{How structure feature map guides the reconstruction for the texture feature map?}\label{ilb_1}
\begin{figure}
\setlength{\abovecaptionskip}{0pt}
\setlength{\belowcaptionskip}{0pt}
  \centering
  \includegraphics[width=\linewidth]{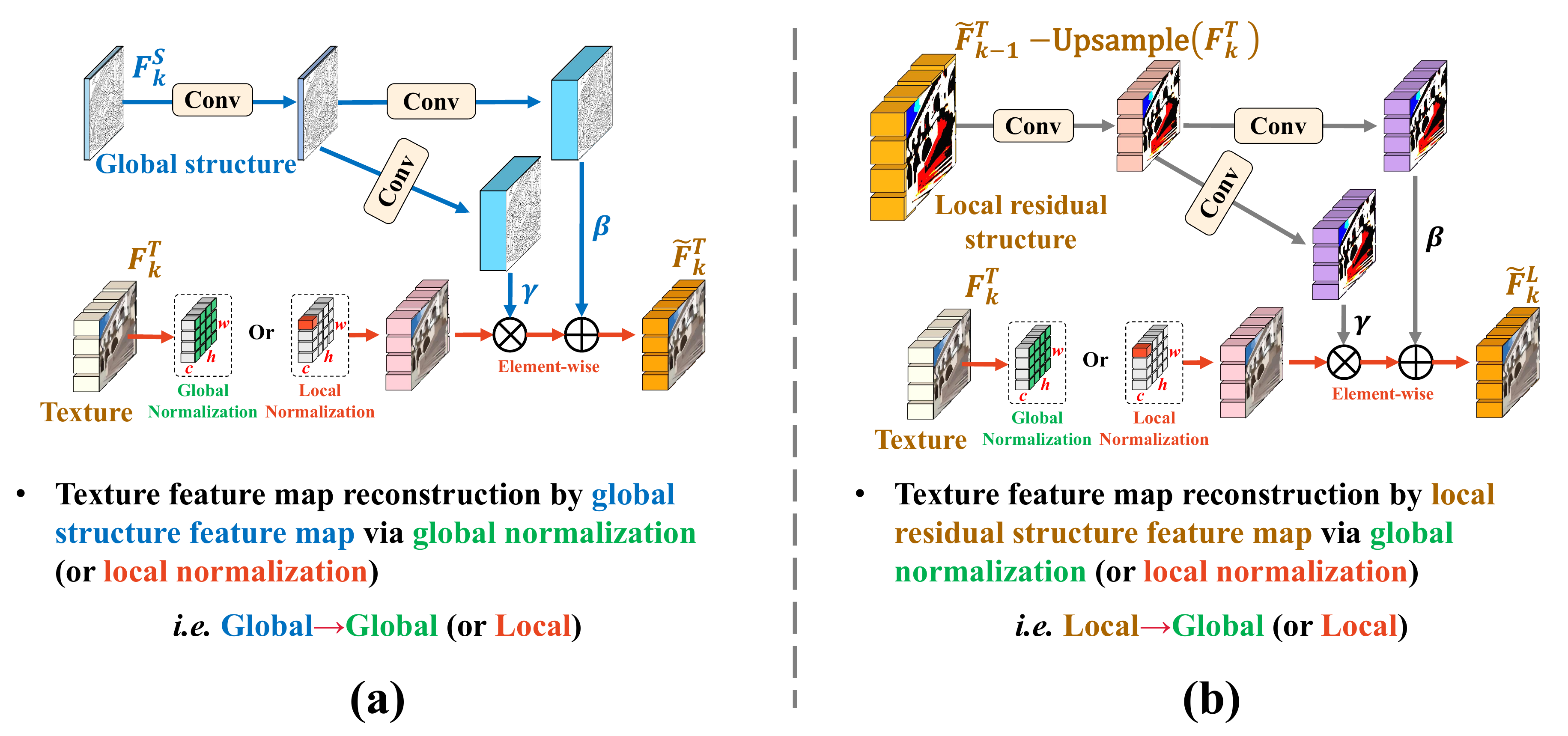}
  \caption{Illustration of how to guide the texture feature map reconstruction via the structure feature map (a)(b), four strategies are discussed, \emph{i.e.},  (a) \textbf{Global$\rightarrow$Global (Local)}: global structure  {feature map} over global (local) texture {feature map}, (b) \textbf{Local$\rightarrow$Global (Local)}: local residual structure {feature map} over global (local)  texture {feature map}.}\label{different}
\vspace{-15pt}
\end{figure}

\begin{figure}
\setlength{\abovecaptionskip}{0pt}
\setlength{\belowcaptionskip}{0pt}
  \centering
  \includegraphics[width=0.9\linewidth]{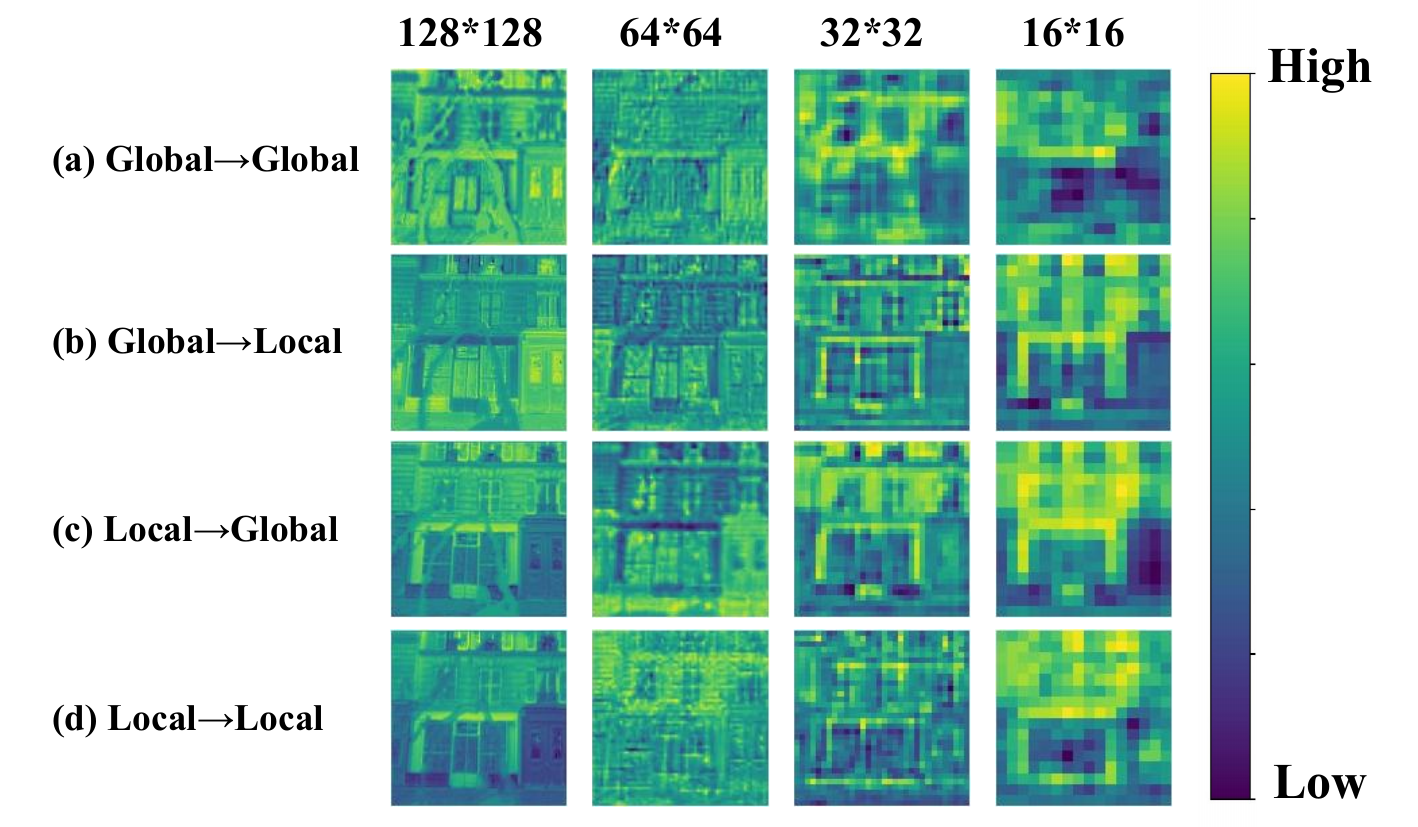}
  \caption{Comparison of the reconstructed texture {feature map}(a)(b) under the global structure {feature map} under global and local normalizations over texture feature map; (c)(d) exhibit the reconstructed texture feature map under the local residual structure {feature map} guidance. The higher value within the yellow area indicates more structure or texture information from feature map.}\label{featuremap_b}
\vspace{-15pt}
\end{figure}

Compared with texture feature map that focuses on the local details, structure feature map prefers global information, hence, as shown in Fig.\ref{global_local_structure}(a), we always call it global structure feature map. Following that, we discuss two modules as the texture feature map reconstruction by global structure {feature map} via \textit{global} and \textit{local} normalization during the convolutional downsampling process, denoted as \emph{Global$\rightarrow$Global} and \emph{Global$\rightarrow$Local}, as shown in Fig.\ref{different}(a). To choose the better matching, we offer their reconstructed texture feature maps under global and local normalization in Fig.\ref{featuremap_b}(a)(b), where the reconstructed texture feature map under local normalization shown as Fig.\ref{featuremap_b}(b) exhibits more clear local details while enjoying the similar structure information as global normalization in Fig.\ref{featuremap_b}(a).

\begin{figure}
\setlength{\abovecaptionskip}{0pt}
\setlength{\belowcaptionskip}{0pt}
  \centering
  \includegraphics[width=0.9\linewidth]{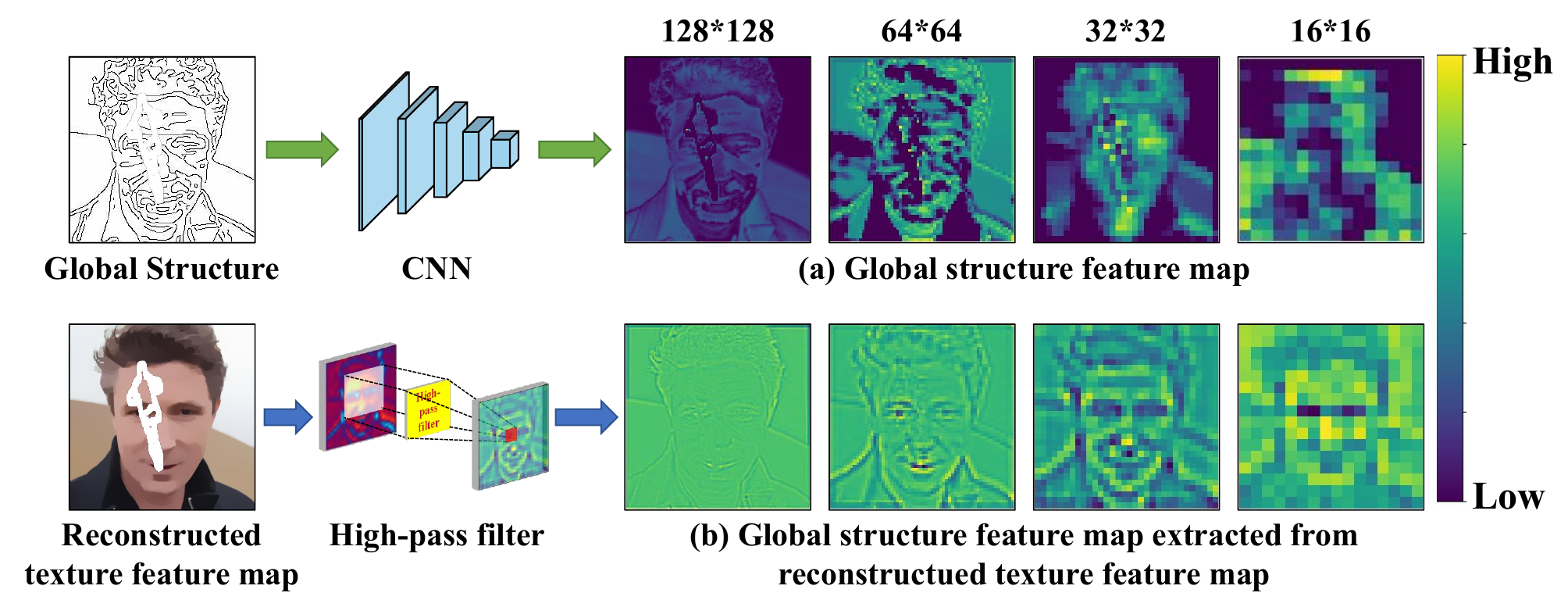}
  \caption{Illustration on  how to obtain the global structure {feature map} (a) and the global structure feature map extracted from reconstructed texture feature map (b). The higher value within the yellow area indicates more structure or texture information over feature map. (a) We run CNNs to downsample the edge map to encode the multi-scale global structure feature map. However, the KL divergence fails to compare the global structure feature map from the \textit{k}-th layer against that from the (\textit{k}-1)-th layer feature map of the baseline model with \textit{varied} sizes. Hence, we propose to extract the global structure  feature map (b) from the reconstructed texture feature maps via high-pass filter operator, which share the same channel number as the reconstructed texture feature maps.}\label{recon_global_structure}
\vspace{-15pt}
\end{figure}

\begin{figure}
\setlength{\abovecaptionskip}{0pt}
\setlength{\belowcaptionskip}{0pt}
  \centering
  \includegraphics[width=0.8\linewidth]{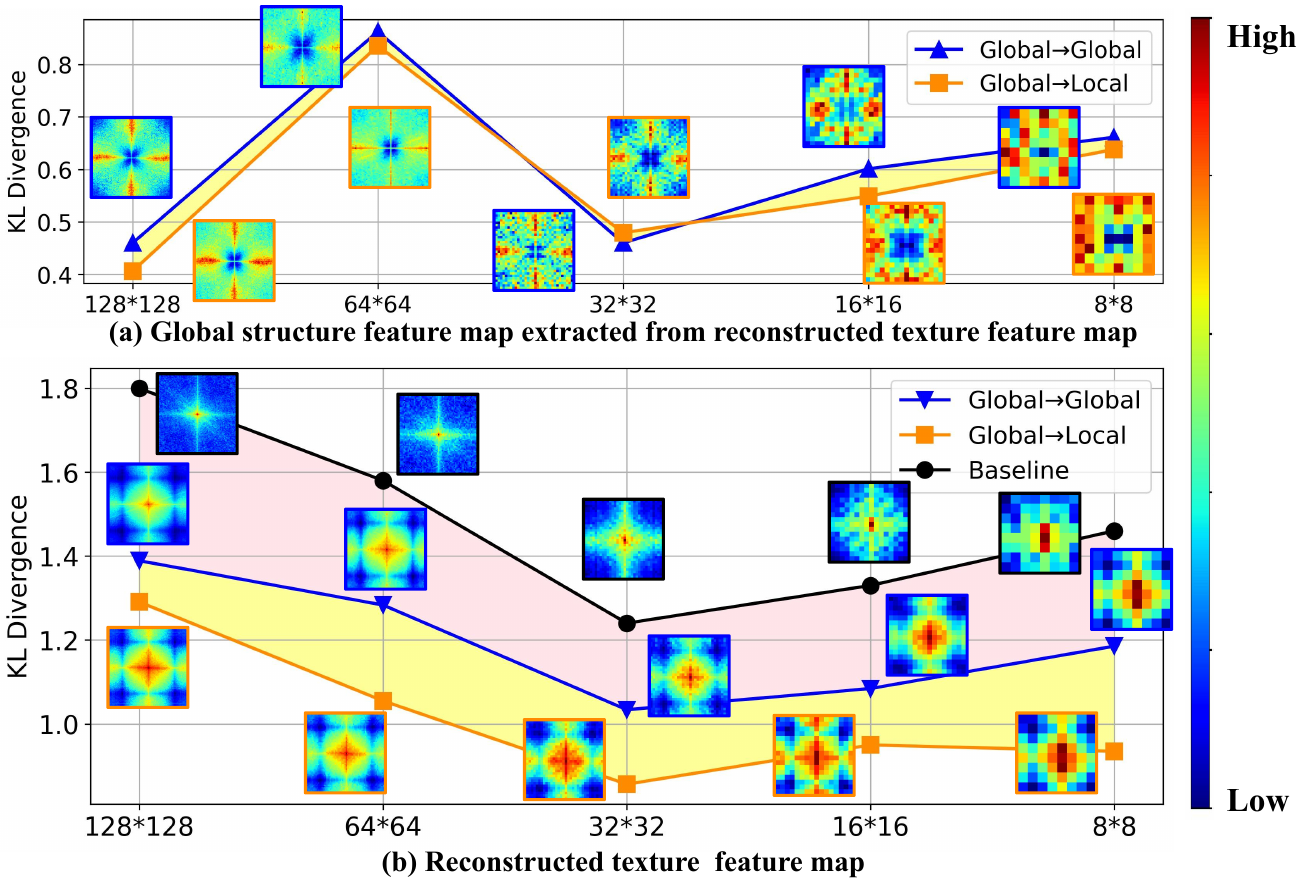}
  \caption{Comparison of global structure feature map (a) extracted from the reconstructed texture feature maps via high-pass and reconstructed texture feature map (b). The heat map denotes the spectrogram (the frequency gradually increasing from the center to the surrounding), serving as the frequency domain version of the extracted global structure and reconstructed texture feature map, particularly: the higher value for the red area represents the more structure or texture information in (a) or (b) {within the global structure or texture feature map}.}\label{spectrogram_single-1}
\vspace{-15pt}
\end{figure}

To further validate such intuition, we report the Kullback–Leibler (KL) divergence over the distributions of the feature maps w.r.t. both global structure and texture feature map, to validate the reconstruction from the global structure feature map to the texture feature map, between our modules and \emph{baseline model\footnote{We abandon the guidance from the structure {feature map} to the texture {feature map}, as the \textit{baseline model.}}}. We measure the KL divergence between the reconstructed texture feature map on the $k$-th layer of CNNs for our modules and the texture feature map on the $(k-1)$-th layer of CNNs, which enjoys the higher resolution with more texture information, for the baseline model. However, the KL divergence fails to compare the global structure feature map from the \textit{k}-th layer against that from the (\textit{k}-1)-th layer feature map of the baseline model with \textit{varied} sizes. Hence, we propose to extract the global structure feature map from the reconstructed texture feature maps via high-pass filter operator\footnote{Recalling Sec.\ref{pc}, different from the global structure feature map, we claim that the global structure feature map extracted from the reconstructed texture feature map via high-pass filter operator is only used to validate the effectiveness on how global structure {feature map} guides the texture {feature map}'s reconstruction.}, as shown in Fig.\ref{recon_global_structure}, which share the same channel number as the reconstructed texture feature maps; to this end,  we transform these multi-scale global structure feature maps extracted from the reconstructed texture feature map into the vectors with the dimension as the number of all channels for KL divergence. Instead of the global structure feature map extracted from the reconstructed texture feature map, we compare the reconstructed texture feature map with the texture feature map before downsampling of the baseline model to validate whether the texture feature map is reconstructed well. Following that, the smaller KL divergence indicates less texture feature map loss. The results indicate that the reconstructed texture feature map under local normalization exhibits much smaller KL divergence value (see Fig.\ref{spectrogram_single-1}(b)), implying less texture feature map loss, while possessing the same level of structure feature map loss as global normalization in Fig.\ref{spectrogram_single-1}(a).

\subsubsection{How to reconstruct the texture feature map under global normalization?}
\begin{figure}
\setlength{\abovecaptionskip}{0pt}
\setlength{\belowcaptionskip}{0pt}
  \centering
  \includegraphics[width=0.9\linewidth]{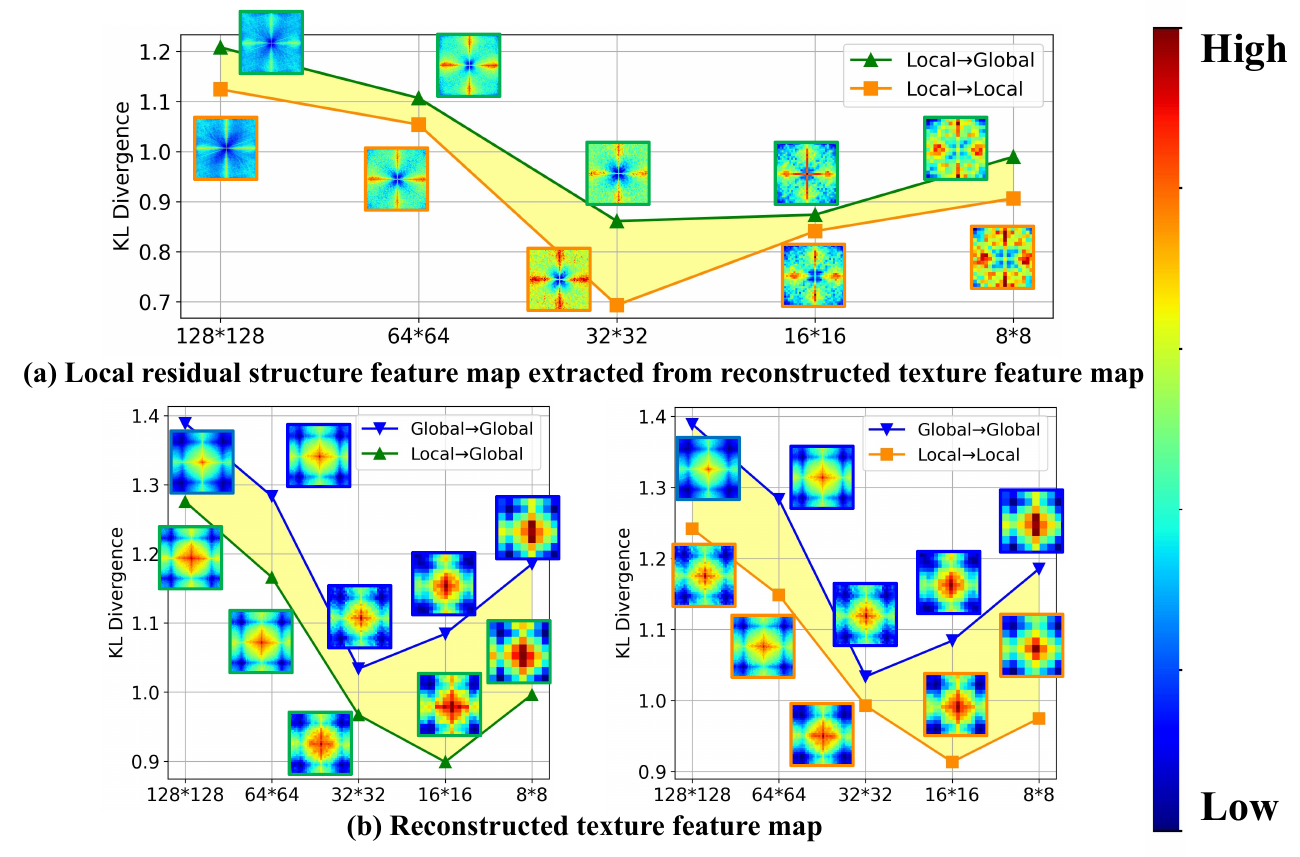}
  \caption{Comparison of local residual structure feature map (a) extracted from the reconstructed texture feature maps via subtracting the feature maps before and after downsampling and reconstructed texture feature map (b). Please refer to Fig.\ref{spectrogram_single-1} for more details about the heat map, particularly: the higher value within the red area indicates more local residual structure or texture feature map in (a) or (b).}\label{spectrogram_single-2}
\vspace{-15pt}
\end{figure}

The above global structure feature map is more adaptive to texture feature map reconstruction under local normalization than global normalization. How to match the global normalization over texture feature map remains to be answered, we need another variant high-frequency structure information while attending more on the local information. Besides, since the global structure {feature map} is too sparse to be easily broken down by the information from the adjacent regions over the lower-resolution feature map at the late stage. As shown in Fig.\ref{global_local_structure}(a), in the higher resolution (\textit{e.g.} 128$\times$128 and 64$\times$64), the edge map well encode more global structure information, yet less at the later stage (\textit{e.g.} 32$\times$32).  So the global structure feature map need to be complemented at different stages. To address such problem, as shown in Fig.\ref{global_local_structure}(b), we obtain the structure feature map from texture feature map, where the texture feature map is upsampled with the same size as the previous layer, then being subtracted (\emph{residual}) to yield such structure feature map, named \emph{local residual structure feature map}\footnote{Unlike global structure {feature map}, it is closely related to local texture information; residual subtraction conducts over each position, rather than frequency variation over regions, hence complementary to global structure {feature map}.}, to guide the texture feature map reconstruction under global normalization (see \emph{Local$\rightarrow$Global} in Fig.\ref{different}(b)), where the reconstructed texture feature map is shown in Fig.\ref{featuremap_b}(c), we can see that the reconstructed texture feature map guided by local residual structure feature map is more clear than the global one, especially over the unmasked regions, which is benefited from the intuition that the local residual structure {feature map} pays more attention on local details than global structure {feature map}. To further validate that, we study the variant combinations by reconstructing the texture  {feature map} under local normalization via local residual structure {feature map} (see \emph{Local$\rightarrow$Local} in Fig.\ref{different}(b)), where the reconstructed texture feature map is shown in the Fig.\ref{featuremap_b}(d). In particular, the texture feature map is more clear while sharing a similar structure information as Fig.\ref{featuremap_b}(a), stemming from global normalization. Such observation is consistent with the KL divergence in Fig.\ref{spectrogram_single-2}(b), \textit{i.e.,} both the modules from Fig.\ref{different}(b) indicate a similar gap upon Fig.\ref{different}(a), implying the importance of local residual structure feature map. Furthermore, we measure the KL divergence between local residual structure feature map extracted from reconstructed texture feature map, on the $k$-th layer of CNNs for our modules and the texture feature map on the $(k-1)$-th layer of CNNs for the baseline model. Following that, the smaller KL divergence indicates less local residual structure feature map loss\footnote{Since the KL divergence that fails to compare the global structure feature map with varied sizes, we turn to propose the global structure feature map extracted from reconstructed texture feature map. The multi-scale local residual structure feature maps share the same channel number as the texture feature map, so we subtract the reconstructed texture feature maps before and after downsampling as the local residual structure feature map, then transforming the local residual structure feature map  extracted from reconstructed texture feature map into the vectors with the dimension as the number of all channels for KL divergence.}. As shown in Fig.\ref{spectrogram_single-2}(a), since \textit{Local→Local} strategy is benefited from the augmented local details promoted by the guidance to local texture normalization {feature map} from local residual structure {feature map} than \textit{Local→Global} strategy. The results indicate that \textit{Local→Local} exhibits much smaller KL divergence value, implying less local residual structure feature map loss.

\subsubsection{How to simultaneously reconstruct the texture {feature map} via the global and local residual structure {feature map}?}
\begin{figure}
\setlength{\abovecaptionskip}{0pt}
\setlength{\belowcaptionskip}{0pt}
  \centering
  \includegraphics[width=0.9\linewidth]{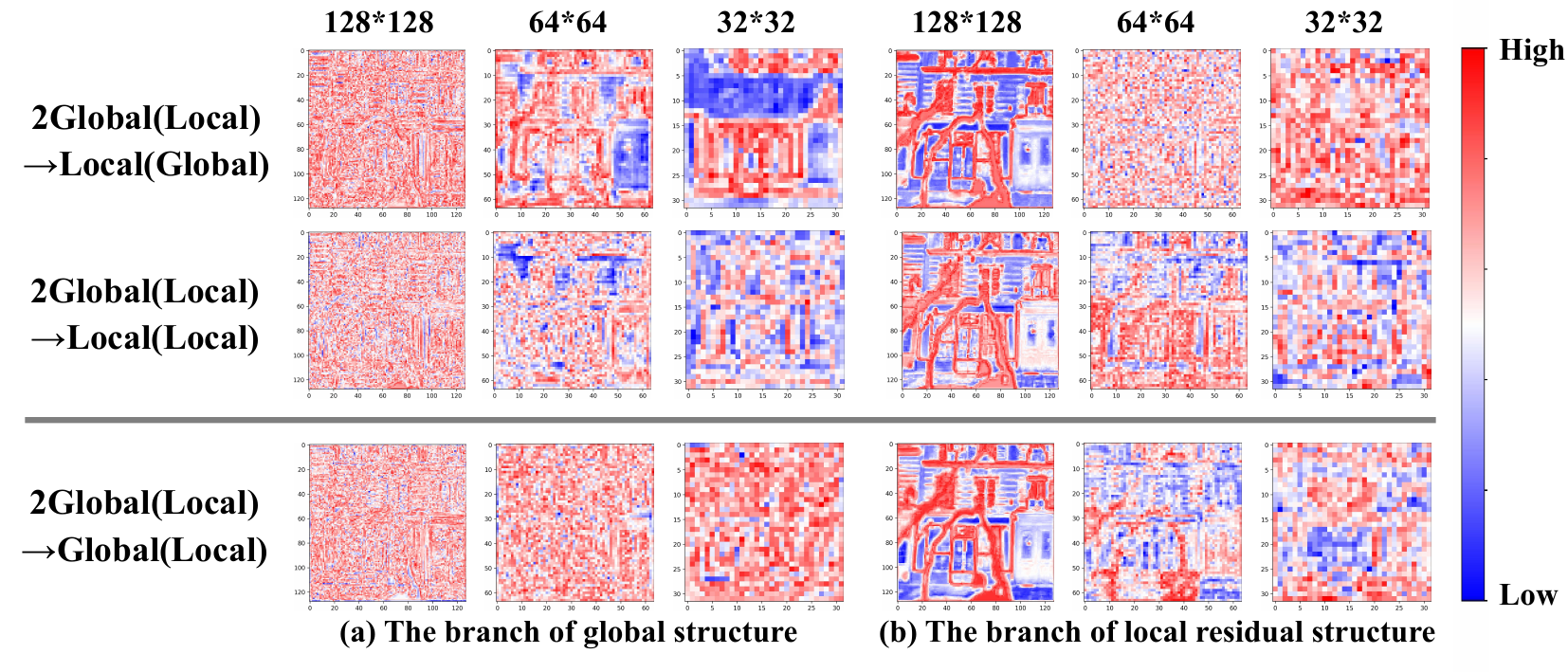}
  \caption{The information entropy (channel dimension) of the reconstructed texture feature map under the branch of global (a) and local residual(b) structure {feature map}.  In the heat map, the higher entropy value within the red area represents more texture information {over the feature maps}. }\label{entropy}

\end{figure}

To motivate our final architecture, let us first consider the cases of reconstructing texture {feature map} via local normalization by global structure {feature map}, while reconstructing the texture {feature map} via global normalization by local residual structure feature map. Fusing all these reconstructed feature maps via element-wise addition, and subsequently to be convolved for downsampling to generate the reconstructed texture feature map for the next layer. Despite the aforementioned intuition on the guidance of global (local residual) structure {feature map} over local (global) texture {feature map}, we observe that both the global and local residual structure information decreases rapidly. To be contrary, the texture feature map under both normalization strategies reduce much slowly, which is benefited from the reconstruction from both global and local residual structure {feature map}. Such intuition is validated in the 1st row of the Fig.\ref{entropy}, where, with the convolutional downsampling progresses, the global structure feature map value decreases, while popping up the local texture information by the high feature map value via local normalization, shown in Fig.\ref{entropy}(a); while local residual structure feature map is decreasing and progressively to be globally widespread caused by the reconstructed texture feature map via global normalization, shown in Fig.\ref{entropy}(b).

To further validate such intuition, we study another module on both the reconstruction from global and local residual structure {feature map} to texture {feature map} under local normalization, and illustrate the feature map value during the convolutional downsampling process in the 2nd row of Fig.\ref{entropy}. Fig.\ref{entropy}(a) shares the similar result, owing to decreasing the global structure feature maps, while Fig.\ref{entropy}(b) progressively attends more on the local details, benefited from \emph{the augmented local details promoted by the reconstruction to local texture feature map under local normalization from local residual structure feature map}.

\subsubsection{Our architecture: Reconstructing texture {feature map} via structure {feature map} under both global and local side}
\label{our_arch}
As inspired above, we finalize our strategy as: the globally normalized texture {feature map} is reconstructed via global structure {feature map}, with locally normalized texture {feature map} reconstructed via local residual structure {feature map}, their fused reconstructed feature map via element-wise addition further serves as the input of the convolution for the next layer, named \emph{2Global(Local)$\rightarrow$Global(Local)}
, where the feature maps for such strategy are illustrated in the 3rd row of the Fig.\ref{entropy}. Fig.\ref{entropy}(a) shows that the global texture information {over the feature maps} is salient for unmasked regions during convolutional downsapling process due to global normalization, while augmented by global structure {feature map} guidance for reconstruction. The similar observation holds for local texture feature maps for all layers due to local normalization, while augmented by local residual structure {feature map} guidance, as shown in Fig.\ref{entropy}(b). All these observations essentially disclose the intuition of reconstructing texture {feature map} via structure {feature map} under both global and local side, which is critical during convolutional downsampling to image inpainting.  For final strategy \emph{2Global(Local)$\rightarrow$Global(Local)},  one crucial question is why the global and local residual structure {feature map} can mutually complement each other to guide the texture {feature map}'s reconstruction, we answer that in the next subsection. (\textit{more analysis can be seen in Appendix Sec.B})

%
%

\begin{figure}
\setlength{\abovecaptionskip}{0pt}
\setlength{\belowcaptionskip}{0pt}
  \centering
  \includegraphics[width=\linewidth]{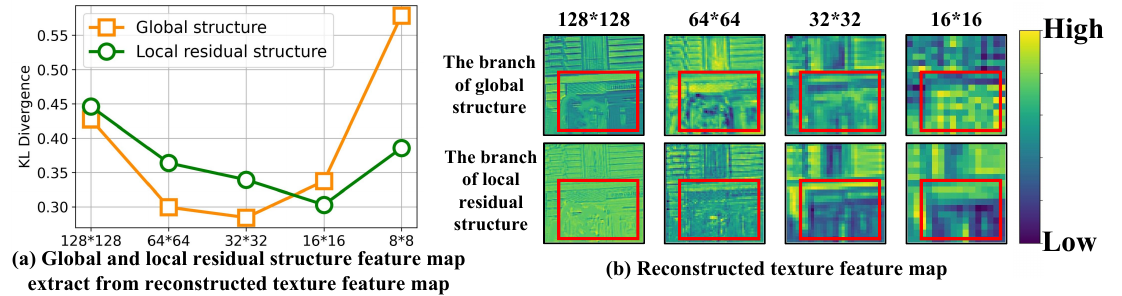}
  \caption{Illustration of the complementary relationship between global and local residual structure feature map extracted from reconstructed texture feature map. (a) KL divergence is deployed to measure the distribution difference between feature maps from current and previous stages.  (b) The global (local residual) structure {feature map} delivers clearer reconstructed texture feature map (marked by the red box) at the early (later) stage. The higher value within the yellow area indicates more texture information from feature map.}\label{assist}
\vspace{-15pt}
\end{figure}

\subsubsection{Intuition on the reciprocal support between global and local residual structure {feature map} for texture {feature map}'s reconstruction}\label{gls}
Intuitively, we extract the global structure feature map from the reconstructed texture feature map via high-pass filter and local residual structure from the reconstructed texture feature map by subtracting the  {reconstructed} texture feature maps before and after downsampling,  then calculating their KL divergence upon the baseline model while observing that the global structure feature map decreases slower than the local residual structure over the higher-resolution texture feature map during the early stages of the convolutional downsampling process, yet surpasses that at the late stages over lower-resolution feature maps, as illustrated in Fig.\ref{assist}, which implies their complementary to each other at different stages. Going one step further, we deeply investigate the followings:  \textit{On one hand}, at the early stage, global structure {feature map} exhibits a clear information, however, it is always sparse so as to be easily broken down by global structure information from the adjacent regions over the lower-resolution feature map at the late stage, such fact is illustrated in Table.\ref{lgh}(b), which visualizes the global structure feature map over masked image across three layers during convolutional process. In particular, at the early stage (\textit{e.g.} 128$\times$128 and 64$\times$64 of \textit{w/o. texture feature map}), the feature map well encodes more global structure information (\textit{i.e.}, the global structure {feature map} consistent with the masked image), yet less at the late stage (\textit{e.g.} 32$\times$32 of \textit{w/o. texture feature map}). \textit{On the other hand}, unlike global structure {feature map}, the local texture feature map reduces slowly since the texture feature map is benefited from the reconstruction from both global and local residual structure feature map, while the local residual structure {feature map} is conducted over each position of the texture {feature map} so as to alleviate the degradation, benefited from the augmented local details of texture {feature map}, \textit{e.g.,} the Fig.\ref{spectrogram_single-2}(a) validates such intuition that the augmented local details promoted by the guidance to local texture normalization {feature map} from local residual structure {feature map}, implying less local residual structure feature map loss. To this end, we conduct the guidance from global structure {feature map} \emph{twice} for balance during the late stages of convolutional downsampling process.

\begin{figure}
\setlength{\abovecaptionskip}{0pt}
\setlength{\belowcaptionskip}{0pt}
  \centering
  \includegraphics[width=\linewidth]{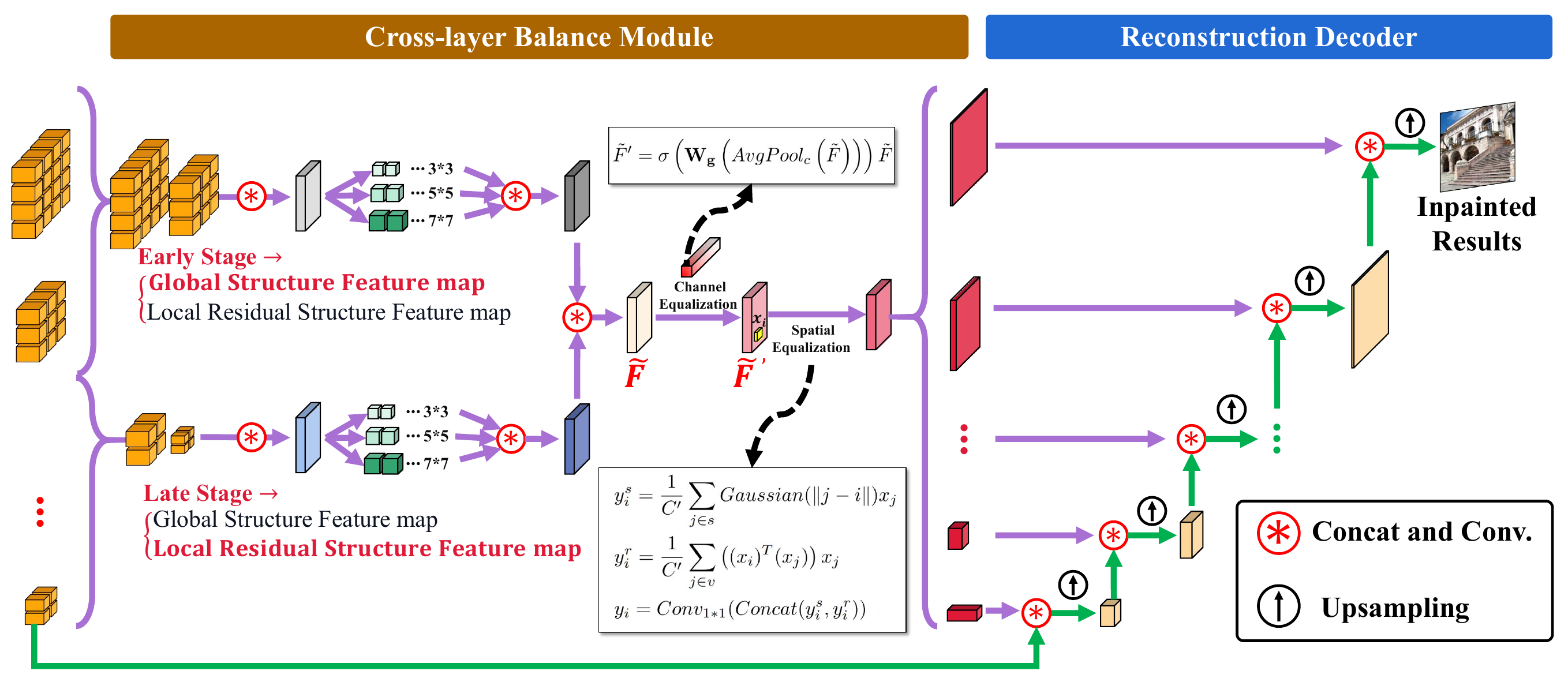}
  \caption{Illustration of the cross-layer balance module. We balance the global and local residual structure feature map to reconstruct the texture feature map under local and global normalization via a cross-layer balance module for upsampling from the decoder.}\label{architecture_cross}
\vspace{-15pt}
\end{figure}

\subsubsection{Cross-layer balance module decoder for upsampling.}\label{cross}
Based on the above, the reconstructed texture feature map over different layers regarding different scales within the encoder, served as the input to   our proposed cross-layer balance module, in particular:  1) reconstructing texture feature maps dominated by global structure feature maps at the early stage; 2) reconstructing texture feature maps dominated by local residual structure feature maps at the late stage; 3) reconstructing texture feature map regarding the last layer of CNNs at the last stage, where each stream contains five partial convolutions with the same kernel size differed among different streams. Afterwards, the reconstructed texture feature maps from three streams are concatenated to obtain the output feature map with the same size. Finally, we fuse the feature maps from the local residual and global structure streams, denoted as $\tilde{F}$, for the upsampling from decoder, then we adopt the feature equalization method \cite{Liu2019MEDFE} to balance the local residual and global structure feature maps from different stages of CNNs for decoder (see Fig.\ref{architecture_cross}). For the channel domain, the feature map $\tilde{F}$ is equalized by
\begin{align}
        & \tilde{F}' = \sigma\left(\mathbf{W_{g}}\left(AvgPool_{c}\left(\tilde{F}\right)\right)\right)\tilde{F}, \label{eq:glo}
\end{align}
where $AvgPool_{c}(\cdot)$ denotes the global average pooling in each channel and $\mathbf{W_{g}}$ is the learnable linear projection matrix; $\sigma(\cdot)$ is the sigmoid function.
Following that, in the spatial domain, assume $x_i$ to be the feature vector at the position $i$ of $\tilde{F}'$ and $x_j$ to be a neighboring feature channel around $i$ at position $j$, the equalization is achieved by
\begin{small}
\begin{align}
        & y_{i}^{s} = \frac{1}{C'} \sum_{j \in s}Gaussian(\|j-i\|)x_{j}, \label{eq:gau}\\
        & y_{i}^{r} = \frac{1}{C'} \sum_{j \in v}\left((x_{i})^{Tr}(x_{j})\right)x_{j}, \label{eq:dis}\\
        & y_{i} = Conv_{1*1}(Concat(y_{i}^{s}, y_{i}^{r})), \label{eq:con}
\end{align}
\end{small}
where $y_{i}^{s}$ and $y_{i}^{r}$ are the feature vectors after spatial and range similarity measurements. $Gaussian(\cdot)$ represents a Gaussian function and $C'$ is a normalization factor. $s$ and $v$ are the corresponding neighboring regions.
With Eq.(\ref{eq:glo}-\ref{eq:con}), $\tilde{F}$ is corrected to yield the consistent local residual and global structure feature maps.

We process the texture feature maps for the cross-layer balance module to be with the same size as the input reconstructed texture feature map set,
so as to be exploited for upsampling via bottom-up strategy to yield the final image inpainting. Specifically, we process the output of the encoder's final layer as input for the decoder, which are further concatenated with the texture feature maps from the reconstructed texture feature map set channel-wise. The upsampling is conducted via the fusion through convolution. This iterative process continues until the desired resolution is reached. The whole process is illustrated in Fig.\ref{architecture_cross}.

\begin{table*}[t]
\setlength{\abovecaptionskip}{0pt}
\setlength{\belowcaptionskip}{0pt}
    \centering
    \tiny
    \caption{Comparison of quantitative results (\emph{i.e.}, PSNR, SSIM and FID) under varied mask ratios on PSV, CelebA and Places2 with irregular mask dataset. The best results are reported with \textbf{boldface}. $\uparrow$: Higher is better; $\downarrow$: Lower is better; - : no results are reported.}
    \label{tab:compare1}
    \resizebox{0.7\linewidth}{!}{
    \begin{tabular}{@{}|l|c||c|c|c|c||c|c|c|c||c|c|c|c|@{}}
    \hline
	\multicolumn{2}{|c||}{\textbf{Metrics}} & \multicolumn{4}{c||}{\textbf{PSNR$\uparrow$}}  &\multicolumn{4}{c||}{\textbf{SSIM$\uparrow$}}& \multicolumn{4}{c|}{ \textbf{FID$\downarrow$} }\\\hline
     \textbf{Method} &\textbf{Venue} &\textbf{10-20\%} & \textbf{20-30\%} &\textbf{30-40\%} &\textbf{40-50\%} &\textbf{10-20\%} & \textbf{20-30\%} &\textbf{30-40\%} &\textbf{40-50\%} &\textbf{10-20\%} & \textbf{20-30\%} &\textbf{30-40\%} &\textbf{40-50\%} \\\hline
    \hline
    \multicolumn{14}{|l|}{\cellcolor{blue!20}\textbf{CelebA ($256 \times 256$)}} \\\hline
    \textit{PENNet} \cite{yan2019PENnet} &CVPR' 19 &30.54 &27.78 &26.01 &23.15 &0.967 &0.934 &0.894 &0.845 &3.14 &6.37 &9.33 &10.74\\
     \textit{MEDFE} \cite{Liu2019MEDFE} &ECCV' 20 &31.34 &28.09 &25.71 &23.85 &0.976 &0.953 &0.923 &0.889 &2.81 &5.68 &8.99 &10.08\\
    \textit{RFR} \cite{li2020recurrent} &CVPR' 20 &30.93 &28.94 &27.11 &25.47 &0.970 &0.958 &0.939 &0.913 &3.17 &4.01 &4.89 &6.11\\
    \textit{HiFill} \cite{yi2020contextual} &CVPR' 20 &30.80 &27.92 &25.42 &23.61 &0.974 &0.950 &0.921 &0.885 &3.27 &6.07 &9.17  &10.38\\
    \textit{CTSDG} \cite{Guo_2021_ICCV} &ICCV' 21 &32.91 &29.51 &27.02 &25.13 &0.981 &0.962 &0.937 &0.905 &2.08 &3.86 &6.06 &8.58\\
    \textit{MMT} \cite{yu2022unbiased}  &ECCV' 22 &29.94 &- &26.88 &- &0.970 &- &{0.950} &- &6.47 &- &9.32&- \\
    \textit{HAN} \cite{deng2022hourglass} &ECCV' 22 &33.04 &{29.94} &27.53 &25.62 &{0.983} &{0.967} &0.945 &{0.918} &{1.49} &{2.58} &{3.93} &{5.39} \\
    \textit{DGTS} \cite{liu2022delving} &MM' 22 &33.09  &30.23 &28.01  &25.92 &0.984 &0.971 &0.957 &0.937 &1.42 &2.43 &4.27 &5.53\\
    \textit{IR-SDE} \cite{luo2023image} &ICML' 23 &33.19 &30.74 &28.83 &25.97 &0.992 &0.980 &0.967 &0.951 &1.21 &2.30 &4.13 &5.38\\
    \hline
   \cellcolor{green!20}\textbf{\textit{Ours(PENNet)}} &\cellcolor{green!20}- &\cellcolor{green!20}{33.23} &\cellcolor{green!20}{30.34} &\cellcolor{green!20}{29.01} &\cellcolor{green!20}{25.86} &\cellcolor{green!20}{0.991} &\cellcolor{green!20}{0.978} &\cellcolor{green!20}{0.968} &\cellcolor{green!20}{0.952}&\cellcolor{green!20}{1.42}&\cellcolor{green!20}{2.55} &\cellcolor{green!20}{3.89}&\cellcolor{green!20}{5.47}\\
    \cellcolor{green!20}\textbf{\textit{Ours(HiFill)}} &\cellcolor{green!20}-   &\cellcolor{green!20}{33.39} &\cellcolor{green!20}{30.66} &\cellcolor{green!20}{29.47} &\cellcolor{green!20}{26.04} &\cellcolor{green!20}\textbf{0.994} &\cellcolor{green!20}{0.981} &\cellcolor{green!20}{0.973} &\cellcolor{green!20}\textbf{0.963} &\cellcolor{green!20}{1.25} &\cellcolor{green!20}{2.32} &\cellcolor{green!20}{3.77} &\cellcolor{green!20}{5.34}\\
    \cellcolor{green!20}\textbf{\textit{Ours}} &\cellcolor{green!20}-  &\cellcolor{green!20}\textbf{33.45} &\cellcolor{green!20}\textbf{30.92}  &\cellcolor{green!20}\textbf{29.76} &\cellcolor{green!20}\textbf{26.13} &\cellcolor{green!20}{0.993}  &\cellcolor{green!20}\textbf{0.984} &\cellcolor{green!20}\textbf{0.979}  &\cellcolor{green!20}\textbf{0.963}  &\cellcolor{green!20}\textbf{1.17}  &\cellcolor{green!20}\textbf{1.98} &\cellcolor{green!20}\textbf{3.24}  &\cellcolor{green!20}\textbf{4.35} \\
    \hline
    \hline
    \multicolumn{14}{|l|}{\cellcolor{blue!20}\textbf{Places2 ($256 \times 256$)}} \\\hline
    \textit{PENNet} \cite{yan2019PENnet} &CVPR' 19 &26.78&23.53 &21.80&20.48  &0.946 &0.894 &0.841 &0.781 &20.40 &37.22 &53.74 &67.08\\
    \textit{MEDFE} \cite{Liu2019MEDFE} &ECCV' 20 &28.13 &25.02 &22.98 &21.53 &0.958 &0.919 &0.874 &0.825 &14.41 &27.52 &38.45 &53.05\\
    \textit{RFR} \cite{li2020recurrent} &CVPR' 20 &27.26 &24.83 &22.75 &21.11 &0.929 &0.891 &0.830 &0.756 &17.88 &22.94 &30.68 &38.69 \\
    \textit{HiFill} \cite{yi2020contextual} &CVPR' 20 &27.50 &24.64 &22.81 &21.23 &0.954 &0.914 &0.873 &0.816 &16.95 &29.30 &42.88 &57.12\\
    \textit{CTSDG} \cite{Guo_2021_ICCV} &ICCV' 21  &28.91 &25.36 &22.94 &21.21 &0.952 &0.901 &0.834 &0.755 &15.72 &27.88 &42.44 &57.78 \\
    \textit{ZITS} \cite{dong2022incremental} &CVPR' 22 &28.31 &25.40 &23.51 &22.11 &0.942 &0.902 &0.860 &0.817 &- &- &- &- \\
    \textit{FAR} \cite{cao2022learning} &ECCV' 22 &28.36 &{25.48} &23.60 &22.18 &0.942 &0.903 &{0.861} &{0.818} &- &- &- &- \\
    \textit{HAN} \cite{deng2022hourglass} &ECCV' 22 &28.93 &25.44 &23.06 &21.38 &{0.957} &{0.903} &0.839 &0.762 &12.01 &{20.15} &{28.85} &{37.63} \\
    \textit{DGTS} \cite{liu2022delving} &MM' 22 &29.11&27.19&24.99&22.60&0.969&0.945&0.909&0.859&10.17&19.69&28.49&37.90\\
    \textit{IR-SDE} \cite{luo2023image} &ICML' 23 &29.58 &27.23 &25.05 &22.46 &0.973 &0.941 &0.908 &0.865 &9.55 &19.04 &28.22 &38.38\\
    \hline
   \cellcolor{green!20}\textbf{\textit{Ours(PENNet)}} &\cellcolor{green!20}- &\cellcolor{green!20}{29.76} &\cellcolor{green!20}\textbf{27.48} &\cellcolor{green!20}{25.15} &\cellcolor{green!20}{22.62} &\cellcolor{green!20}{0.961} &\cellcolor{green!20}{0.937} &\cellcolor{green!20}{0.908} &\cellcolor{green!20}{0.862} &\cellcolor{green!20}{9.75} &\cellcolor{green!20}{19.22} &\cellcolor{green!20}{26.90} &\cellcolor{green!20}{36.98} \\
    \cellcolor{green!20}\textbf{\textit{Ours(HiFill)}} &\cellcolor{green!20}-  &\cellcolor{green!20}{29.97}  &\cellcolor{green!20}{27.26}  &\cellcolor{green!20}{25.47}  &\cellcolor{green!20}{22.57}  &\cellcolor{green!20}{0.978}  &\cellcolor{green!20}{0.944}  &\cellcolor{green!20}{0.912}  &\cellcolor{green!20}{0.873}  &\cellcolor{green!20}{9.41}  &\cellcolor{green!20}{18.24}  &\cellcolor{green!20}{24.32} &\cellcolor{green!20}{35.95}  \\
    \cellcolor{green!20}\textbf{\textit{Ours}} \cellcolor{green!20}&\cellcolor{green!20}-  &\cellcolor{green!20}\textbf{30.10} &\cellcolor{green!20}{27.30}  &\cellcolor{green!20}\textbf{25.49} &\cellcolor{green!20}\textbf{22.74} &\cellcolor{green!20}\textbf{0.982}  &\cellcolor{green!20}\textbf{0.955} &\cellcolor{green!20}\textbf{0.923}  &\cellcolor{green!20}\textbf{0.887}  &\cellcolor{green!20}\textbf{8.40}  &\cellcolor{green!20}\textbf{16.98} &\cellcolor{green!20}\textbf{22.88}  &\cellcolor{green!20}\textbf{34.80} \\
    \hline
    \end{tabular}
  \label{table:compare1}
  }
\vspace{-15pt}
\end{table*}


\subsection{Loss Functions}
\label{loss_fun}
To this end,  we introduce several loss functions during the training, including: reconstruction loss to improve the image quality, auxiliary loss to enhance the structure information, apart from adversarial loss to ensure contents consistency.

\noindent \textbf{Reconstruction Loss.} We employ the reconstruction loss \cite{yan2019PENnet} to measure the pixel-wise difference between inpainting result $I_{out}$ and the ground truth $I_{gt}$, expressed as
\begin{equation}
\begin{aligned}
        & \mathcal{L}_{rec} = \|I_{out} - I_{gt}\|_{1},
\end{aligned}
    \label{eq:bpa}
\end{equation}
where $||\cdot||_1$ is the $\ell_1$ norm.

\noindent \textbf{Auxiliary Loss.}
To reconstruct  the texture via the structure, we devise the auxiliary loss function to promote the extraction of the structure information. Specifically, we introduce the decoder $\mathcal{D}_{S}$ to inpaint the edge map $\tilde{I}_{S}$. Based on that, we measure the pixel-wise difference between $\mathcal{D}_{S}(F_{N}^{S})$  and the unmasked image ${I}_{H}$, formulated as
\begin{equation}
\begin{aligned}
        & \mathcal{L}_{aux}^{str} = \|\mathcal{D}_{S}(F_{N}^{S}) - I_{S}\|_{1},\\
\end{aligned}
    \label{eq:bpa}
\end{equation}

\noindent \textbf{Adversarial Loss.} To enhance perception of images quality, we utilize the adversarial loss  to distinguish the fake inpainted image $I_{out}$  from the ground truth by a discriminator $\mathcal{D}$, which is expressed as
\begin{equation}
\begin{aligned}
        & \mathcal{L}_{adv}^G = -\mathbb{E}_{I_{out}}[log(\mathcal{D}(I_{out}))],\\
        & \mathcal{L}_{adv}^D = -\mathbb{E}_{I_{gt}}[log(\mathcal{D}(I_{gt}))] -\mathbb{E}_{I_{out}}[log(1-\mathcal{D}(I_{out}))].
\end{aligned}
    \label{eq:bpa}
\end{equation}

\noindent \textbf{Overall Loss. }In summary, we finalize the overall loss
as
\begin{equation}
\begin{aligned}
        & \mathcal{L}_{Conv} = \lambda_{r}\mathcal{L}_{rec} + \lambda_{adv}\mathcal{L}_{adv}^G + \lambda_{r}^{str}\mathcal{L}_{aux}^{str},
\end{aligned}
    \label{eq:all}
\end{equation}
where $\lambda_{r}$, $\lambda_{adv}$ and $\lambda_{r}^{str}$  are the balance hyper-parameters, which are empirically set as 1, 0.1 and 1 in our experiments.
Eq.(\ref{eq:all}) is minimized to train the pipeline.

\begin{figure}
\setlength{\abovecaptionskip}{0pt}
\setlength{\belowcaptionskip}{0pt}
  \centering
  \includegraphics[width=\linewidth]{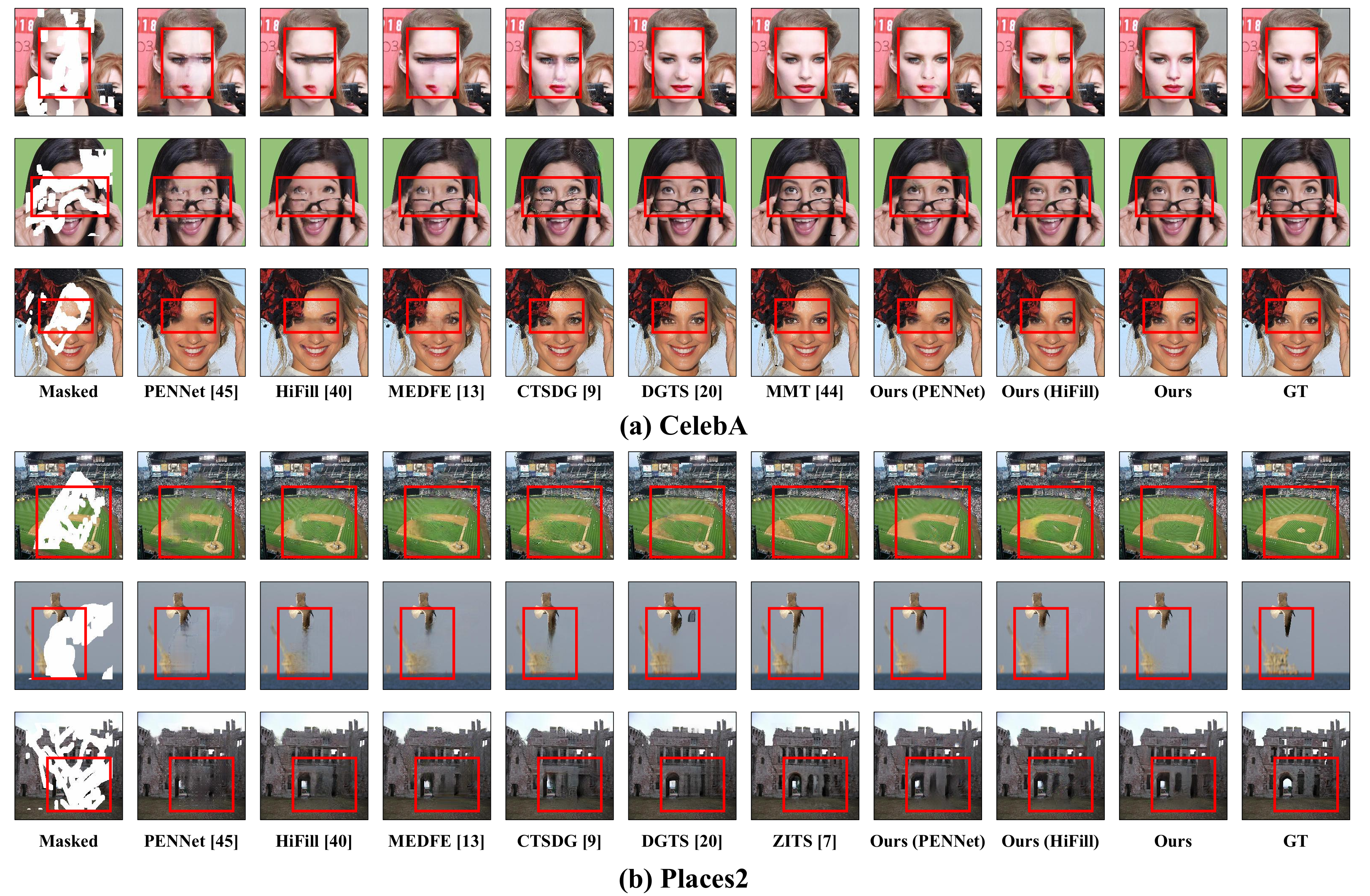}
  \caption{Comparison of the final inpainted results with the state-of-the-arts on CelebA (a) and Places2 (b) with irregular masks. Our method delivers the desirable inpainting results (marked as the red box) against others. }\label{visual_analysis1}
\vspace{-15pt}
\end{figure}

\section{Experiment}
\subsection{Implementation Details}
We validate the proposed method over three typical datasets with different characteristics, including: \textbf{Paris StreetView (PSV)} is a collection of street view images in Paris, consisting of 14,900 images for training and 100 images for validation; \textbf{CelebA} is the human face dataset with 30,000 aligned face images, which is split into 28,000 training images and 2,000 validation images; \textbf{Places2} contains more than 1.8M natural images from varied scenes.
The partial convolutional head, composed of two partial convolutional blocks, is responsible for downsampling. The 4-layer transformer body features block numbers and window sizes of \{2, 2, 6, 2\} and \{8, 8, 8, 8\}, respectively. The 5-layer structure encoder has channel numbers set to \{64, 128, 256, 512, 1024\}. Specifically, our framework extracts multi-scale features in 6 layers; the previous 5 layers downsample the feature maps, while the final layer concatenates all previous stages and processes the resulting feature map within a convolutional layer. In the cross-layer balance module, the layer number within both the early and late stages are set to 3, and skip connections concatenate the feature maps from the encoder with the corresponding ones in the decoder. We adopt the Adam optimizer with $\beta_1$ = 0.5 and $\beta_2$ = 0.99, where the learning rates for generator and discriminator are set to $5\times10^{-4}$ and $10^{-4}$, respectively. We train the model for the total of 150 epochs on PSV and CelebA, while it is 20 epochs on Places2. All the experiments are implemented with the pytorch framework and run on 4 NVIDIA 2080TI GPU (\textit{more analysis can be seen in Appendix  Sec.C}).

\subsection{Comparison with State-of-the-arts}
To evaluate our method and state-of-the-arts, we deploy three evaluation metrics: 1) peak signal-to-noise ratio (PSNR); 2) structural similarity index (SSIM) \cite{wang2004image}; and 3) Fréchet Inception Score (FID) \cite{heusel2017gans}. In particular, $\ell_1$ error is defeated since it prefers blurry inpainted image. PSNR and SSIM are used to compare the low-level differences over pixel level between the generated image and ground truth. FID evaluates the perceptual quality by measuring the feature distribution distance between the synthesized and real images.The irregular masked regions of the image are testified with varied ratios over the whole image size.


\begin{table}[t]
\setlength{\abovecaptionskip}{0pt}
\setlength{\belowcaptionskip}{0pt}
		\begin{center}
		\caption{Comparison of quantitative results (\emph{i.e.}, FID and LPIPS) under varied mask ratios on Places2 ($512 \times 512$). We also provide the FLoating-point OPerations (FLOPs) and parameters (Params). M and G denote $10^{6}$ and $10^{9}$,  The best results are reported with \textbf{boldface}. $\downarrow$: Lower is better.}\label{tab:compare512}
			\resizebox{\linewidth}{!}{
				\begin{tabular}{|l|c|| c|c || c |c || c| c ||  c|c || }
					\hline
					\multirow{3}{*}{\textbf{Method}} &\multirow{3}{*}{\textbf{Venue}} & \multirow{3}{*}{\textbf{Params}} & \multirow{3}{*}{\textbf{FLOPs}} & \multicolumn{6}{c||}{\cellcolor{blue!20}{\textbf{Places} ($512 \times 512$)}}  \\
					~ & ~ & ~& ~ & \multicolumn{2}{c||}{\textbf{Narrow Mask}} & \multicolumn{2}{c||}{\textbf{Medium Mask}} &\multicolumn{2}{c||}{\textbf{Wide Mask}}\\
					~ & ~ & ~ & ~ & \textbf{FID}$\downarrow$ & \textbf{LPIPS}$\downarrow$  & \textbf{FID}$\downarrow$ & \textbf{LPIPS}$\downarrow$  & \textbf{FID}$\downarrow$ & \textbf{LPIPS}$\downarrow$  \\
					\hline
                    EdgeConnect~\cite{Nazeri_2019_ICCV} &ICCV' 19 &22.0M &490.4G  &1.33 &0.111 &3.66 &0.135 &8.37 &0.160 \\
                    DeepFill v2~\cite{yu2018generative}  &CVPR'19 &4.00M  &100.8G  &1.06 &0.104 &2.68 &0.130 &5.20 &0.155 \\
                    RegionNorm~\cite{yu2020region} &AAAI' 20 &12.0M &216.0G &2.13 &0.120 &6.27 &0.146 &15.7 &0.176 \\
                    HiFill~\cite{yi2020contextual} &CVPR' 20 &3.00M   &72.10G  &9.24 &0.218 &7.54 &0.157 &12.8 &0.180 \\
                    CoModGAN~\cite{zhao2021large} &ICLR' 20 &109M  &121.09G  &0.82 &0.111 &1.34 &0.128 &1.82 &0.147 \\
                    MADF~\cite{zhu2021image}  &TIP' 21 &85.0M   &183.7G &0.57 &0.085 &1.62 &0.113 &3.76 &0.139\\
                    RegionWise~\cite{ma2022regionwise}  &TCYB' 22 &47.0M  &35.00G  &0.90 &0.102 &2.42 &0.125 &4.75 &0.149  \\
                    AOT GAN~\cite{zeng2022aggregated}  &TVCG' 22 &15.0M  &291.5G  &0.79 &0.091 &2.29 &0.119 &5.94 &0.149 \\
                    LaMa~\cite{suvorov2022resolution} &WACV' 22 &27.0M    &171.3G  &0.63 &0.090 &1.30 &0.112 &2.21 &0.135 \\
                    Big-LaMa~\cite{suvorov2022resolution} &WACV' 22 &51.0M    &264.2G  &\textbf{0.56}  &0.083  &1.27  &\textbf{0.105}  &1.72  &0.120  \\
                    MAT~\cite{li2022mat} &CVPR' 22 &60.0M    &288.8G  &0.98  &0.105  &2.45  &0.127  &4.76  &0.151  \\
                    ZITS~\cite{dong2022incremental} &CVPR' 22  &78.4M &666.9G &1.27 &0.096  &3.02  &0.112  &5.56  &0.137 \\
                    MI-GAN~\cite{Sargsyan_2023_ICCV} &ICCV' 23  &5.98M &15.69G   &0.58 &0.087 &1.29 &0.108  &1.74  &0.123  \\
                    StrDiffusion~\cite{Liu_2024_CVPR} &CVPR' 24 &300M    &3177G  &0.89  &0.107  &2.33  &0.130  &5.57  &0.153 \\
                    \hline
                    \cellcolor{green!20}Ours (LaMa)   &\cellcolor{green!20}- &\cellcolor{green!20}34.0M  &\cellcolor{green!20}463.0G  &\cellcolor{green!20}0.58 &\cellcolor{green!20}\textbf{0.081} &\cellcolor{green!20}\textbf{1.26} &\cellcolor{green!20}{0.106} &\cellcolor{green!20}\textbf{1.65} &\cellcolor{green!20}\textbf{0.119} \\
                    \hline
				\end{tabular}
			}
		\end{center}
\vspace{-15pt}
	\end{table}


\subsubsection{Quantitative analysis}
We compare our method with the typical image inpainting approaches, including: PENNet \cite{yan2019PENnet}, RFR \cite{li2020recurrent}, HAN \cite{deng2022hourglass}, FAR \cite{cao2022learning} and HiFill \cite{yi2020contextual} advocate an encoder-decoder paradigm; MEDFE \cite{Liu2019MEDFE}, MMT \cite{yu2022unbiased} and DGTS \cite{liu2022delving} focus on the texture feature map, overlooking the structure feature map; CTSDG \cite{Guo_2021_ICCV} presents mutually guidance between the structure and texture feature map; ZITS \cite{dong2022incremental} combining two types of high-frequency structure feature map, to further fuse the texture feature map during the convolutional process. Notably, to validate the advantages of our convolutional downsampling process, we \emph{substitute} all the encoders with ours from the latest arts \cite{yi2020contextual, yan2019PENnet}, leading to two variants of our architecture, namely \textit{Ours (PENNet)} and \textit{Ours (HiFill)}.
For fairness, we directly report the comparable results from \cite{deng2022hourglass, yu2022unbiased, dong2022incremental,cao2022learning}.

Table.\ref{tab:compare1} summarizes our findings below:
our method and its \emph{variants} enjoy a much smaller FID score, together with larger PSNR and SSIM than the competitors, confirming that our method can \emph{reconstruct the texture feature map via the structure feature map under both global and local side} (Sec.\ref{ilb}), \emph{benefiting our encoder with the significant advantages over others}. Notably, MMT, DGTS and MEDFE suffer from the non-ideal results since they both only focus on the texture feature map in the encoder, thus fail to extract the ideal structure and texture feature map, validating the benefits of reconstructing the texture feature map via the structure feature map during downsampling. Despite the intuition of the mutual guidance between the structure and texture, CTSDG receives an obvious performance loss, which verifies our proposal --- \emph{the structure feature map should guide the texture feature map reconstruction rather the inverse} (Sec.\ref{ilb}). Particularly, on Places2, our method outperforms ZITS with a large margin, implying the importance of \emph{alleviating the texture feature map loss during downsampling} (Sec.\ref{intro}), we also provide the quantitative results on high-resolution image in Table.\ref{tab:compare512} to prove that (\textit{more analysis can be seen in Appendix  Sec.D}).

\begin{figure}
\setlength{\abovecaptionskip}{0pt}
\setlength{\belowcaptionskip}{0pt}
  \centering
  \includegraphics[width=\linewidth]{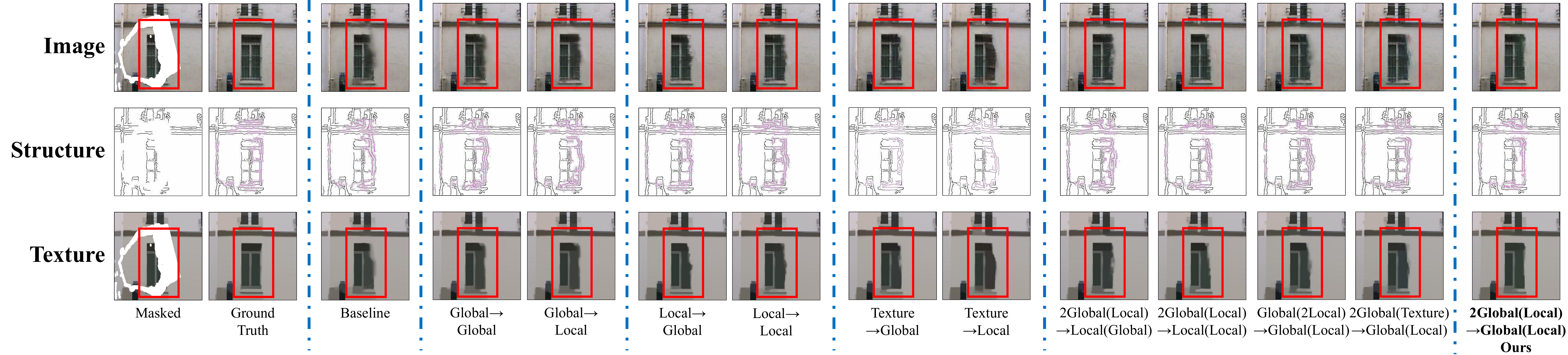}
  \caption{Visual analysis on the variants of our architecture. The consistent structure and texture (marked by the red box) for the inpainting results are visualized for our architecture.}\label{visual_different_arch}
\end{figure}

\begin{table}
\setlength{\abovecaptionskip}{0pt}
\setlength{\belowcaptionskip}{0pt}
    \scriptsize
    \centering
    \caption{Ablation study about the variants of our architecture. Our method, \emph{i.e.}, {2Global(Local) $\rightarrow$ Global(Local)}, achieves the best results (reported with \textbf{boldface}). }
    \resizebox{0.8\linewidth}{!}{
    \begin{tabular}{@{}|c||c|c||c|c||c|c|@{}}
    \hline
    \multirow{2}*{{Module}} & \multicolumn{2}{c||}{{PSNR$\uparrow$}}  &\multicolumn{2}{c||}{{SSIM$\uparrow$}}& \multicolumn{2}{c|}{{FID$\downarrow$} }\\
    &{10-30\%} &{30-50\%} &{10-30\%} &{30-50\%} &{10-30\%} &{30-50\%}  \\\hline\hline
\multicolumn{7}{|l|}{\cellcolor{blue!20}\textbf{\textit{\Rmnum{1}. Texture feature map only}}}\\\hline
    Baseline &29.93&24.66&0.963&0.891&19.47&51.90\\\hline
    \hline
\multicolumn{7}{|l|}{\cellcolor{blue!20}\textbf{\textit{\Rmnum{2}. Global structure feature map reconstructs the texture feature map}}}\\
    \hline 
    Global$\rightarrow$Global   &30.74&25.25&0.966&0.902&16.45&44.56\\
    Global$\rightarrow$Local  &31.02&25.52&0.971&0.910&14.69&41.81\\
    \hline
    \hline
\multicolumn{7}{|l|}{\cellcolor{blue!20}\textbf{\textit{\Rmnum{3}. Local residual structure feature map reconstructs the texture feature map}}}\\
    \hline 
    Local$\rightarrow$Global  &31.08&25.55&0.971&0.910&14.20&40.13\\
    Local$\rightarrow$Local &31.04&25.37&0.970&0.906&14.40&41.32\\\hline
    \hline
\multicolumn{7}{|l|}{\cellcolor{blue!20}\textbf{\textit{\Rmnum{4}. Texture feature map reconstructs the texture feature map}}}\\
    \hline 
    Texture$\rightarrow$Global  &30.74 &25.21&0.969 &0.900 &15.11 &43.29\\
    Texture$\rightarrow$Local &29.78  &24.42   &0.962 &0.881 &18.69 &54.11\\\hline
    \hline
\multicolumn{7}{|l|}{\cellcolor{blue!20}\textbf{\textit{\Rmnum{5}. Global and local residual structure simultaneously reconstructs the texture feature map}}}\\\hline
    2Global(Local)$\rightarrow$Local(Global) &31.14&25.66&0.972&0.911&13.93&39.59\\
    2Global(Local)$\rightarrow$Local(Local)  &30.82&25.35&0.970&0.906&16.42&47.15\\
    Global(2Local)$\rightarrow$Global(Local) &31.27&25.72&0.971&0.913&14.12&40.08\\
    2Global(Texture)$\rightarrow$Global(Local) &30.97&25.46&0.970&0.909&15.76&44.10\\
    \hline
  \cellcolor{green!20}\tabincell{c}{\textbf{2Global(Local)$\rightarrow$Global(Local)}  \\ \textbf{(Ours)}  }  &\cellcolor{green!20}\textbf{31.47}&\cellcolor{green!20}\textbf{25.93}&\cellcolor{green!20}\textbf{0.974}&\cellcolor{green!20}\textbf{0.917}&\cellcolor{green!20}\textbf{13.12}&\cellcolor{green!20}\textbf{38.62}\\

    \hline
    \end{tabular}
    \label{tab:compare}
    }
\vspace{-15pt}
\end{table}
\subsubsection{Qualitative analysis}
To shed more light on the advantages of our method, we further perform the visual analysis on the inpainted results.
Fig.\ref{visual_analysis1} delivers the following observations: PENNet, HiFill, MEDFE, DGTS and MMT inevitably generate some artifacts, implying the necessity of the structure information (Sec.\ref{ilb_1}).
Note that, CTSDG fails to inpaint the complex patterns, \emph{e.g.}, the door in the wall (the last row in Fig.\ref{visual_analysis1}(b)), owing to the inappropriate guidance between the structure and texture feature map (Sec.\ref{intro} and \ref{ilb}).
Besides, ZITS often exhibits the unreasonable texture feature map, \emph{e.g.}, the wing of bird (the second row in Fig.\ref{visual_analysis1}(b)), which attributes to the texture feature map loss during downsampling process (\textit{more analysis can be seen in Appendix Sec.D}). Such fact persuasively testifies the importance of \emph{reconstructing the texture feature map via the structure feature map during the downsampling} (Sec.\ref{intro}).

\subsection{Ablation Studies}
\subsubsection{Discussion on the variants of our architecture}
To unfold our contributions for the reconstruction between the structure and texture feature map, we conduct the ablation study on the variants of our architecture. Fig.\ref{visual_different_arch} illustrates that the inpainted images by our method exhibit the consistent structure and texture feature maps. For example, the shape of the window (the twelfth column in Fig.\ref{visual_different_arch}) is successfully restored, which is closely related to the ground truth, as opposed to other variants that produce the blurry ones. Table.\ref{tab:compare} also delivers the quantitative results with the similar trend. Such fact further discloses our core contribution --- \emph{reconstructing the texture {feature map} via the structure {feature map} under both global and local side} (Sec.\ref{ilb}). (\textit{more analysis can be seen in Appendix Sec.E})

\begin{figure}
\setlength{\abovecaptionskip}{0pt}
\setlength{\belowcaptionskip}{0pt}
  \centering
  \includegraphics[width=0.9\linewidth]{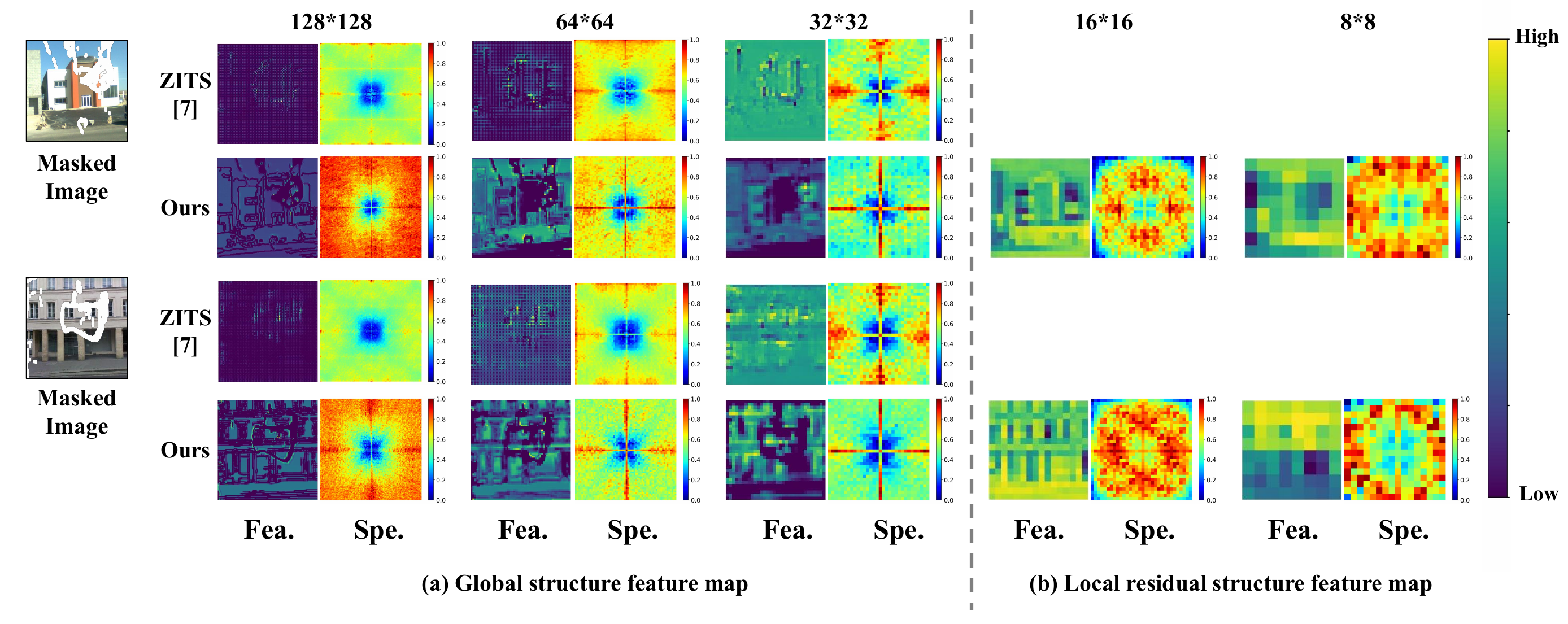}
  \caption{Ablation study about the reciprocal support between the global and local residual structure feature map. \textbf{Spe.}: spectrogram (the frequency increasing from the center to the surrounding) as the frequency domain version of the feature map (\textbf{Fea.}).  Particularly: both the higher value within the red area in \textbf{Spe.} and the yellow area in \textbf{Fea.} indicate more global (a) or local residual (b) structure information within the structure feature map.
  }\label{zits}
\vspace{-15pt}
\end{figure}

\subsubsection{Whether the global and local residual structure feature map really support each other?}
To verify the effectiveness of the reciprocal support between the global and local residual structure feature map, we compare our method with ZITS \cite{dong2022incremental}, which \emph{only contains 3 layers of CNNs} to avoid feature map loss. Fig.\ref{zits} illustrates that, compared to ZITS \cite{dong2022incremental}, our method preserves more global structure feature map (\emph{i.e.}, the structure consistent with the masked image in \textbf{Fea.})  at the early stage (\emph{e.g.}, 128$\times$128 and 64$\times$64), yet less at the later stage (\emph{e.g.}, 32$\times$32); while preserves the substantial local residual structure information (\emph{i.e.}, large red areas) from the reconstructed texture feature map at the later stage (\emph{e.g.}, 16$\times$16 and 8$\times$8), fostering the advantageous inpainting results (also see Fig.\ref{visual_analysis1}), which confirms the benefits of the \emph{reciprocal support} between the global and local residual structure feature map (Sec.\ref{our_arch}).

\begin{figure}
\setlength{\abovecaptionskip}{0pt}
\setlength{\belowcaptionskip}{0pt}
  \centering
  \includegraphics[width=0.8\linewidth]{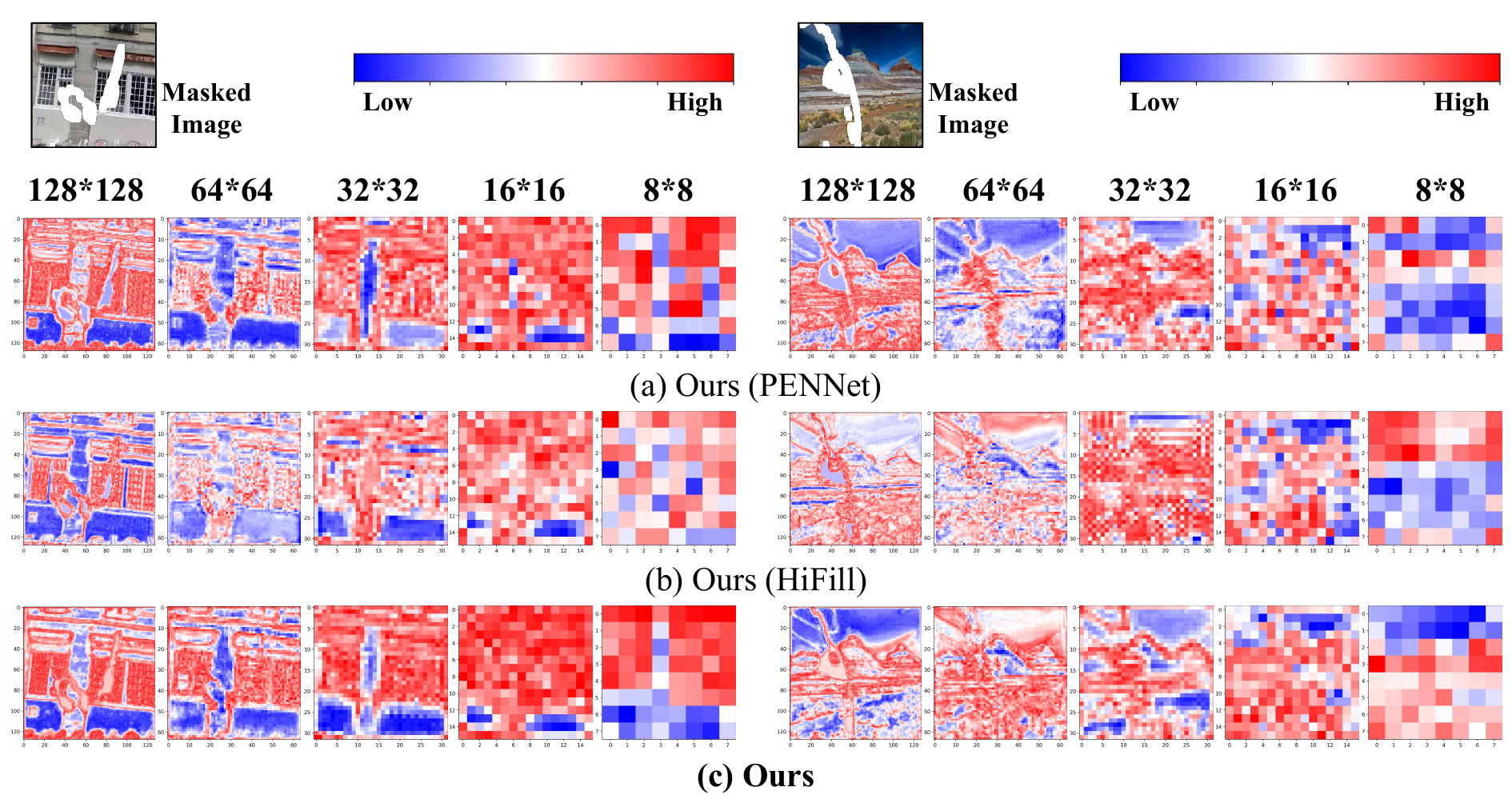}
  \caption{Ablation study about the cross-layer balance module. The heat map represents the information entropy value (channel dimension) of the texture reconstruction feature map. The higher value within the red area (corresponding to higher entropy value) indicates the larger information regarding  structure and texture feature map. }\label{cross_layer}
\vspace{-15pt}
\end{figure}

\subsubsection{Why should global and local residual structure feature map be balanced?}
To validate the effectiveness of the cross-layer balance module for \emph{ours}, we abandon it as \emph{Ours (PENNet)} and \emph{Ours (HiFill)}, then visualize the multi-scale feature maps that is finally fed to the decoder for inpainting in Fig.\ref{cross_layer}. Unlike \emph{Ours (PENNet)} and \emph{Ours (HiFill)}, where the range of red areas fluctuate sharply among different stages, resulting into the unbalanced local residual and global structure feature maps,  our method can focus on the related regions for inpainting (\emph{i.e}, the red area) of the feature maps among different stages, implying the consistency of feature maps from different stages, which confirms that local residual and global structure feature maps from different stages of CNNs for decoder are \emph{well balanced}, as claimed in Sec.\ref{gls}.

\begin{figure}
\setlength{\abovecaptionskip}{0pt}
\setlength{\belowcaptionskip}{0pt}
  \centering
  \includegraphics[width=0.8\linewidth]{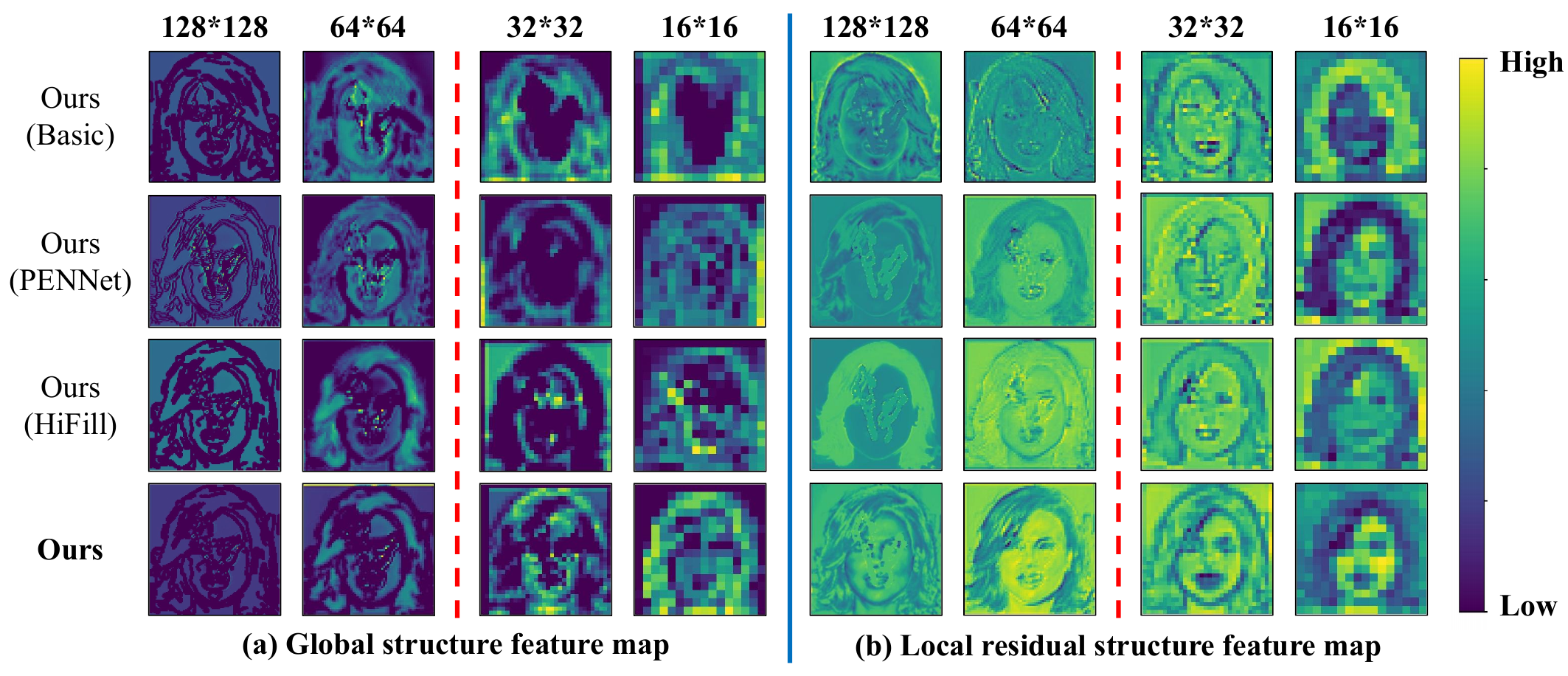}
  \caption{Ablation study about the affects of cross-layer balance module for global and local residual structure feature map. The higher value within the yellow area indicates more global (a) or local residual (b) structure information within the structure feature map.}\label{abs_gls}
\vspace{-15pt}
\end{figure}

\subsubsection{How the cross-layer balance module affects the global and local residual structure {feature map}?}
To further validate the effectiveness of the cross-layer balance module for the reconstruction from the global and local residual structure {feature maps} in our proposed architecture, \textit{Ours}, we remove the cross-layer balance module as \textit{Ours(Basic)}, which relies solely on a conventional decoder with the surrounding areas for inpainting with no additional operations for the masked region. For a more competitive comparison, we also validate inpainting methods based on global attention mechanisms, namely \textit{Ours(PENNet)} and \textit{Ours(HiFill)}. Fig.\ref{abs_gls} illustrates that both the global structure and local residual structure {feature maps} by our method can encode more structure information at multi-scales. For example, as shown in Fig.\ref{abs_gls}(a), at the early stage (\emph{e.g.}, 128$\times$128 and 64$\times$64),  for \textit{Ours(Basic)}, \textit{Ours(PENNet)} and \textit{Ours(HiFill)}, the feature map can well encode the global structure information (\textit{i.e.}, the global structure consistent with the masked image) rather than that over the late stage (\emph{e.g.}, 32$\times$32 and 16$\times$16). 

\section{Conclusion}
In this paper, we systematically answer whether and how structure and texture feature map can help alleviate the feature map loss during the convolutional downsampling process for image inpainting, while claim that reconstructing global (local) texture normalization feature map via global (local residual) structure feature map is the best during the convolutions for the encoder under both global and local side. We balance the reconstruction (denormalization) from global and local structure feature map to the texture feature map under global and local normalization via a cross-layer balance module for upsampling to the decoder. The extensive experimental results over the masked images from low-to-high resolutions validate its merits over state-of-the-arts.\\

{
    \bibliographystyle{IEEEtran}
    \bibliography{main}

@inproceedings{liuone,
  title={One Stone with Two Birds: A Null-Text-Null Frequency-Aware Diffusion Models for Text-Guided Image Inpainting},
  author={Liu, Haipeng and Wang, Yang and Wang, Meng},
  booktitle={The Thirty-ninth Annual Conference on Neural Information Processing Systems}
}

@article{haoran2023fine,
  title={Fine-grained cross-modal fusion based refinement for text-to-image synthesis},
  author={Haoran, Sun and Yang, Wang and Haipeng, Liu and Biao, Qian},
  journal={Chinese Journal of Electronics},
  volume={32},
  number={6},
  pages={1329--1340},
  year={2023},
  publisher={CIE}
}

@ARTICLE{10476709,
  author={Wang, Yang and Qian, Biao and Liu, Haipeng and Rui, Yong and Wang, Meng},
  journal={IEEE Transactions on Pattern Analysis and Machine Intelligence},
  title={Unpacking the Gap Box Against Data-Free Knowledge Distillation},
  year={2024},
  volume={46},
  number={9},
  pages={6280-6291},
  doi={10.1109/TPAMI.2024.3379505}}

@inproceedings{qian2023adaptive,
  title={Adaptive data-free quantization},
  author={Qian, Biao and Wang, Yang and Hong, Richang and Wang, Meng},
  booktitle={Proceedings of the IEEE/CVF Conference on Computer Vision and Pattern Recognition},
  pages={7960--7968},
  year={2023}
}

@inproceedings{qian2022switchable,
  title={Switchable online knowledge distillation},
  author={Qian, Biao and Wang, Yang and Yin, Hongzhi and Hong, Richang and Wang, Meng},
  booktitle={European Conference on Computer Vision},
  pages={449--466},
  year={2022},
  organization={Springer}
}

@InProceedings{Sargsyan_2023_ICCV,
    author    = {Sargsyan, Andranik and Navasardyan, Shant and Xu, Xingqian and Shi, Humphrey},
    title     = {MI-GAN: A Simple Baseline for Image Inpainting on Mobile Devices},
    booktitle = {Proceedings of the IEEE/CVF International Conference on Computer Vision (ICCV)},
    month     = {October},
    year      = {2023},
    pages     = {7335-7345}
}

@InProceedings{Liu_2024_CVPR,
    author    = {Liu, Haipeng and Wang, Yang and Qian, Biao and Wang, Meng and Rui, Yong},
    title     = {Structure Matters: Tackling the Semantic Discrepancy in Diffusion Models for Image Inpainting},
    booktitle = {Proceedings of the IEEE/CVF Conference on Computer Vision and Pattern Recognition (CVPR)},
    month     = {June},
    year      = {2024},
    pages     = {8038-8047}
}

@inproceedings{yu2020region,
  title={Region normalization for image inpainting},
  author={Yu, Tao and Guo, Zongyu and Jin, Xin and Wu, Shilin and Chen, Zhibo and Li, Weiping and Zhang, Zhizheng and Liu, Sen},
  booktitle={Proceedings of the AAAI conference on artificial intelligence},
  volume={34},
  number={07},
  pages={12733--12740},
  year={2020}
}

@inproceedings{wang2021parallel,
  title={Parallel multi-resolution fusion network for image inpainting},
  author={Wang, Wentao and Zhang, Jianfu and Niu, Li and Ling, Haoyu and Yang, Xue and Zhang, Liqing},
  booktitle={Proceedings of the IEEE/CVF International Conference on Computer Vision},
  pages={14559--14568},
  year={2021}
}

@inproceedings{dong2022incremental,
  title={Incremental transformer structure enhanced image inpainting with masking positional encoding},
  author={Dong, Qiaole and Cao, Chenjie and Fu, Yanwei},
  booktitle={Proceedings of the IEEE/CVF Conference on Computer Vision and Pattern Recognition},
  pages={11358--11368},
  year={2022}
}

@inproceedings{deng2022hourglass,
  title={Hourglass Attention Network for Image Inpainting},
  author={Deng, Ye and Hui, Siqi and Meng, Rongye and Zhou, Sanping and Wang, Jinjun},
  booktitle={Computer Vision--ECCV 2022: 17th European Conference, Tel Aviv, Israel, October 23--27, 2022, Proceedings, Part XVIII},
  pages={483--501},
  year={2022},
  organization={Springer}
}

@InProceedings{Guo_2021_ICCV,
    author    = {Guo, Xiefan and Yang, Hongyu and Huang, Di},
    title     = {Image Inpainting via Conditional Texture and Structure Dual Generation},
    booktitle = {Proceedings of the IEEE/CVF International Conference on Computer Vision (ICCV)},
    month     = {October},
    year      = {2021},
    pages     = {14134-14143}
}

@inproceedings{li2022mat,
  title={Mat: Mask-aware transformer for large hole image inpainting},
  author={Li, Wenbo and Lin, Zhe and Zhou, Kun and Qi, Lu and Wang, Yi and Jia, Jiaya},
  booktitle={Proceedings of the IEEE/CVF conference on computer vision and pattern recognition},
  pages={10758--10768},
  year={2022}
}

@article{luo2023image,
  title={Image Restoration with Mean-Reverting Stochastic Differential Equations},
  author={Luo, Ziwei and Gustafsson, Fredrik K and Zhao, Zheng and Sj{\"o}lund, Jens and Sch{\"o}n, Thomas B},
  journal={International Conference on Machine Learning},
  year={2023},
  organization={PMLR}
}

@inproceedings{yu2022unbiased,
  title={Unbiased Multi-modality Guidance for Image Inpainting},
  author={Yu, Yongsheng and Du, Dawei and Zhang, Libo and Luo, Tiejian},
  booktitle={Computer Vision--ECCV 2022: 17th European Conference, Tel Aviv, Israel, October 23--27, 2022, Proceedings, Part XVI},
  pages={668--684},
  year={2022},
  organization={Springer}
}

@inproceedings{liu2022delving,
  title={Delving Globally into Texture and Structure for Image Inpainting},
  author={Liu, Haipeng and Wang, Yang and Wang, Meng and Rui, Yong},
  booktitle={Proceedings of the 30th ACM International Conference on Multimedia},
  pages={1270--1278},
  year={2022}
}

@InProceedings{Liao_2021_CVPR,
    author    = {Liao, Liang and Xiao, Jing and Wang, Zheng and Lin, Chia-Wen and Satoh, Shin'ichi},
    title     = {Image Inpainting Guided by Coherence Priors of Semantics and Textures},
    booktitle = {Proceedings of the IEEE/CVF Conference on Computer Vision and Pattern Recognition (CVPR)},
    month     = {June},
    year      = {2021},
    pages     = {6539-6548}
}

@inproceedings{li2020recurrent,
  title={Recurrent feature reasoning for image inpainting},
  author={Li, Jingyuan and Wang, Ning and Zhang, Lefei and Du, Bo and Tao, Dacheng},
  booktitle={Proceedings of the IEEE/CVF Conference on Computer Vision and Pattern Recognition},
  pages={7760--7768},
  year={2020}
}

@inproceedings{yan2019PENnet,
  author = {Zeng, Yanhong and Fu, Jianlong and Chao, Hongyang and Guo, Baining},
  title = {Learning Pyramid-Context Encoder Network for High-Quality Image Inpainting},
  booktitle = {The IEEE Conference on Computer Vision and Pattern Recognition (CVPR)},
  pages={1486--1494},
  year = {2019}
}

@article{wang2022progressive,
  title={Progressive learning with multi-scale attention network for cross-domain vehicle re-identification},
  author={Wang, Yang and Peng, Jinjia and Wang, Huibing and Wang, Meng},
  journal={Science China Information Sciences},
  volume={65},
  number={6},
  pages={160103},
  year={2022},
  publisher={Springer}
}

@inproceedings{zhang2018unreasonable,
  title={The unreasonable effectiveness of deep features as a perceptual metric},
  author={Zhang, Richard and Isola, Phillip and Efros, Alexei A and Shechtman, Eli and Wang, Oliver},
  booktitle={Proceedings of the IEEE conference on computer vision and pattern recognition},
  pages={586--595},
  year={2018}
}

@inproceedings{suvorov2022resolution,
  title={Resolution-robust large mask inpainting with fourier convolutions},
  author={Suvorov, Roman and Logacheva, Elizaveta and Mashikhin, Anton and Remizova, Anastasia and Ashukha, Arsenii and Silvestrov, Aleksei and Kong, Naejin and Goka, Harshith and Park, Kiwoong and Lempitsky, Victor},
  booktitle={Proceedings of the IEEE/CVF winter conference on applications of computer vision},
  pages={2149--2159},
  year={2022}
}

@article{zhao2021large,
  title={Large scale image completion via co-modulated generative adversarial networks},
  author={Zhao, Shengyu and Cui, Jonathan and Sheng, Yilun and Dong, Yue and Liang, Xiao and Chang, Eric I and Xu, Yan},
  journal={arXiv preprint arXiv:2103.10428},
  year={2021}
}

@article{zhu2021image,
  title={Image inpainting by end-to-end cascaded refinement with mask awareness},
  author={Zhu, Manyu and He, Dongliang and Li, Xin and Li, Chao and Li, Fu and Liu, Xiao and Ding, Errui and Zhang, Zhaoxiang},
  journal={IEEE Transactions on Image Processing},
  volume={30},
  pages={4855--4866},
  year={2021},
  publisher={IEEE}
}

@article{zeng2022aggregated,
  title={Aggregated contextual transformations for high-resolution image inpainting},
  author={Zeng, Yanhong and Fu, Jianlong and Chao, Hongyang and Guo, Baining},
  journal={IEEE Transactions on Visualization and Computer Graphics},
  year={2022},
  publisher={IEEE}
}

@article{ma2022regionwise,
  title={Regionwise generative adversarial image inpainting for large missing areas},
  author={Ma, Yuqing and Liu, Xianglong and Bai, Shihao and Wang, Lei and Liu, Aishan and Tao, Dacheng and Hancock, Edwin R},
  journal={IEEE transactions on cybernetics},
  year={2022},
  publisher={IEEE}
}

@InProceedings{Nazeri_2019_ICCV,
  title = {EdgeConnect: Structure Guided Image Inpainting using Edge Prediction},
  author = {Nazeri, Kamyar and Ng, Eric and Joseph, Tony and Qureshi, Faisal and Ebrahimi, Mehran},
  booktitle = {The IEEE International Conference on Computer Vision (ICCV) Workshops},
  month = {Oct},
  year = {2019}
}

@inproceedings{yu2018generative,
  title={Generative image inpainting with contextual attention},
  author={Yu, Jiahui and Lin, Zhe and Yang, Jimei and Shen, Xiaohui and Lu, Xin and Huang, Thomas S},
  booktitle={Proceedings of the IEEE conference on computer vision and pattern recognition},
  pages={5505--5514},
  year={2018}
}

@inproceedings{cao2022learning,
  title={Learning Prior Feature and Attention Enhanced Image Inpainting},
  author={Cao, Chenjie and Dong, Qiaole and Fu, Yanwei},
  booktitle={Computer Vision--ECCV 2022: 17th European Conference, Tel Aviv, Israel, October 23--27, 2022, Proceedings, Part XV},
  pages={306--322},
  year={2022},
  organization={Springer}
}

@article{liu2025few,
  title={Few-Shot Referring Video Single-and Multi-Object Segmentation Via Cross-Modal Affinity with Instance Sequence Matching},
  author={Liu, Heng and Li, Guanghui and Gao, Mingqi and Zhen, Xiantong and Zheng, Feng and Wang, Yang},
  journal={International Journal of Computer Vision},
  pages={1--19},
  year={2025},
  publisher={Springer}
}

@inproceedings{qian2023rethinking,
  title={Rethinking data-free quantization as a zero-sum game},
  author={Qian, Biao and Wang, Yang and Hong, Richang and Wang, Meng},
  booktitle={Proceedings of the AAAI conference on artificial intelligence},
  volume={37},
  number={8},
  pages={9489--9497},
  year={2023}
}

@inproceedings{he2022masked,
  title={Masked autoencoders are scalable vision learners},
  author={He, Kaiming and Chen, Xinlei and Xie, Saining and Li, Yanghao and Doll{\'a}r, Piotr and Girshick, Ross},
  booktitle={Proceedings of the IEEE/CVF Conference on Computer Vision and Pattern Recognition},
  pages={16000--16009},
  year={2022}
}

@inproceedings{park2019semantic,
  title={Semantic image synthesis with spatially-adaptive normalization},
  author={Park, Taesung and Liu, Ming-Yu and Wang, Ting-Chun and Zhu, Jun-Yan},
  booktitle={Proceedings of the IEEE/CVF conference on computer vision and pattern recognition},
  pages={2337--2346},
  year={2019}
}

@inproceedings{ronneberger2015u,
  title={U-net: Convolutional networks for biomedical image segmentation},
  author={Ronneberger, Olaf and Fischer, Philipp and Brox, Thomas},
  booktitle={Medical Image Computing and Computer-Assisted Intervention--MICCAI 2015: 18th International Conference, Munich, Germany, October 5-9, 2015, Proceedings, Part III 18},
  pages={234--241},
  year={2015},
  organization={Springer}
}

@inproceedings{liu2021pd,
  title={Pd-gan: Probabilistic diverse gan for image inpainting},
  author={Liu, Hongyu and Wan, Ziyu and Huang, Wei and Song, Yibing and Han, Xintong and Liao, Jing},
  booktitle={Proceedings of the IEEE/CVF Conference on Computer Vision and Pattern Recognition},
  pages={9371--9381},
  year={2021}
}

@inproceedings{wan2021high,
  title={High-Fidelity Pluralistic Image Completion with Transformers},
  author={Wan, Ziyu and Zhang, Jingbo and Chen, Dongdong and Liao, Jing},
  booktitle={2021 IEEE/CVF International Conference on Computer Vision (ICCV)},
  pages={4672--4681},
  year={2021},
  organization={IEEE}
}

@inproceedings{Liu2019MEDFE,
  title={Rethinking Image Inpainting via a Mutual Encoder-Decoder with Feature Equalizations},
  author={Liu, Hongyu and Jiang, Bin and Song, Yibing and Huang, Wei and Yang, Chao},
  booktitle={European Conference on Computer Vision},
  pages={725--741},
  year={2020}
}

@misc{yi2020contextual,
    title={Contextual Residual Aggregation for Ultra High-Resolution Image Inpainting},
    author={Zili Yi and Qiang Tang and Shekoofeh Azizi and Daesik Jang and Zhan Xu},
    year={2020},
    eprint={2005.09704},
    archivePrefix={arXiv},
    primaryClass={cs.CV}
}

@inproceedings{liu2018image,
  title={Image inpainting for irregular holes using partial convolutions},
  author={Liu, Guilin and Reda, Fitsum A and Shih, Kevin J and Wang, Ting-Chun and Tao, Andrew and Catanzaro, Bryan},
  booktitle={Proceedings of the European conference on computer vision (ECCV)},
  pages={85--100},
  year={2018}
}

@inproceedings{pathak2016context,
  title={Context encoders: Feature learning by inpainting},
  author={Pathak, Deepak and Krahenbuhl, Philipp and Donahue, Jeff and Darrell, Trevor and Efros, Alexei A},
  booktitle={Proceedings of the IEEE conference on computer vision and pattern recognition},
  pages={2536--2544},
  year={2016}
}

@inproceedings{xie2019image,
  title={Image inpainting with learnable bidirectional attention maps},
  author={Xie, Chaohao and Liu, Shaohui and Li, Chao and Cheng, Ming-Ming and Zuo, Wangmeng and Liu, Xiao and Wen, Shilei and Ding, Errui},
  booktitle={Proceedings of the IEEE/CVF international conference on computer vision},
  pages={8858--8867},
  year={2019}
}

@inproceedings{yu2019free,
  title={Free-form image inpainting with gated convolution},
  author={Yu, Jiahui and Lin, Zhe and Yang, Jimei and Shen, Xiaohui and Lu, Xin and Huang, Thomas S},
  booktitle={Proceedings of the IEEE/CVF international conference on computer vision},
  pages={4471--4480},
  year={2019}
}

@inproceedings{zeng2021generative,
  title={CR-Fill: Generative Image Inpainting with Auxiliary Contextual Reconstruction},
  author={Zeng, Yu and Lin, Zhe and Lu, Huchuan and Patel, Vishal M.},
  booktitle={Proceedings of the IEEE International Conference on Computer Vision},
  year={2021}
}

@inproceedings{li2022misf,
  title={MISF: Multi-level interactive Siamese filtering for high-fidelity image inpainting},
  author={Li, Xiaoguang and Guo, Qing and Lin, Di and Li, Ping and Feng, Wei and Wang, Song},
  booktitle={Proceedings of the IEEE/CVF Conference on Computer Vision and Pattern Recognition},
  pages={1869--1878},
  year={2022}
}

@article{canny1986computational,
  title={A computational approach to edge detection},
  author={Canny, John},
  journal={IEEE Transactions on pattern analysis and machine intelligence},
  number={6},
  pages={679--698},
  year={1986},
  publisher={Ieee}
}

@inproceedings{ren2019structureflow,
  title={Structureflow: Image inpainting via structure-aware appearance flow},
  author={Ren, Yurui and Yu, Xiaoming and Zhang, Ruonan and Li, Thomas H and Liu, Shan and Li, Ge},
  booktitle={Proceedings of the IEEE/CVF International Conference on Computer Vision},
  pages={181--190},
  year={2019}
}

@inproceedings{xu2011image,
  title={Image smoothing via L 0 gradient minimization},
  author={Xu, Li and Lu, Cewu and Xu, Yi and Jia, Jiaya},
  booktitle={Proceedings of the 2011 SIGGRAPH Asia conference},
  pages={1--12},
  year={2011}
}

@article{heusel2017gans,
  title={Gans trained by a two time-scale update rule converge to a local nash equilibrium},
  author={Heusel, Martin and Ramsauer, Hubert and Unterthiner, Thomas and Nessler, Bernhard and Hochreiter, Sepp},
  journal={Advances in neural information processing systems},
  volume={30},
  year={2017}
}

@article{wang2004image,
  title={Image quality assessment: from error visibility to structural similarity},
  author={Wang, Zhou and Bovik, Alan C and Sheikh, Hamid R and Simoncelli, Eero P},
  journal={IEEE transactions on image processing},
  volume={13},
  number={4},
  pages={600--612},
  year={2004},
  publisher={IEEE}
}
}

\appendix

\subsection{Overview}
Due to page limitation of the main body, as indicated by our submission, the supplementary material offers further discussion on more qualitative and quantitative results with high-resolution image and more ablation studies results, which are summarized below:
\begin{itemize}
    \item The discussion on the model complexity of our encoder, as mentioned in Sec.III-B4 of the main body (Sec.\ref{sec1})
    \item More descriptions on the implementation details, as mentioned in Sec.IV-A of the main body (Sec.\ref{sec2})
    \item Additional quantitative results and qualitative analysis on \emph{high-resolution image} for the comparison with state-of-the-arts, as mentioned in Sec.IV-B of the main body (Sec.\ref{sec3}).
    \item Additional results for the ablation study about more discussion on the variants of our architecture, as mentioned in Sec.IV-C1 of the main body (Sec.\ref{sec4})
    \item Additional results for the ablation study about various structure guidance to alleviate the texture feature map loss during the convolutional downsampling process, as mentioned in Sec.III-A1 of the main body (Sec.\ref{sec5})
    \item Additional results for the ablation study about hyperparameter sensitivity analysis of $\tau$ in adjusted transformer blocks, as mentioned in Sec.III-A2 of the main body (Sec.\ref{sec6})
\end{itemize}

\subsection{The discussion on model complexity of our encoder}\label{sec1}
When analyzing the overall complexity of the model, we need to consider the computational load of each component. In our architecture, we employ the attention mechanism based on the Swin Transformer to handle texture information. This mechanism divides the image into fixed-size windows and performs self-attention computation within each window. The advantage of this design is that the attention computation within each window depends only on the pixels within the window, rather than the entire image. Specifically, for a texture feature map $F_{k}^{T} \in \mathbb{R}^{C_{T} \times H_{k} \times W_{k}}$, the complexity of the attention mechanism can be expressed as $O\left((H_{k} \times W_{k}) \times (h_{k} \times w_{k}) \times C_{T}\right)$, where $h_{k} \times w_{k}$ represents the size of the window.

Additionally, we utilize CNNs to process high-frequency structure information. For a structure feature map $F_{k}^{S} \in \mathbb{R}^{C_{S} \times H_{k} \times W_{k}}$, the complexity is typically represented as $O\left((H_{k} \times W_{k}) \times (k_{H} \times k_{W}) \times C_{S}\right)$, where $k_{H} \times k_{W}$ represents the size of the convolutional kernel.

As the structure feature map guides the texture feature map, we normalize the texture feature map and then reconstruct it via denormalization using the structure feature map. Since both normalization and denormalization are element-wise operations, their complexity is $O\left((H_{k} \times W_{k}) \times C_{T}\right)$.

In summary, the overall complexity of the model exhibits linear computational complexity with respect to the image size. Finally, we present our complexity via FLoating-point OPerations (FLOPs) w.r.t. $H_{k} \times W_{k}$   and parameters (Params) w.r.t. $k_{H} \times k_{W}$ of the model, while compare it with other baseline models, as shown in Table.\ref{complexity}. Our inpainting model demonstrates the ability to minimize the number of model parameters and computational resources required, while outperforming other baseline models.

\begin{table}[t]
\setlength{\abovecaptionskip}{0pt}
\setlength{\belowcaptionskip}{0pt}
\centering
\caption{Model complexity. We provide the Floating-point Operations (FLOPs) and parameters (Params) of the model. Our inpainting model demonstrates the ability to minimize the size of model parameters and computational resources. M and G denote $10^{6}$ and $10^{9}$, respectively. }\label{complexity}
\resizebox{\linewidth}{!}{%
\begin{tabular}{c|cccccc|c}
\hline
\multirow{2}{*}{\textbf{Model}} & \textbf{GC} & \textbf{RFR} & \textbf{CTSDG} & \textbf{HAN} & \textbf{MMT} & \textbf{ZITS} & \multirow{2}{*}{\textbf{Ours}} \\
& \cite{yu2019free} & \cite{li2020recurrent} & \cite{Guo_2021_ICCV} & \cite{deng2022hourglass} & \cite{yu2022unbiased} & \cite{dong2022incremental} & \\
\hline
\textbf{FLOPs} & 103.1G & 206.1G & 75.9G & 137.7G & 98.3G & 112.4G & 95.1G \\
\textbf{Params} & 16.0M & 30.6M & 52.1M & 19.4M & 51.3M & 86.6M & 37.8M \\
\hline
\end{tabular}%
}
\vspace{-15pt}
\end{table}

\subsection{More descriptions on the implementation details}\label{sec2}
Due to page limitation, we offer more descriptions of the implementation details. For $256 \times 256$ images, the network is trained using the NVIDIA Irregular Mask Dataset \cite{liu2018image}. The mask sizes are based on different ratios of the masked region to the entire image, categorized into six groups: (0.01, 0.1], (0.1, 0.2], (0.2, 0.3], (0.3, 0.4], (0.4, 0.5], and (0.5, 0.6]. Each group is further divided into two types: masks that are either close to the border or distant from it. For the distant type, masks are placed at least 50 pixels away from the image border. For $512 \times 512$ images, we adopt the masks from \cite{suvorov2022resolution} that cover more than 50\% of the image. These masks are categorized into three sizes: \textit{Narrow}, \textit{Medium}, and \textit{Wide}. They include the types from polygonal chains dilated by a large random width (wide masks) and rectangles of arbitrary aspect ratios (box masks).
In our framework, the channel number of the texture feature map is tied to the token vector dimensions. Lower dimensions can limit the tokens' representational capacity, while higher dimensions increase the computational burden of the self-attention mechanism. To address this, we propose multi-scale texture feature maps with a uniform channel number of 180 across all layers in both the convolutional head and the transformer body.

\subsection{Additional quantitative and qualitative analysis on high-resolution image}\label{sec3}
Due to page limitation, we offer more quantitative and qualitative analysis on high-resolution image. We apply our strategy to masked images over high resolutions of $\mathbf{512 \times 512}$. \textit{Previous methods \cite{Nazeri_2019_ICCV}, \cite{yu2018generative}, \cite{ma2022regionwise}, \cite{yi2020contextual}, \cite{zeng2022aggregated}, \cite{zhu2021image}, \cite{zhao2021large}, \cite{Liu_2024_CVPR}, \cite{Sargsyan_2023_ICCV} encountered challenges in reducing texture feature map loss during the convolutional downsampling process}. \textit{As a solution, they opted for fewer convolutional layers, which may compromise the capture of semantic information.} The typical higher resolution image inpainting approach LaMa \cite{suvorov2022resolution} achieves a global receptive field within the limited downsampling layers, \textit{i.e.,} 3 convolutional downsampling layers only for the texture feature map, together with 9 residual blocks.  Hence, LaMa struggles to capture advanced semantic information, resulting into the generation of stable repeating texture feature maps rather than clear inpainted regions. With the same number of downsampling layers, Big-LaMa  \cite{suvorov2022resolution} extends LaMa with 18 residual blocks, to capture more semantic information. However, \textit{Big-LaMa still failed to remedy the texture feature map loss caused by convolutional downsampling. Therefore, it can only downsample to $64 \times 64$ consistent with LaMa}; MAT \cite{li2022mat} focuses on the texture feature map, overlooking the structure feature map; In contrast, our method reconstructs the texture feature map under global (local) normalization via global (local residual) structure feature maps, thus alleviating texture feature map loss as convolutional downsampling progresses to deeper layers.
To validate this, we substitute the encoder of LaMa with ours while making up additional 2 convolutional downsampling layers, totally 5 layers, to enhance its semantic information, resulting into the variant of our architecture, denoted as \textit{Ours (LaMa)}.

\begin{table}[t]
\setlength{\abovecaptionskip}{0pt}
\setlength{\belowcaptionskip}{0pt}
		\begin{center}
		\caption{Comparison of quantitative results (\emph{i.e.}, FID and LPIPS) under varied mask ratios on Places2 ($512 \times 512$). We also provide the FLoating-point OPerations (FLOPs) and parameters (Params). M and G denote $10^{6}$ and $10^{9}$,  The best results are reported with \textbf{boldface}. $\downarrow$: Lower is better.}\label{tab:compare512}
			\resizebox{\linewidth}{!}{
				\begin{tabular}{|l|c|| c|c || c |c || c| c ||  c|c || }
					\hline
					\multirow{3}{*}{\textbf{Method}} &\multirow{3}{*}{\textbf{Venue}} & \multirow{3}{*}{\textbf{Params}} & \multirow{3}{*}{\textbf{FLOPs}} & \multicolumn{6}{c||}{\cellcolor{blue!20}{\textbf{Places} ($512 \times 512$)}}  \\
					~ & ~ & ~& ~ & \multicolumn{2}{c||}{\textbf{Narrow Mask}} & \multicolumn{2}{c||}{\textbf{Medium Mask}} &\multicolumn{2}{c||}{\textbf{Wide Mask}}\\
					~ & ~ & ~ & ~ & \textbf{FID}$\downarrow$ & \textbf{LPIPS}$\downarrow$  & \textbf{FID}$\downarrow$ & \textbf{LPIPS}$\downarrow$  & \textbf{FID}$\downarrow$ & \textbf{LPIPS}$\downarrow$  \\
					\hline
                    EdgeConnect~\cite{Nazeri_2019_ICCV} &ICCV' 19 &22.0M &490.4G  &1.33 &0.111 &3.66 &0.135 &8.37 &0.160 \\
                    DeepFill v2~\cite{yu2018generative}  &CVPR'19 &4.00M  &100.8G  &1.06 &0.104 &2.68 &0.130 &5.20 &0.155 \\
                    RegionNorm~\cite{yu2020region} &AAAI' 20 &12.0M &216.0G &2.13 &0.120 &6.27 &0.146 &15.7 &0.176 \\
                    HiFill~\cite{yi2020contextual} &CVPR' 20 &3.00M   &72.10G  &9.24 &0.218 &7.54 &0.157 &12.8 &0.180 \\
                    CoModGAN~\cite{zhao2021large} &ICLR' 20 &109M  &121.09G  &0.82 &0.111 &1.34 &0.128 &1.82 &0.147 \\
                    MADF~\cite{zhu2021image}  &TIP' 21 &85.0M   &183.7G &0.57 &0.085 &1.62 &0.113 &3.76 &0.139\\
                    RegionWise~\cite{ma2022regionwise}  &TCYB' 22 &47.0M  &35.00G  &0.90 &0.102 &2.42 &0.125 &4.75 &0.149  \\
                    AOT GAN~\cite{zeng2022aggregated}  &TVCG' 22 &15.0M  &291.5G  &0.79 &0.091 &2.29 &0.119 &5.94 &0.149 \\
                    LaMa~\cite{suvorov2022resolution} &WACV' 22 &27.0M    &171.3G  &0.63 &0.090 &1.30 &0.112 &2.21 &0.135 \\
                    Big-LaMa~\cite{suvorov2022resolution} &WACV' 22 &51.0M    &264.2G  &\textbf{0.56}  &0.083  &1.27  &\textbf{0.105}  &1.72  &0.120  \\
                    MAT~\cite{li2022mat} &CVPR' 22 &60.0M    &288.8G  &0.98  &0.105  &2.45  &0.127  &4.76  &0.151  \\
                    ZITS~\cite{dong2022incremental} &CVPR' 22  &78.4M &666.9G &1.27 &0.096  &3.02  &0.112  &5.56  &0.137 \\
                    MI-GAN~\cite{Sargsyan_2023_ICCV} &ICCV' 23  &5.98M &15.69G   &0.58 &0.087 &1.29 &0.108  &1.74  &0.123  \\
                    StrDiffusion~\cite{Liu_2024_CVPR} &CVPR' 24 &300M    &3177G  &0.89  &0.107  &2.33  &0.130  &5.57  &0.153 \\
                    \hline
                    \cellcolor{green!20}Ours (LaMa)   &\cellcolor{green!20}- &\cellcolor{green!20}34.0M  &\cellcolor{green!20}463.0G  &\cellcolor{green!20}0.58 &\cellcolor{green!20}\textbf{0.081} &\cellcolor{green!20}\textbf{1.26} &\cellcolor{green!20}{0.106} &\cellcolor{green!20}\textbf{1.65} &\cellcolor{green!20}\textbf{0.119} \\
                    \hline
				\end{tabular}
			}
		\end{center}
\vspace{-15pt}
	\end{table}

\begin{figure}
\setlength{\abovecaptionskip}{0pt}
\setlength{\belowcaptionskip}{0pt}
  \centering
  \includegraphics[width=\linewidth]{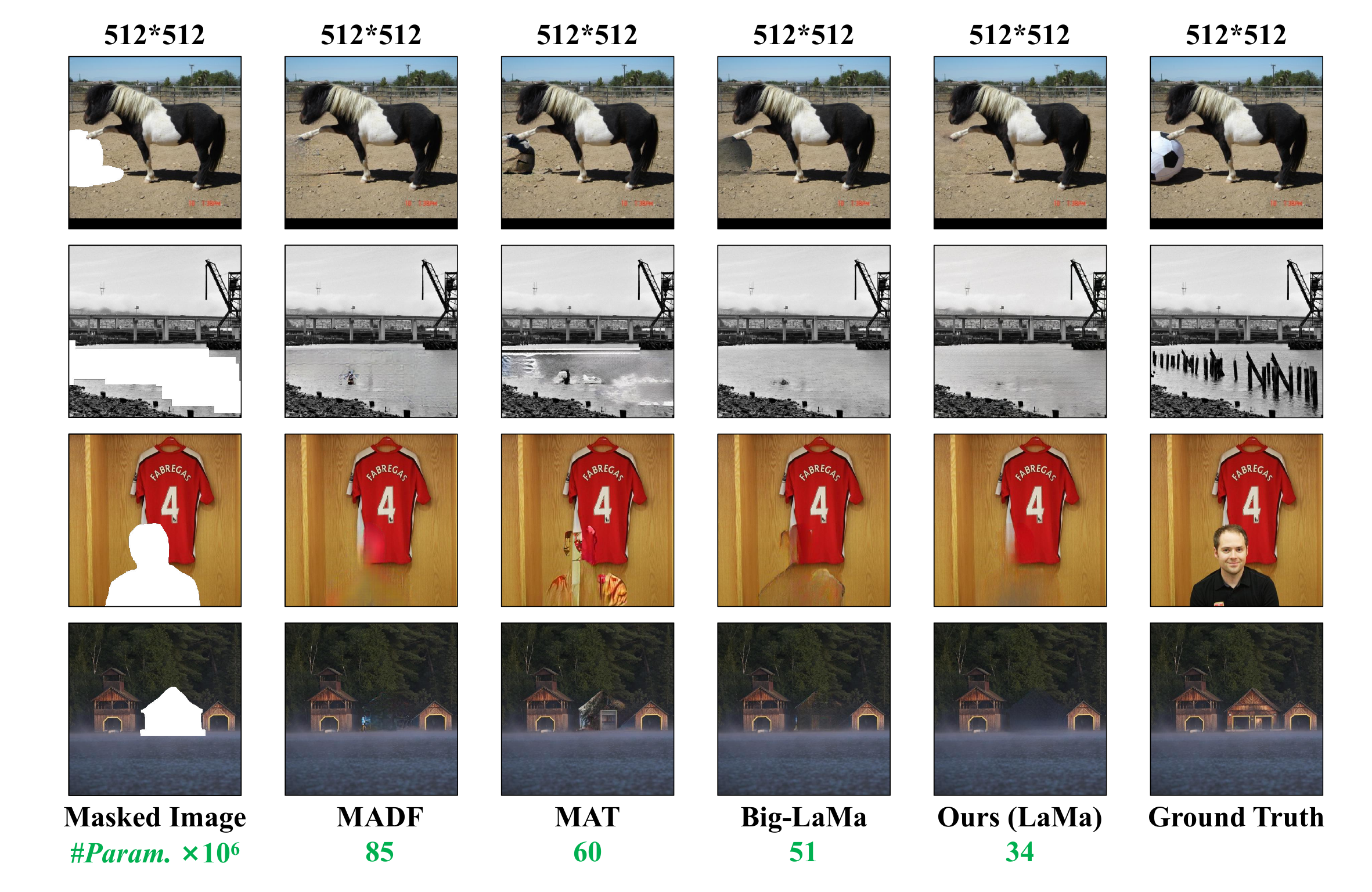}
  \caption{Comparison of the final inpainted results with the state-of-the-arts on Places2 ($512 \times 512$) with object-shaped masks. \textit{Ours (LaMa)} delivers the desirable inpainting results against others.}\label{fig:compare512}
\vspace{-15pt}
\end{figure}
We utilize FID and LPIPS \cite{zhang2018unreasonable} to assess the performance of all compared methods with irregular masks from \cite{suvorov2022resolution}, namely narrow, medium, and wide masks. Table.\ref{tab:compare512} summarizes our findings below: our method \textit{Ours (LaMa)} enjoys a much smaller FID and LPIPS score than the competitors, confirming that our method can well alleviate the texture feature map loss via the reconstruction from the structure feature map at both global and local sides  (Sec.III-B), thereby endowing our encoder with the advantages over competing methods.  Notably, LaMa remains the large performance margins (at most 25.3\% for FID and 11.8\% for LPIPS) compared to \textit{Ours (LaMa)}, owing to the texture feature map loss during downsampling. Fig.\ref{fig:compare512} also proves that. \textit{Ours (LaMa)} outperforms the recent work MI-GAN \cite{Sargsyan_2023_ICCV} and StrDiffusion \cite{Liu_2024_CVPR} in terms of FID and LPIPS. \textit{In particular, MI-GAN \cite{Sargsyan_2023_ICCV} distills an informative texture feature map from the heavyweight neural network, CoModGAN \cite{zhao2021large}, using multiple convolutional kernels in each downsampling layer to closely match the network’s capacity with that of its teacher network. However, this texture-only method overlooks structural information, and the extreme lightweight design sacrifices the performance, while suffers from heavy computational complexity for training, which cost nearly one month with 8 Nvidia A6000 GPUs}. StrDiffusion \cite{Liu_2024_CVPR} addresses the semantic discrepancy between masked and unmasked regions during the denoising process of DDPM-based inpainting methods by incorporating auxiliary structural information. However, maintaining semantic coherence between the denoised texture and structure during inference becomes challenging, especially for high-resolution images, where the downsampling process disrupts both structural and textural information.

We offer the Floating-point Operations (FLOPs) and parameters (Params) of our model, while compare with other baseline models, as shown
in Table.\ref{tab:compare512}. In term of parameter comparison, our method is more efficient than Big-LaMa and MAT, while comparable to the LaMa, yet outperform it for inpainting result in term of reconstructed texture feature map. Fig.\ref{fig:compare512} demonstrates that \textit{Ours (LaMa)} is significantly more parameter-efficient than Big-LaMa and MAT, while enjoy ideal inpainting results. Our floating-point operation demonstrates the advantages of our model with fewer computational resources than those structure-guided inpainting methods, i.e., ZITS and StrDiffusion, which also prove that using structure to reduce information loss during texture downsampling is the most effective way to leverage structure in image inpainting.

Specifically, due to the fewer downsampling layers, LaMa struggles to capture sufficient semantic information, as seen in the football regions in the first row of Fig.\ref{fig:compare512}. Besides, the MAT often exhibits the unreasonable texture feature map, \textit{e.g.}, the last row of Fig.\ref{fig:compare512}, which attributes to the texture feature map loss without the reconstruction from the structure feature map.  Such fact  testifies the importance of \emph{reconstructing the texture feature map via the structure feature map during the convolutional downsampling}, which is consistent to the intuitions as stated in Sec.I of the main body.

\begin{figure}
\setlength{\abovecaptionskip}{0pt}
\setlength{\belowcaptionskip}{0pt}
  \centering
  \includegraphics[width=0.9\linewidth]{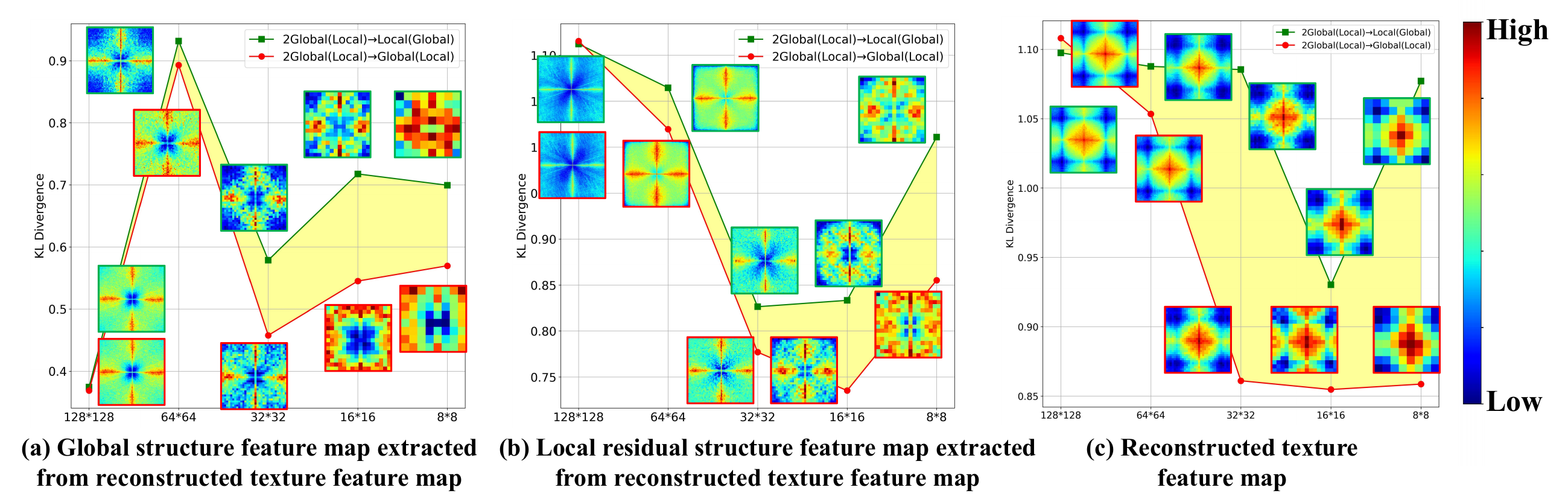}
  \caption{Comparison of global structure feature map (a) extracted from the reconstructed texture feature map via high-pass filter operator and reconstructed texture feature map (c), apart from local residual structure feature map (b) acquired by subtracting the \textit{reconstructed} texture feature maps before and after downsampling. Please refer to Fig.9 of the main body for more details about the heat map, particularly:  the higher value within the red area indicates more structure or texture information in (a)(b) or (c) {over the feature maps}.}\label{spectrogram_single-c}

\vspace{-15pt}
\end{figure}
\subsection{More Discussion on the variants of our architecture}\label{sec4}
As mentioned in Sec.IV-C1 of the main body, due to page limitation, we further provide more results to validate advantages of reconstructing texture feature maps via structure feature maps from global and local side, named \textbf{2Global(Local) → Global(Local)} than others (see Sec.II-B4), we extract the global structure feature map from the reconstructed texture feature map via high-pass filter and local residual structure feature map from the reconstructed texture feature map via subtracting the texture feature maps before and after downsampling to validate the effectiveness of the reconstruction from the global and local residual structure feature map. As shown in Fig.\ref{spectrogram_single-c}, \textit{2Global(Local)→Local(Global)} misses the global and local residual structure feature map augment the texture feature map, its global and local residual structure extracted from reconstructed texture feature map exhibit much bigger KL divergence value (a)(b), so the reconstructed texture feature map exhibits much bigger KL divergence value (c), representing more texture feature map loss.

\begin{figure}
\setlength{\abovecaptionskip}{0pt}
\setlength{\belowcaptionskip}{0pt}
  \centering
  \includegraphics[width=0.9\linewidth]{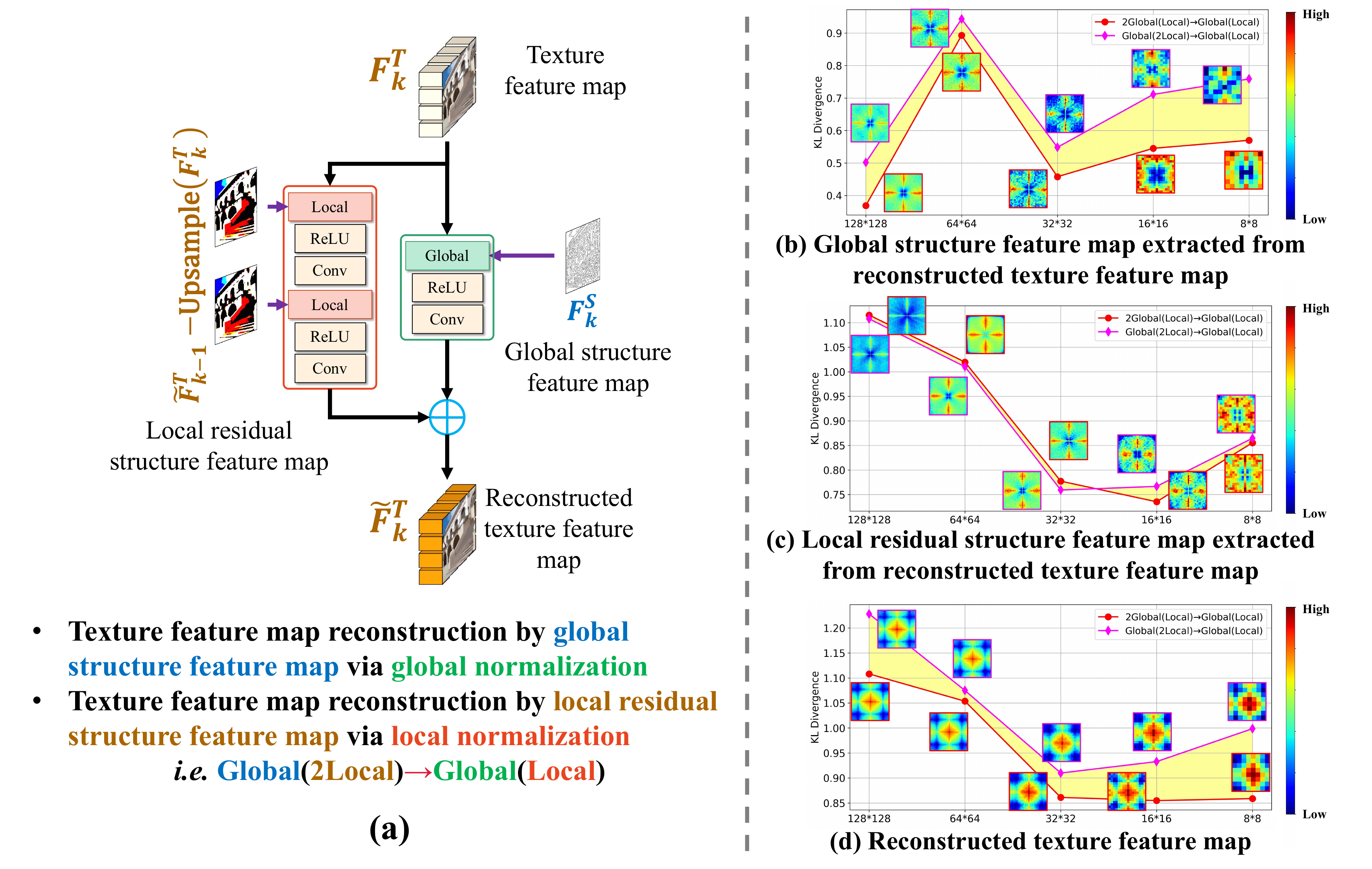}
  \caption{Illustration of \textbf{Global(2Local)→} \textbf{Global(Local)} (a); the comparison of global structure feature map (b) extracted from the reconstructed texture feature maps via high-pass filter operator and reconstructed texture feature map (d), apart from local residual structure  feature map (c) acquired by subtracting the texture feature maps before and after downsampling. Please refer to Fig.9 of the main body for more details about the heat map, particularly: the higher value within the red area indicates more structure or texture information in (b)(c) or (d) {over the feature maps}.}\label{twice}
\vspace{-15pt}
\end{figure}

Recalling Sec.II-B6 of the main body, we conduct the reconstruction from global structure feature map twice for balance during the late stage of convolutional downsampling process. To validate such intuition, we conduct a comparison over the reconstruction from local residual structure {feature map} twice (\textit{Global(2Local)→Global(Local)}), where the visual comparison of global structure feature map extracted from reconstructed texture feature map, local residual structure extracted from reconstructed texture feature map and reconstructed texture feature map with \textit{higher resolution} are illustrated in Fig.\ref{twice}, which indicates that \textit{Global(2Local)→Global(Local)} misses global structure feature map to keep balance with local residual structure feature map, its global structure feature map exhibits much bigger KL divergence value (see Fig.\ref{twice}(b)), implying more global structure feature map loss, so the reconstructed texture feature map of our architecture can exhibit much smaller KL divergence value (see Fig.\ref{twice}(d)),  implying less texture feature map loss, while possessing the same level of local residual structure feature map loss as \textit{Global(2Local)→Global(Local)} in Fig.\ref{twice}(c).

Notably, focusing on sparse local residual structural features rather than dense texture features can help alleviate texture information loss during the convolutional downsampling process, we conducted an ablation study on a variant of our architecture: texture feature map reconstruction using the global structure feature map via global normalization and texture reconstruction feature map using the texture feature map via local normalization (\textbf{2Global(Texture)} \textbf{→Global(Local)}). As shown in Table.IV of the main body, compared to \textbf{2Global(Local)→Global(Local)}, the results for \textbf{2Global(Texture)→Global(Local)} are inferior due to the lack of guidance from sparse local residual structure information, which is essential for highlighting different semantic regions in the local branch. However, the augmented local details promoted by reconstructing the local texture feature map under local normalization result in slightly better outcomes than \textbf{Global→Global}. This fact verifies that \textit{the intuition of reconstructing texture feature maps via structure feature maps on both global and local scales is critical during convolutional downsampling for image inpainting} (see Sec.II-B4). We also performed texture feature map  reconstruction using texture feature maps via global normalization (\textbf{Texture→Global}) and via local normalization (\textbf{Texture→Local}) to further validate this. The results for \textbf{Texture→Global} and \textbf{Texture→Local} were inferior to those of their corresponding comparison models, \textbf{Local→Global} and \textbf{Local→Local}, respectively. This evidence suggests that \textit{sparse structure information helps bound distinct regions within the feature map, effectively mitigating dense texture information loss during the convolutional downsampling process} (see Sec.II). Fig.16 of the main body (Columns six through nine) further supports this claim.

\begin{table}[t]
\setlength{\abovecaptionskip}{0pt}
\setlength{\belowcaptionskip}{0pt}
    \centering
    \caption{Ablation study about the cross-layer balance module for the global structure image and local residual structure image from the inpainted image. Our method, \emph{i.e.}, \textbf{Ours}, achieves the best results (reported with \textbf{boldface}). }
    \resizebox{0.9\linewidth}{!}{
    \begin{tabular}{@{}|c||c|c|c|c||c|c|c|c|@{}}
    \hline
    \multirow{2}*{Module}
    & \multicolumn{4}{c||}{{\textbf{Global Structure}}}  & \multicolumn{4}{c|}{{\textbf{Local Residual Structure}}}\\
    &PSNR$\uparrow$ &SSIM$\uparrow$ &FID$\downarrow$   &Visual     &PSNR$\uparrow$ &SSIM$\uparrow$ &FID$\downarrow$ &Visual     \\\hline\hline

    \multirow{2}*{Ours(Basic)}
    &16.81  &0.824 &15.56 &\multirow{2}{*}{\begin{minipage}{0.09\columnwidth}
        \centering
    	\raisebox{-.5\height}{\includegraphics[width=0.75\columnwidth]{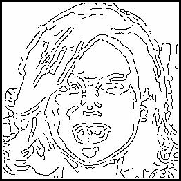}}
	   \end{minipage}} &35.59 &0.939 &4.61 &\multirow{2}{*}{\begin{minipage}{0.09\columnwidth}
        \centering
    	\raisebox{-.5\height}{\includegraphics[width=0.75\columnwidth]{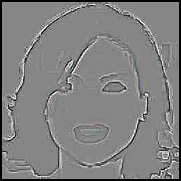}}
	   \end{minipage}}\\
    &11.58 &0.581 &32.17 &~ &29.34 &0.850 &15.46 &~\\\hline\hline

    \multirow{2}*{Ours(PENNet)}
    &17.04 &0.853 &13.21 &\multirow{2}{*}{\begin{minipage}{0.09\columnwidth}
        \centering
    	\raisebox{-.5\height}{\includegraphics[width=0.75\columnwidth]{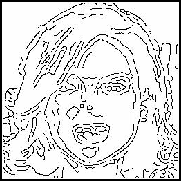}}
	   \end{minipage}} &36.85 &0.959 &4.23 &\multirow{2}{*}{\begin{minipage}{0.09\columnwidth}
        \centering
    	\raisebox{-.5\height}{\includegraphics[width=0.75\columnwidth]{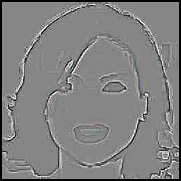}}
	   \end{minipage}}\\
    &12.29 &0.622 &30.49 &~ &30.92 &0.878 &11.02 &~\\\hline\hline

    \multirow{2}*{Ours(HiFill)}
    &17.24 &0.862 &12.60 &\multirow{2}{*}{\begin{minipage}{0.09\columnwidth}
        \centering
    	\raisebox{-.5\height}{\includegraphics[width=0.75\columnwidth]{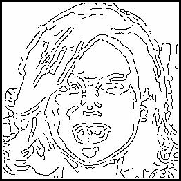}}
	   \end{minipage}} &37.49&0.957&3.68 &\multirow{2}{*}{\begin{minipage}{0.09\columnwidth}
        \centering
    	\raisebox{-.5\height}{\includegraphics[width=0.75\columnwidth]{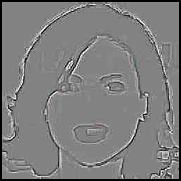}}
	   \end{minipage}}\\
    &12.41 &0.631 &29.83 &~ &31.16 &0.877 &9.88 &~\\\hline\hline

    \multirow{2}*{\textbf{\textit{Ours}}}
    &\cellcolor{green!20}\textbf{17.73} &\cellcolor{green!20}\textbf{0.876} &\cellcolor{green!20}\textbf{11.64} &\multirow{2}{*}{\begin{minipage}{0.09\columnwidth}
        \centering
    	\raisebox{-.5\height}{\includegraphics[width=0.75\columnwidth]{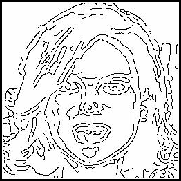}}
	   \end{minipage}} &\cellcolor{green!20}\textbf{39.15} &\cellcolor{green!20}\textbf{0.967} &\cellcolor{green!20}\textbf{2.58}  &\multirow{2}{*}{\begin{minipage}{0.09\columnwidth}
        \centering
    	\raisebox{-.5\height}{\includegraphics[width=0.75\columnwidth]{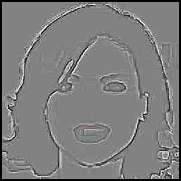}}
	   \end{minipage}}\\
    &\cellcolor{green!20}\textbf{12.95} &\cellcolor{green!20}\textbf{0.675} &\cellcolor{green!20}\textbf{27.26} &~ &\cellcolor{green!20}\textbf{31.68} &\cellcolor{green!20}\textbf{0.904} &\cellcolor{green!20}\textbf{8.14}&~ \\\hline

    \end{tabular}
  \label{table:cross}
  }
\end{table}

\begin{figure}
\setlength{\abovecaptionskip}{0pt}
\setlength{\belowcaptionskip}{0pt}
  \centering
  \includegraphics[width=0.9\linewidth]{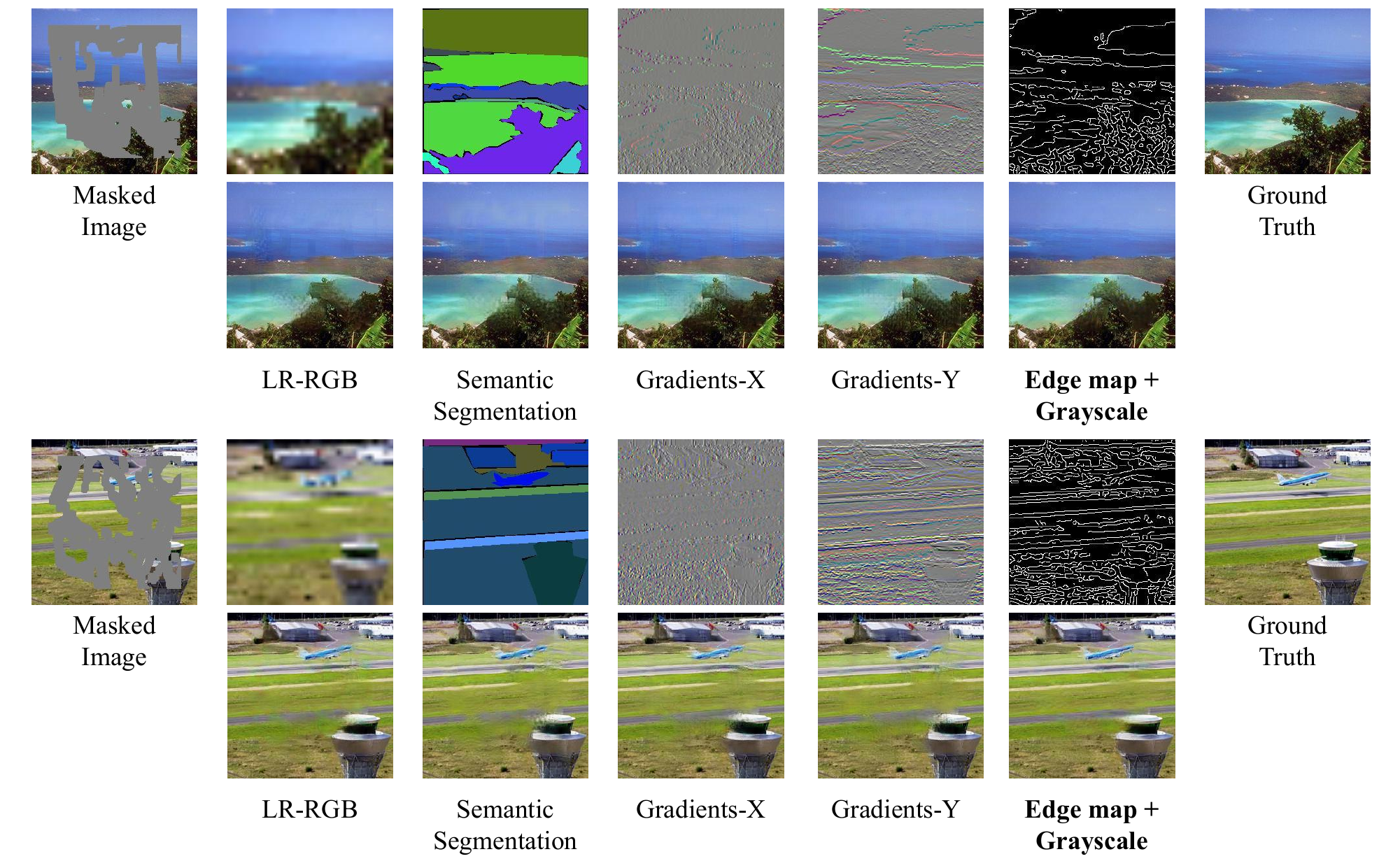}
  \caption{Ablation study about comparing the final inpainted results using various priors on ADE20K with irregular masks reveals that our \textbf{Edge map + Grayscale} approach yields the most desirable inpainting results compared to other methods.}\label{fig:prior}
\vspace{-15pt}
\end{figure}

\subsection{The structure guidance to alleviate the texture feature map loss during the convolutional downsampling process.}\label{sec5}
As mentioned in Sec.III-A1 of the main body, due to page limitation, we further provide more results to evaluate the effectiveness of using the \textbf{Edge Map + Grayscale} as the structural guidance, we conducted ablation studies comparing several structure guidance to reconstruct the texture during downsampling process. These include: 1). \textbf{Low-resolution RGB Image}: This image is downsampled from $256 \times 256$ to $32 \times 32$ at the pixel level. 2). \textbf{Semantic Segmentation Map}: We used segmentation priors from the ADE20K dataset, where colors represent different classes based on the ground truth. 3). \textbf{Gradient Map}: We employed Sobel filters to extract RGB gradients in two directions, resulting in \textbf{Gradients-X} (horizontal) and \textbf{Gradients-Y} (vertical) gradient maps. 4). \textbf{Edge Map + Grayscale}: We generated the edge map using the Canny edge detector and combined it with grayscale images, which enhance pixel intensity information for improved training. As shown in Fig.\ref{fig:prior}, our sparse structure, \textbf{Edge Map + Grayscale}, exhibits superior semantic accuracy in the inpainted results compared to other methods. Specifically, it outperforms the \textbf{Semantic Segmentation Map}, which tends to overlook fine structural details. The two types of \textbf{Gradient Maps} mainly capture local intensity changes by analyzing pixel neighborhoods, which limits their ability to capture global structures. Additionally, the \textbf{Low-resolution RGB Image} suffers from inaccurate edge information due to downsampling, resulting in inferior performance compared to our approach. Table.\ref{tab:prior} further indicates that our \textbf{Edge Map + Grayscale} yields significant improvements, with up to 26.2\% better FID, 7.6\% higher PSNR, and 3.6\% improved SSIM compared to other methods. This evidence suggests that \textit{sparse high-frequency global structure information effectively mitigates texture information loss during the convolutional downsampling process} (see Sec.II of the main body).

\subsection{Hyperparameter sensitivity analysis of $\tau$ in adjusted transformer blocks.}\label{sec6}
As mentioned in Sec.III-A2 of the main body, due to page limitation, we further provide more results to analyse the parameter $\tau$ in adjusted transformer blocks. This parameter $\tau$ is responsible for suppressing the attention scores of masked texture patches and ensuring that the unmasked patches are primarily reconstructed by semantic content. We set the $\tau$ values to \{0; 50; 100; 150; 200\}. As shown in Fig.\ref{fig:tau}, the best performance is achieved with $\tau=100$ across various mask ratios, where the attention scores of the masked patch are nearly zero, which can be attributed to the following reasons: 1) When $\tau$ is too small  (e.g. 0 and 50), the blank semantics of the masked patches interfere with the connections between unmasked patches. 2) Conversely, when $\tau$ is too large  (e.g. 150 and 200), it exaggerates the differences in attention scores between patches, limiting each patch to form strong connections with only a few others, thus weakening the global correlations. Based on different values of $\tau$, we balance the extraction of global correlations while avoiding the influence of the masked patches.

\begin{table}
\setlength{\abovecaptionskip}{0pt}
\setlength{\belowcaptionskip}{0pt}
    \scriptsize
    \centering
    \caption{Ablation study about several priors as the structure information on ADE20K dataset. Our method, \emph{i.e.}, \textbf{Edge map + Grayscale}, achieves the best results (reported with \textbf{boldface}).}
    \resizebox{0.9\linewidth}{!}{
    \begin{tabular}{@{}|c||c|c||c|c||c|c|@{}}
    \hline
    \multirow{2}*{{Priors}} & \multicolumn{2}{c||}{{PSNR$\uparrow$}}  &\multicolumn{2}{c||}{{SSIM$\uparrow$}}& \multicolumn{2}{c|}{{FID$\downarrow$} }\\
    &{20-30\%} &{40-50\%}  &{20-30\%} &{40-50\%}  &{20-30\%} &{40-50\%}  \\\hline\hline
    Low-Resolution RGB  &30.16 &23.83 &0.924 &0.857 &38.69 &80.71 \\\hline
    Semantic Segmentation  &29.22 &24.33 &0.908 &0.849 &44.21 &92.16 \\\hline
    Gradients-X  &28.44 &24.22 &0.900 &0.842 &47.84 &94.06 \\\hline
    Gradients-Y  &29.24 &24.42 &0.910 &0.849 &46.64 &90.76 \\\hline
    \cellcolor{green!20}\textbf{Edge map + Grayscale} &\cellcolor{green!20}\textbf{30.62} &\cellcolor{green!20}\textbf{24.72} &\cellcolor{green!20}\textbf{0.933} &\cellcolor{green!20}\textbf{0.865} &\cellcolor{green!20}\textbf{35.27} &\cellcolor{green!20}\textbf{72.25} \\\hline
    \end{tabular}
    \label{tab:prior}
    }
\vspace{-15pt}
\end{table}

\begin{figure}
\setlength{\abovecaptionskip}{0pt}
\setlength{\belowcaptionskip}{0pt}
  \centering
  \includegraphics[width=0.8\linewidth]{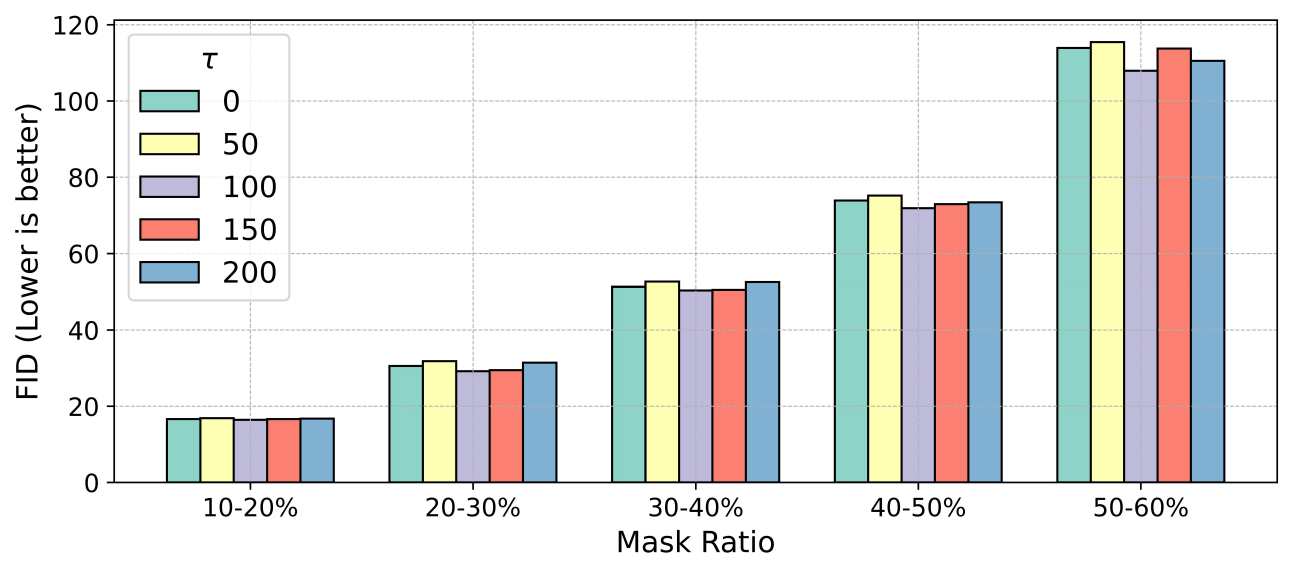}
  \caption{Ablation study about hyperparameter $\tau$ under varied mask ratios on PSV, Our method, \emph{i.e.}, \textbf{$\tau = 100$}, achieves the best results}\label{fig:tau}
\vspace{-15pt}
\end{figure}

\end{document}